%% file: main.tex
\documentclass[preprint,5p,twocolumn]{elsarticle}
\usepackage[english]{babel}
\usepackage{amsmath}
\usepackage{graphicx}
\usepackage{color}       
\usepackage{transparent} 
\usepackage{xcolor}
\usepackage{booktabs}
\usepackage{tabularx}
\usepackage{siunitx}
\usepackage[colorlinks=true,allcolors=blue]{hyperref}
\usepackage[textsize=small]{todonotes}
\usepackage{framed} 
\usepackage{multicol} 
\usepackage{nomencl} 
\usepackage{ifthen}
\makenomenclature

\renewcommand*{\nompreamble}{%
  \markboth{\nomname}{\nomname}%
  \begin{multicols}{2}%
  \setlength{\columnsep}{1em}%
  \raggedcolumns%
}
\renewcommand*{\nompostamble}{%
  \end{multicols}%
}

\renewcommand{\nomgroup}[1]{%
  \item[\bfseries
    \ifthenelse{\equal{#1}{A}}{Acronyms}{%
    \ifthenelse{\equal{#1}{L}}{Latin symbols}{%
    \ifthenelse{\equal{#1}{G}}{Greek symbols}{%
    \ifthenelse{\equal{#1}{S}}{Subscripts / qualifiers}{Other}}}}%
  ]%
}

\graphicspath{{./figures/}}

\newif\ifcoauthornotes
\makeatletter
\def\ps@pprintTitle{%
  \let\@oddhead\@empty
  \let\@evenhead\@empty
    \def\@oddfoot{\footnotesize\itshape Neubauer et al.: Preprint submitted to Energy and Buildings\hfill August 31, 2026}%
  \let\@evenfoot\@oddfoot}
\makeatother

\begin{document}

\onecolumn

\ifcoauthornotes
\section*{Notes and Open Questions for Co-Authors}
\begin{itemize}

  \item \textbf{General information}
    \begin{itemize}
      \item The intro should now read as one story, not a list of facts.
      \item The field itself shifted from M1 to M2 over 2023--2026 (documented in the corpus).
      \item Emerging topics (RAG, tool use, agentic workflows, MCP) are covered under M2.
      \item Novelty vs.\ other reviews: no prior review jointly codes task family, LLM method, evidence realism, deployment readiness, and responsibility boundary.
      \item Other reviews are broader (AECO/AI) or algorithm-focused; this one is HVAC-specific and deployment-focused.
      \item The status of the papers is current as of April 1. I will likely update the papers again during the revision process.
    \end{itemize}

  \item \textbf{Open questions}
    \begin{itemize}
      \item Does the intro story work?
      \item Is the M1--M2 field-shift argument convincing?
      \item Any current topics still missing?
    \end{itemize}

  \item \textbf{Tasks}
    \begin{itemize}
      \item Check the intro story
      \item Give feedback in general
    \end{itemize}
    
\vspace{1cm}
\textbf{Thank you for your comments and support}
\end{itemize}
\fi 

\clearpage

\begin{frontmatter}

\title{Large Language Models for HVAC Operations in Building Energy Systems: A Critical Review of Methods, Applications, and Deployment Readiness}

\author[tu,lb]{Alexander Neubauer\corref{cor1}}
\ead{a.neubauer@tu-berlin.de}
\author[lb]{Tianzhen Hong}
\author[lb]{Han Li}
\author[tu]{Mengbo Yu}
\author[tu]{Amin Darbandi}
\author[tu]{Yannick Fürst}
\author[tu]{Martin Kriegel}
\cortext[cor1]{Corresponding author at: TU Berlin, Marchstr. 4, 10587 Berlin, Germany}
\address[tu]{Hermann-Rietschel-Institut, Technische Universität Berlin, 10587 Berlin, Germany}
\address[lb]{Lawrence Berkeley National Laboratory, Berkeley, CA 94720, USA}
\setcounter{page}{0}

\begin{abstract}
Building automation systems generate rich sensor data yet remain insight-poor because heterogeneous point naming, missing metadata, and fragmented documentation obstruct their operational use. This systematic review analyses and codes 66 peer-reviewed studies on large language models (LLMs) for HVAC operations published between 2023 and March~2026. Each study is classified across five application families and three LLM method families and assessed for evidence realism, deployment readiness, and the responsibility boundary between the LLM and physical HVAC decisions. The corpus is concentrated in building energy modelling (BEM, 32 of 66 papers), while load forecasting remains too sparse for subfield-level conclusions. Only four studies reach pilot-level evidence, and none reports sustained operational deployment. No study was classified as ready-now for industry adoption; three were near-term and 63 research-only. Nevertheless, several bounded, human-in-the-loop uses merit near-term trials, including point-name normalisation, document-grounded operator support, BEM workflow assistance, and advisory interfaces around physics-based controllers. Conventional machine learning (ML), model predictive control (MPC), reinforcement learning (RL) and ontology-based tools remain more adopted for high-frequency control, short-horizon numerical forecasting, and well-posed ontology mapping, while autonomous agentic operation and unvalidated occupant proxies remain research-stage. Current evidence therefore supports LLMs primarily as semantic and workflow layers rather than autonomous HVAC controllers. Future work should prioritise field-validated benchmarks, orchestration evaluation under operational constraints, and LLM--MPC/RL architectures with bounded latency and verifiable safety properties.
\end{abstract}

\begin{graphicalabstract}
\centering
\resizebox{0.99\textwidth}{!}{\fontsize{6pt}{6pt}\selectfont{
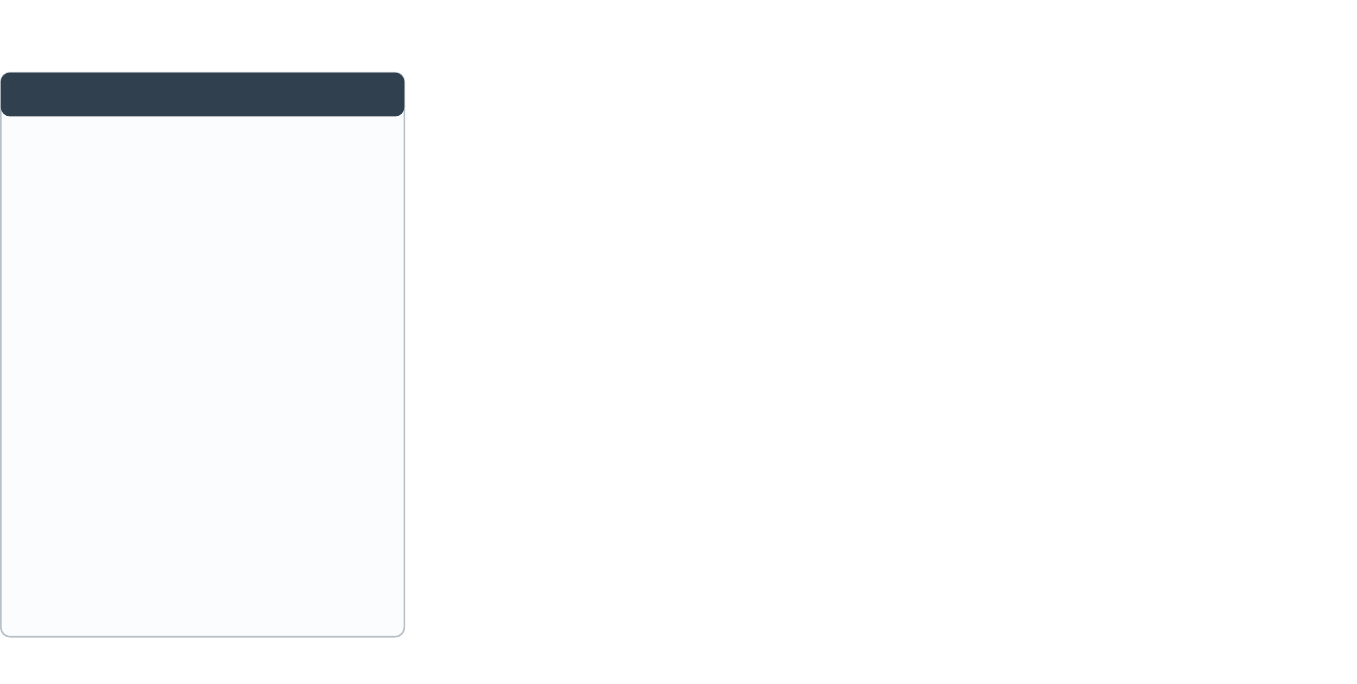
}}
\end{graphicalabstract}

\onecolumn
\begin{highlights}
    \item 66 LLM-for-HVAC studies coded across 5 application and 3 method categories
    \item No study qualified as ready-now; 3 were near-term and 63 research-only
    \item Framework links evidence realism, readiness, and responsibility boundaries
    \item Bounded trial roles include point-name cleanup, operator support, and BEM assistance
    \item Fit-for-purpose map: ML, MPC and RL remain popular applications of AI for HVAC 
\end{highlights}
\clearpage
\twocolumn

\thispagestyle{pprintTitle}

\begin{keyword}
Large language models \sep Generative artificial intelligence \sep Retrieval-augmented generation \sep LLM agents \sep Building energy modelling \sep Fault detection and diagnosis \sep Systematic review
\end{keyword}

\end{frontmatter}

\nomenclature[A]{AECO}{Architecture, engineering, construction, and operations}
\nomenclature[A]{AHU}{Air handling unit}
\nomenclature[A]{AI}{Artificial intelligence}
\nomenclature[A]{API}{Application programming interface}
\nomenclature[A]{ASHRAE}{American Society of Heating, Refrigerating and Air-Conditioning Engineers}
\nomenclature[A]{BAS}{Building automation system}
\nomenclature[A]{BEM}{Building energy modelling}
\nomenclature[A]{BEMS}{Building energy management system}
\nomenclature[A]{BERT}{Bidirectional Encoder Representations from Transformers}
\nomenclature[A]{BIM}{Building information modelling}
\nomenclature[A]{BMS}{Building management system}
\nomenclature[A]{BOPTEST}{Building Optimization Performance Test framework}
\nomenclature[A]{Brick}{Brick schema for semantic building metadata}
\nomenclature[A]{CRPS}{Continuous ranked probability score}
\nomenclature[A]{FDD}{Fault detection and diagnosis}
\nomenclature[A]{GC-ARM}{Graph-constrained association rule mining}
\nomenclature[A]{GDPR}{General Data Protection Regulation}
\nomenclature[A]{GPT}{Generative pre-trained transformer}
\nomenclature[A]{HVAC}{Heating, ventilation, and air conditioning}
\nomenclature[A]{ICL}{In-context learning}
\nomenclature[A]{IDF}{Input Data File (EnergyPlus model file format)}
\nomenclature[A]{IEQ}{Indoor environmental quality}
\nomenclature[A]{LBNL}{Lawrence Berkeley National Laboratory}
\nomenclature[A]{LLM}{Large language model}
\nomenclature[A]{LoRA}{Low-rank adaptation}
\nomenclature[A]{MAE}{Mean absolute error}
\nomenclature[A]{MCP}{Model Context Protocol}
\nomenclature[A]{ML}{Machine learning}
\nomenclature[A]{MPC}{Model predictive control}
\nomenclature[A]{NL}{Natural language}
\nomenclature[A]{O\&M}{Operations and maintenance}
\nomenclature[A]{PEFT}{Parameter-efficient fine-tuning}
\nomenclature[A]{PI}{Proportional-integral}
\nomenclature[A]{PMV}{Predicted mean vote}
\nomenclature[A]{PRISMA}{Preferred Reporting Items for Systematic Reviews and Meta-Analyses}
\nomenclature[A]{RAG}{Retrieval-augmented generation}
\nomenclature[A]{Re2G}{Retrieve-rerank-generate}
\nomenclature[A]{RL}{Reinforcement learning}
\nomenclature[A]{RMSE}{Root mean square error}
\nomenclature[A]{RP-1312}{ASHRAE research project 1312 fault-detection benchmark}
\nomenclature[A]{SMOTE}{Synthetic Minority Over-sampling Technique}
\nomenclature[A]{VAV}{Variable air volume}

\nomenclature[S]{A0}{Background papers: reviews, infrastructure, or semantic middleware}
\nomenclature[S]{A1}{Fault detection and diagnosis application category}
\nomenclature[S]{A2}{Energy load forecasting application category}
\nomenclature[S]{A3}{Building energy modelling and simulation application category}
\nomenclature[S]{A4}{HVAC control and optimisation application category}
\nomenclature[S]{A5}{Thermal comfort and occupant interaction application category}
\nomenclature[S]{M1}{Prompting and supervised adaptation method category}
\nomenclature[S]{M2}{Context-grounded orchestration method category}
\nomenclature[S]{M3}{Multimodal input method category}
\nomenclature[S]{FM}{Failure mode}
\nomenclature[S]{FW}{Future-work priority}
\nomenclature[S]{RQ}{Research question}

\nomenclature[L]{$A \times M$}{Cross-tabulation of application categories and method categories}
\nomenclature[L]{$F_1$}{F1 score}
\nomenclature[L]{$n$}{Number of papers in a category or cell}
\nomenclature[L]{$N$}{Total number of papers in the reviewed corpus}
\nomenclature[L]{$R^2$}{Coefficient of determination}
\nomenclature[X]{$\sim$}{Approximately}

\begin{table*}[t]
\begin{framed}
\begin{scriptsize}
\makeatletter\nom@tempdim=1.2cm\makeatother\input{main.nls} 
\end{scriptsize}
\end{framed}
\end{table*}

\section{Introduction}\label{sec:intro}
Modern buildings are rich in data but poor in insight. A building automation system records thousands of points, yet the information an engineer needs to diagnose a fault or tune a controller is scattered across cryptic, vendor-specific point names, missing metadata, and unstructured manuals that analytics pipelines cannot read directly~\cite{zhang_opportunities_2025}. This fragmentation limits not only diagnostics and maintenance, but also the deployment of advanced analytics and control methods. Approaches such as model predictive control (MPC) and reinforcement learning (RL) have shown strong performance in controlled, largely simulated evaluations~\cite{amangeldy_review_2025,amangeldy_ai-powered_2025}, yet their
large-scale adoption in routine building operation remains limited. The binding
obstacles are institutional and operational rather than numerical: buildings are
often decades old, repeatedly retrofitted, and insufficiently documented, so
integration effort frequently exceeds the projected savings, and the specialised
expertise needed to bridge this gap is unevenly distributed and difficult to
retain~\cite{zhang_opportunities_2025}. Three barriers summarise the deployment
gap: \emph{trust} (operators cannot interpret or override black-box controllers),
\emph{integration} (heterogeneous point naming and missing metadata prevent clean data pipelines), and \emph{governance} (liability and auditability requirements are difficult to satisfy when machine learning (ML) systems are opaque), barriers that extend beyond numerical optimisation to include semantic, technical, organisational, and governance constraints.
Because heating, ventilation, and air-conditioning (HVAC) systems are among the largest end uses in buildings~\cite{gonzalez_building-review_2022} and a primary lever for electrification and demand flexibility~\cite{neubauer_heatshift_2026}, the consequence extends beyond individual buildings: unresolved semantic bottlenecks slow electrification, limit demand-response participation, and constrain portfolio-scale deployment of efficient control and retrofit strategies. Meanwhile, large language model (LLM) capabilities are advancing across scientific and energy-sector domains~\cite{minaee_llm-survey_2025,iea_energy-ai_2026}; given the scale of HVAC energy use, this progress bears directly on building operations.

LLMs may help address these bottlenecks through language-based interpretation and coordination of heterogeneous information. Their value in HVAC operations lies not primarily in replacing physics-based controllers, but in acting as a \emph{semantic and workflow layer}. It connects fragmented information sources such as manufacturer manuals, building automation system (BAS) point data, building information modelling (BIM) models, semantic ontologies, work-order histories, maintenance logs, and historical operational records, and makes that information accessible in forms that operators, technicians, and engineers can act on. Building operations and maintenance (O\&M) practice is strongly document-centred: diagnostic and maintenance decisions draw on equipment manuals, inspection records, and institutional knowledge that are rarely integrated into a single accessible system. LLMs have the potential to partially address this gap, not by automating decisions, but by improving access to the information needed to make them. 

Early LLM-for-HVAC work focused primarily on prompting and fine-tuning
approaches (M1), showing that pre-trained models can classify faults,
generate energy-model code, and answer domain-specific queries with limited
task-specific training. More recent work increasingly relies on
\emph{context-grounded orchestration} (M2), in which LLMs are connected to
curated knowledge bases, external tools, simulation environments, sensor
databases, and building documents. In such systems, performance depends not
only on the underlying model, but also on retrieval quality, tool-interface
design, and orchestration logic. These approaches can ground outputs in
building-specific information and make retrieved sources, tool calls, and
intermediate artefacts more traceable. At the same time, they introduce new
failure modes related to incomplete retrieval, tool misuse, latency, and
error propagation across multi-step workflows.  A third, smaller line of work extends these systems beyond text through multimodal inputs such as floor plans, piping and instrumentation diagrams, and spatial heatmaps processed by vision--language models (M3). The transition from M1 toward M2 mirrors a broader shift in LLM practice from question-answering-oriented use, in which the model classifies, explains, or answers, toward task-oriented systems that plan and act through tools. Within the peer-reviewed HVAC corpus, this shift is traced in Section~\ref{sec:llm_trends}.

The rapid availability of LLM-based tools for building management has outpaced evidence-based deployment guidance. Practitioners currently lack a systematic basis for distinguishing which LLM roles are suitable for bounded field use, which require further validation, and which are not yet supported for safety-relevant building applications. This paper addresses that gap by organising the growing literature on LLMs for HVAC operations into a structured analytical framework. The central question is therefore not simply whether LLMs can support building operations, but where they can credibly improve operational decision-making without introducing unacceptable safety, governance, or reproducibility risks. A related fit-for-purpose question is equally important: which HVAC tasks benefit specifically from language-based interpretation and orchestration, and which are better handled by conventional ML, control, optimisation, or ontology-based tools?

Five HVAC application (A) categories are identified: A1 fault detection and diagnosis (FDD), A2 energy load forecasting (Forecast), A3 building energy modelling (BEM) and simulation, A4 control and optimisation (Control), A5 thermal comfort and occupant interaction (Comfort). Together with the three method categories M1--M3 introduced above and formally defined in Section~\ref{sec:background_methods}, each included study is located on this A$\times$M grid.

A critical design constraint shapes the analysis throughout. HVAC operation is \emph{physics-constrained, time-dependent, and safety-relevant}: plausibility is not an adequate success criterion. Current fault-diagnosis evidence already shows why: \citet{zhang_domain-specific_2025} demonstrate that near-perfect benchmark accuracy on a fine-tuned FDD model does not transfer across equipment types without separate retraining, and \citet{xiao_exploring_2024} report run-to-run variation in the recommendations generated by a multi-agent optimisation workflow. This instability coexists with strong headline accuracy numbers, which is why standard metrics (F1, R\textsuperscript{2}, energy savings percentage) are insufficient to characterise \emph{deployment readiness}.

The study's main practical output is a map that distinguishes bounded, human-supervised roles supported by the current evidence from roles that require additional safety mechanisms, field validation, and governance frameworks before deployment. The analysis also identifies broader patterns across the A$\times$M grid, including the growing prevalence of context-grounded orchestration, the concentration of studies in building energy modelling, and the persistent lack of real-building validation. Finally, recurring failure modes are translated into concrete future-work priorities derived from the limitations observed across the reviewed corpus.

Four research questions (RQ) organise the synthesis:
\begin{description}
\item[\textbf{RQ1}] What is the distribution of studies across application categories (A1--A5) and LLM method categories (M1--M3), and what structural gaps remain unexplored?
\item[\textbf{RQ2}] What is the deployment maturity of reviewed LLM roles, assessed against evidence realism, closed-loop validation status, and safety-critical responsibility boundaries?
\item[\textbf{RQ3}] What recurring failure modes and cross-cutting barriers constrain broader deployment, and which future-work directions address them most directly?
\item[\textbf{RQ4}] Which HVAC tasks benefit specifically from language reasoning, and where do conventional ML, control, optimisation, or ontology tools remain the more appropriate technical choice?
\end{description}

To address these questions, the paper makes four contributions:
\begin{enumerate}
\item A \emph{task-structured, evidence-coded} systematic review of 66 primary studies mapping the A1--A5~$\times$~M1--M3 grid with explicit \emph{evidence realism} classification, and an analysis of adoption trends and methodological shifts, including the frontier shift from prompt-centred toward context-grounded orchestration, across the 2023--2026 publication window.

\item A deployment-readiness assessment that classifies the reviewed studies as \emph{ready-now}, \emph{near-term}, or \emph{research-only} based on evidence realism, closed-loop validation, role boundedness, and the responsibility boundary between LLM outputs and physical HVAC decisions.

\item A \emph{safety-aware} analysis of responsibility boundaries and failure modes relevant to \emph{cyber-physical} HVAC systems.

\item Concrete practitioner guidance for five \emph{low-risk, bounded workflow roles} that merit near-term practitioner attention, an assessment of what is already usable to support the energy transition, and four actionable future-work priorities derived from identified failure modes and field evidence gaps.
\end{enumerate}

Beyond mapping the literature, this review distinguishes between \emph{technical plausibility} and \emph{deployment defensibility} in HVAC LLM applications. It therefore contributes not only an application--method taxonomy, but also an operational interpretation layer that links evidence realism, safety-criticality, and responsibility boundaries to concrete near-term deployment decisions.

The paper proceeds as follows. Background and methodology are covered in Sections~\ref{sec:background}--\ref{sec:method}; the search strategy, screening protocol, and coding scheme are described there. Each application category is then analysed in Sections~\ref{sec:a1}--\ref{sec:a5}, followed by cross-cutting synthesis in Section~\ref{sec:crosscutting}. The discussion (Section~\ref{sec:discussion}) translates the evidence into deployment guidance, a responsibility boundary framework, and future-work priorities, and Section~\ref{sec:conclusion} concludes.

\section{Background}\label{sec:background}
This section establishes the domain context and methodological vocabulary used throughout the review. Operational HVAC bottlenecks motivating LLM adoption are characterised in Section~\ref{sec:background_hvac}; the three LLM adaptation categories M1--M3 that structure the analysis are defined in Section~\ref{sec:background_methods}; and the relationship to adjacent reviews is discussed in Section~\ref{sec:background_reviews}.

\subsection{HVAC Systems and Operational Bottlenecks}\label{sec:background_hvac}
HVAC systems are typically among the largest end uses in buildings, accounting for approximately \qty{38}{\percent} of building energy use globally, though the exact share varies considerably with building type, climate, and use pattern~\cite{gonzalez_building-review_2022}. They are central to three intersecting priorities: electrification of heating, demand flexibility for grid-interactive operation, and occupant comfort and indoor air quality. Digital control capabilities have expanded substantially, but the practical gap between installed capability and realised performance remains wide. Retrofit analysis, fault resolution and operational optimisation all require domain expertise, which is lacking among most facility management teams. This creates a persistent barrier between simulation models and field decisions~\cite{amangeldy_review_2025,zhang_opportunities_2025}.
Typical building O\&M practice remains strongly preventive, inspection-based, and document-centred. Facility teams rely on asset inventories, condition reports, log books, scheduled inspections, work-order systems, and escalation to trained specialists to organise maintenance activities. 

The integration barrier is technical in origin: heterogeneous point naming across manufacturers, missing or inconsistent metadata, and the absence of shared ontologies mean that analytics pipelines must be rebuilt for every building~\cite{zheng_mastering_2025}. The trust barrier is operational: operators need to understand, override, and audit automated control decisions, and opaque numeric controllers cannot provide that transparency~\cite{liu_large_2025}. Both barriers are fundamentally semantic, and this is precisely where LLMs, with their natural-language competence, add a capability that numeric controllers have never had.

\begin{figure*}[t!]
\centering
\resizebox{0.9\textwidth}{!}{\fontsize{5.5pt}{5.5pt}\selectfont{
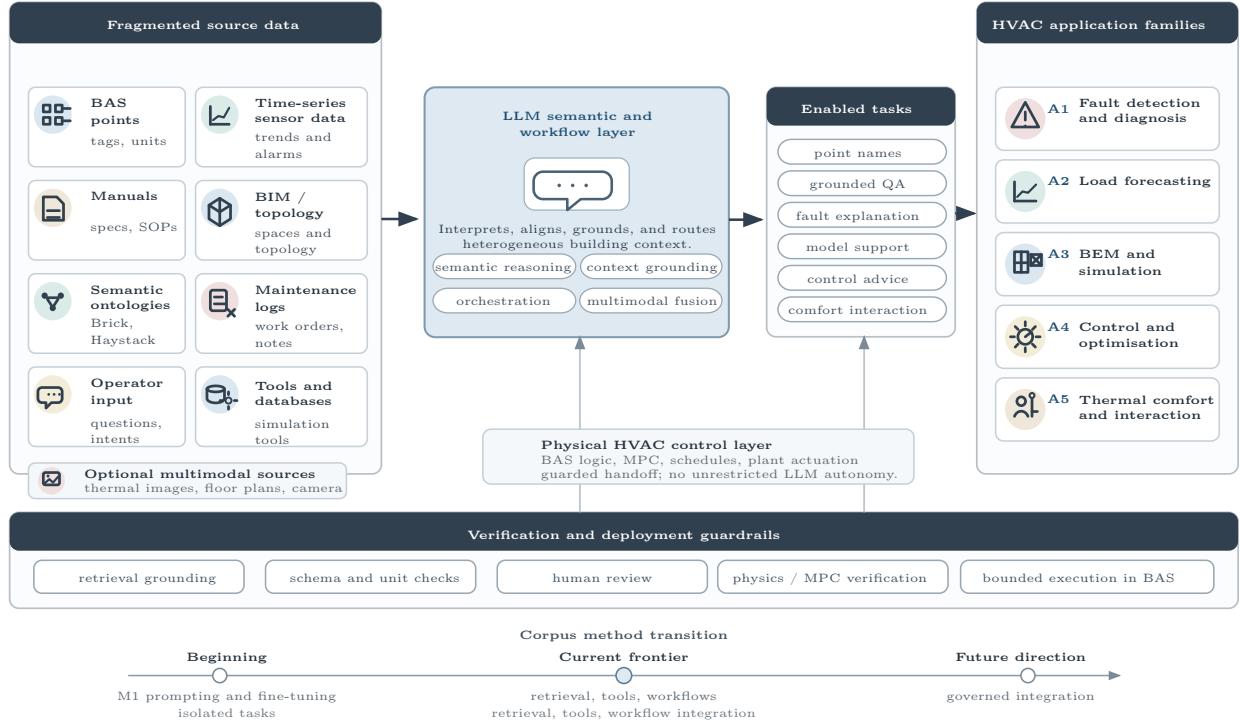
}}
\caption{%
  Conceptual role of the LLM semantic and workflow layer in HVAC operations.
  Fragmented building information sources (left) are interpreted, aligned,
  and routed by the LLM layer (centre) toward five HVAC application categories
  (right). Verification and deployment guardrails separate the LLM semantic
  zone from bounded physical actuation. The timeline (bottom) reflects the
  observed corpus shift from prompt-centred approaches (M1) toward
  context-grounded orchestration (M2) and the emerging priority of governed
  harness engineering.
}\label{fig:llm_semantic_workflow}
\end{figure*}

\subsection{LLM Adaptation Methods: M1--M3}\label{sec:background_methods}
The three method categories classify studies according to their dominant integration logic rather than their performance quality or every individual algorithmic subtype. M1 retains the pre-trained language model as the primary source of task knowledge and adapts its behaviour through prompting or supervised training. M2 grounds model outputs in externally curated information and surrounding orchestration, including retrieval, memory, tool use, and multi-step workflows. M3 extends the system through non-textual input modalities, such as images, plans, and spatial heatmaps, processed by a vision--language encoder. 

The M1/M2 boundary is defined architecturally rather than mechanistically. M2 applies when task-relevant context enters at run time through maintained infrastructure such as a retrieval index, a database, a tool interface, or an orchestration loop. Prompt-only and fine-tuned systems achieve task competence in fundamentally different ways, yet both remain M1 because neither depends on such run-time infrastructure. A one-way hand-off to a downstream solver is not orchestration and remains M1. Because real systems fall along a spectrum, borderline cases were assigned by this criterion (Section~\ref{sec:method_coding}).

These review-level categories preserve the principal architectural distinctions while avoiding sparse and unstable subcategories within the 66-paper corpus. Table~\ref{tab:method_categories} summarises the three categories.

\textit{M1: Prompting and supervised adaptation} encompasses in-context learning (ICL), few-shot and zero-shot prompting, prompt engineering, and supervised fine-tuning, including parameter-efficient methods such as Low-Rank Adaptation (LoRA) and Parameter-Efficient Fine-Tuning (PEFT). These approaches either exploit knowledge encoded in a pre-trained model with minimal domain-specific data or adapt the model using labelled task examples. A representative HVAC example is the LoRA-fine-tuned GPT-3.5 developed by \citet{zhang_domain-specific_2025}, which increased FDD accuracy from \qty{29.5}{\percent} to nearly \qty{100}{\percent} on the American Society of Heating, Refrigerating and Air-Conditioning Engineers (ASHRAE) RP-1312 benchmark while generalising to unseen air handling unit (AHU) datasets. Transfer to other equipment classes, including variable air volume (VAV) and chiller systems, nevertheless required separate fine-tuning.

\textit{M2: Context-grounded orchestration} augments LLMs with externally curated context, such as manuals, BIM documents, ontologies, sensor logs, memory, and tool outputs. These systems frequently provide tool-calling capabilities, including code execution, simulation APIs, and database queries. More complex implementations use multi-step orchestration workflows in which intermediate retrievals, tool outputs, or model responses are fed back into the process. Retrieval-augmented generation (RAG) is therefore treated not as a separate category but as one grounding mechanism within the broader category of context- and orchestration-driven systems. A representative HVAC example is the four-agent workflow developed by \citet{zhang_automatic_2025}, which achieves a \qty{100}{\percent} Input Data File (IDF) success rate and reduces expert BEM authoring time from four hours to nine minutes by generating, executing, and debugging EnergyPlus models.

\textit{M3: Multimodal input approaches} combine vision--language models with non-textual inputs such as floor plans, piping and instrumentation diagrams, spatial heatmaps, sensor plots, or video. These approaches extend the information modalities available to the LLM beyond text. The category is defined by input modality alone: run-time coupling of an LLM to a physics-based model, an optimiser, or a control layer is an orchestration pattern and is coded M2, so a text-only system that invokes a simulation tool within its run-time workflow is M2 rather than M3. A representative HVAC example is the Office-in-the-Loop system developed by \citet{sawada_office---loop_2025}, which integrates spatial overhead heatmaps, occupant feedback, and IoT sensor data as multimodal prompts to a model that issues HVAC setpoints in a real office.

\begin{table*}[t]\footnotesize
\centering
\caption{Comparison of LLM adaptation categories and their HVAC deployment characteristics.}\label{tab:method_categories}
\renewcommand{\arraystretch}{1.35}
\begin{tabularx}{\textwidth}{p{2.5cm} X p{3cm} p{3cm} p{3.5cm}}
\toprule
Category & Mechanisms & Typical HVAC inputs & Primary risk & Representative HVAC use \\
\midrule
M1 Prompting / Supervised &
  ICL, prompt constraints, fine-tuning, PEFT &
  Instructions, labelled sensor data &
  Prompt brittleness; dataset shift &
  FDD classification; BEM code generation \\
M2 Context-grounded orchestration &
  Retrieval over manuals, BIM documents, ontologies, memory, and sensor logs; tool-calling interfaces; agentic loops and multi-step debugging workflows &
  Building documents; ontologies; memory; tool outputs &
  Retrieval errors; latency; error propagation across chained steps &
  Diagnostic retrieval; multi-step BEM workflows; model context protocol (MCP)-enabled simulation access \\
M3 Multimodal input &
 Vision--language encoding of non-textual inputs &
  Floor plans; P\&ID diagrams; spatial heatmaps; sensor plots; video &
  Integration complexity; governance &
  Multimodal setpoint supervision; video-derived modelling guidance \\
\bottomrule
\end{tabularx}
\end{table*}

The M1--M3 categories describe how LLMs acquire domain competence, but the operational challenge they address is common to all three: building environments accumulate fragmented, heterogeneous information (BAS point lists, equipment manuals, BIM models, sensor histories, semantic ontologies, and operator queries) that no single pipeline integrates automatically.
Figure~\ref{fig:llm_semantic_workflow} situates the LLM within this environment as a semantic and workflow layer that interprets, aligns, and routes context from these fragmented sources toward the five HVAC application categories reviewed here.
Two constraints frame the entire analysis and recur throughout this review: LLM outputs are channelled through verification guardrails before reaching physical actuation, which keeps the model in a bounded rather than autonomous role; and the predominant integration approach has shifted from prompt-centred methods (M1) toward context-grounded orchestration (M2) over the 2023--2026 corpus.

\subsection{Existing Reviews and Positioning}\label{sec:background_reviews}
Existing reviews cluster into three adjacent but incomplete perspectives. First, \emph{broad building-energy and building-industry reviews} such as \citet{zhang_opportunities_2025}, \citet{liu_large_2025}, \citet{arslan_large_2026}, and \citet{liang_pre-training_2025} map opportunities, challenges, or model families across a wide application space, but they do not focus specifically on HVAC operations as a safety- and latency-constrained deployment context. Second, \emph{adjacent smart-building control reviews} such as \citet{amangeldy_review_2025} analyse artificial intelligence (AI), deep reinforcement learning, and resource-management strategies in depth, but they primarily evaluate numeric control paradigms rather than LLM-specific operational roles. Third, \emph{perspective-style review papers} such as \citet{ma_ten_2026} surface important governance and evaluation questions, but they are not structured as coded corpus syntheses. A fourth strand reviews other AI modalities entering HVAC practice, such as robotics for inspection and maintenance~\cite{jiang_robotics_2026}. These address physical rather than language-based automation and are therefore outside the scope adopted here.

This fragmentation explains why no prior review combines the dimensions used here. Reviews with very broad scope prioritise coverage across buildings, architecture, engineering, construction, and operations (AECO), or AI categories; controller-focused reviews prioritise algorithmic performance; and perspective papers prioritise open questions over paper-level coding. As a result, the literature has not yet been synthesised through the joint lens of \emph{task family}, \emph{LLM integration method}, \emph{evidence realism}, \emph{deployment constraint}, and \emph{responsibility boundary}. This paper addresses that \emph{gap} by narrowing the focus to HVAC operations and coding each study along exactly those dimensions, which are the ones most relevant for real deployment decisions. The A$\times$M grid therefore complements rather than replaces model-family and control-performance taxonomies: it re-organises the same studies by operational task, integration pattern, and deployment evidence.

\section{Methodology}\label{sec:method}

The methodology is described in two parts: (i) the search strategy and corpus
construction (Section~\ref{sec:method_search}), and (ii) the taxonomy and coding scheme applied to classify each paper by application category, LLM method, evidence type, and deployment constraint (Section~\ref{sec:method_coding}).

\subsection{Search Strategy and Corpus}
\label{sec:method_search}

The following Scopus query was executed to identify candidate papers:

\begin{quote}
\scriptsize\ttfamily\raggedright
TITLE-ABS-KEY(\\
\quad (``large language model*'' OR LLM* OR GPT* OR ChatGPT OR\\
\quad\quad ``generative AI'' OR RAG OR ``retrieval augmented generation'' OR\\
\quad\quad ``retrieval-augmented generation'')\\
\quad AND (HVAC OR ``building energy'' OR ``building performance'' OR\\
\quad\quad ``building management system*'')\\
)\\
\quad AND (DOCTYPE(ar) OR DOCTYPE(re) OR DOCTYPE(cp))
\end{quote}

The query combines (i)~LLM and generative-AI terms with (ii)~HVAC / building-energy context terms, followed by a document-type restriction to articles, reviews, and conference papers; this design was meant to recover papers that make a substantive contribution at
the intersection of LLM and HVAC or building-energy systems without prematurely narrowing the corpus through task-specific keywords. The evidence base was deliberately limited to Scopus-indexed, English-language, peer-reviewed publications. This boundary serves two purposes: (i)~it restricts the corpus to contributions with a minimum level of reviewed evaluative content, excluding rapidly changing preprints, vendor demonstrations, and posters whose results cannot be adequately assessed; and (ii)~it anchors deployment-readiness claims to bibliographically stable, reproducibly retrievable records. Using a single curated index also improves search reproducibility and keeps the corpus definition transparent. Scopus was selected for its curated indexing of the engineering and building-science venues in which HVAC research appears. Its query interface is date-stamped, so searches can be re-run verbatim. Google Scholar was not used as a primary source because its results are neither stable nor exactly reproducible and it does not separate peer-reviewed from grey literature. The principal trade-off is that much current LLM engineering practice is documented only in arXiv preprints and GitHub repositories. The corpus therefore captures the peer-reviewed evidence base rather than the fastest-moving capability frontier. This is acknowledged as a scope limitation in Section~\ref{sec:limitations}.

\begin{figure*}[t!]
\centering
\resizebox{0.99\textwidth}{!}{\fontsize{6pt}{6pt}\selectfont{
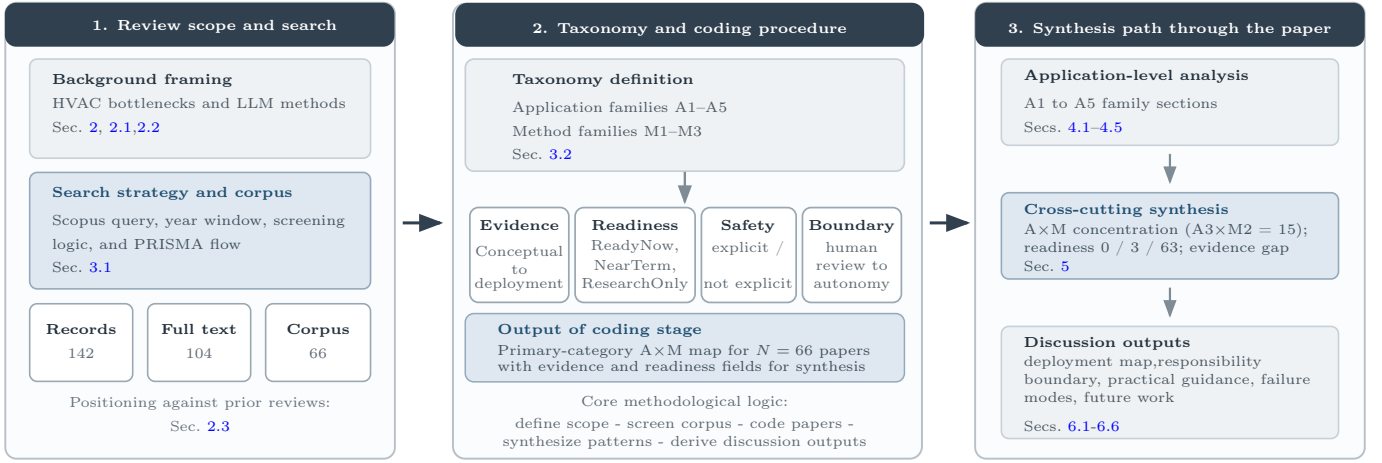
}}
\caption{Review workflow, from search and screening to taxonomy coding and cross-cutting synthesis.}\label{fig:workflow}
\end{figure*}

\begin{figure}[!h]
  \centering
  \includegraphics[width=0.95\linewidth]{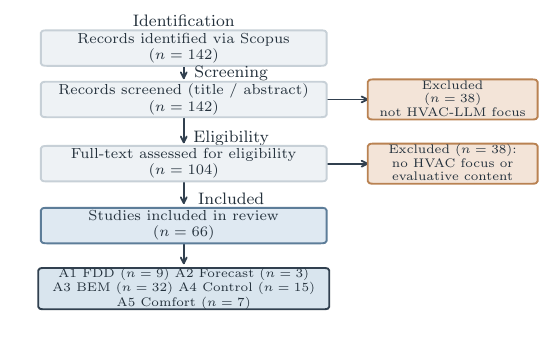}
\caption{PRISMA flow diagram: records identified via Scopus,
  screened, assessed for eligibility, and included in the review.}\label{fig:prisma}
\end{figure}

The search was restricted to journal articles, conference papers, and review articles published from 2023 onwards and indexed in Scopus by 31~March 2026, the day before the search was executed on 1~April 2026. The 2023 start follows the public release of ChatGPT on 30~November 2022, which made general-purpose LLMs broadly accessible; 2023 is thus the first full year in which HVAC-oriented LLM studies could appear, and the corpus reflects this with one 2023 paper against 10 in 2024 and 45 in 2025. The 2026 records therefore cover only the first quarter of that year. Review and perspective articles retrieved by the query were used for positioning and citation checking but were excluded from the primary-study coding corpus, which contains only studies reporting their own evaluative evidence. The initial query returned 142 records. After title/abstract screening, 38 records were excluded for lacking an HVAC-LLM focus, leaving 104 records assessed in full text. Of these, a further 38 were excluded for lacking a substantive HVAC operations focus or evaluative content.\footnote{The identical exclusion count (38) at the title/abstract and full-text stages is coincidental.} The final included corpus comprises 66 primary studies, distributed across five application categories: A1 ($n=9$), A2 ($n=3$), A3 ($n=32$), A4 ($n=15$), and A5 ($n=7$). The overall review workflow is illustrated in Fig.~\ref{fig:workflow}. The Preferred Reporting Items for Systematic Reviews and Meta-Analyses (PRISMA) flow is shown in Fig.~\ref{fig:prisma}.

Borderline papers were retained only when the full text contained a concrete HVAC operational task, system, or dataset together with some form of evaluative evidence. Studies addressing load forecasting were included when they evaluated the prediction of future HVAC or building-energy demand for an explicit operational purpose; studies limited to generating synthetic load profiles or simulation inputs without evaluating a forecasting task were excluded. Further exclusions comprised: (i)~LLM-adjacent building energy simulation work without an operational HVAC task; (ii)~Natural language processing or text-mining studies applied to building-energy literature or policy documents rather than operational HVAC data; (iii)~non-HVAC building applications such as structural monitoring, cost estimation, or general BIM processing; and (iv)~broad smart-building concept papers, generic AECO studies, or papers mentioning HVAC only peripherally. The review therefore focuses on studies in which an LLM directly participates in an HVAC operational workflow (fault detection, load forecasting, energy modelling, control, or occupant interaction) and is evaluated using empirical, simulated, laboratory, or field evidence.

\subsection{Taxonomy and Coding}\label{sec:method_coding}
\textit{Application categories.} Each paper was assigned one primary application category according to its dominant HVAC task, following the A1--A5 definitions introduced in Section~\ref{sec:intro}. Papers with substantial contributions to a second task were additionally assigned a secondary application category. All quantitative category counts and A$\times$M cross-tabulations are based on the primary assignment.

\textit{Method-category coding.} Each paper was assigned one method category according to the M1--M3 definitions in Section~\ref{sec:background_methods} and Table~\ref{tab:method_categories}. Assignment followed the dominant integration logic of the system. Agentic multi-step workflows were coded as M2, as were systems in which solver, optimiser, or control-layer outputs feed back into the model's run-time workflow, since that coupling is an orchestration pattern rather than an input modality. One-way execution of a generated artefact by a downstream solver remained M1 (Section~\ref{sec:background_methods}). Only systems that ingest non-textual inputs through a vision--language encoder were coded as M3.

\textit{Evidence and deployment coding.}
Each paper was coded for evidence type, deployment readiness, and responsibility boundary. The presence of an explicit safety discussion was recorded narratively rather than as a separate coded field. The operational definitions used for these dimensions are summarised in Table~\ref{tab:deployment_codebook}.

\begin{table*}[t]\footnotesize
\centering
\caption{Operational definitions for evidence and deployment coding.}
\label{tab:deployment_codebook}
\renewcommand{\arraystretch}{1.12}
\begin{tabularx}{\textwidth}{
  >{\raggedright\arraybackslash}p{1.6cm}
  >{\raggedright\arraybackslash}p{3.7cm}
  >{\raggedright\arraybackslash}X}
\toprule
Dimension & Code & Operational definition \\
\midrule

Evidence type
& Conceptual
& No empirical evaluation. \\

& Simulation-only
& Evaluation based exclusively on a simulated building, system, environment,
or synthetic dataset. \\

& Lab study
& Evaluation in a structured experimental or user-evaluation setting on
purpose-collected data, including hardware or testbed experiments and participant
or expert evaluations. The LLM is not deployed as part of routine building
operation. \\

& Retrospective field data
& Offline analysis of historical data collected from an operational building.
The LLM is not deployed in the live workflow. \\

& Pilot deployment
& Short-term deployment in a real building with measured operational or user
outcomes. \\

& Operational deployment
& Sustained or repeated use under routine operational conditions. \\

\addlinespace
Deployment readiness
& Ready-now
& A bounded and reusable workflow has been validated through sustained or
repeated operational deployment under human oversight, without granting the
LLM autonomous authority over safety-critical HVAC actions. \\

& Near-term
& Feasibility has been demonstrated for the intended operational role, but
additional field validation, latency or cost optimisation, integration work,
or governance clarification is required. Interface-usability evidence alone
does not qualify. \\

& Research-only
& Evidence remains conceptual, simulation-based, or otherwise insufficient
for operational use, or a fundamental gap remains between the demonstrated
capability and the reliability, safety, or validation requirements of the
intended role. \\

\addlinespace
Responsibility boundary
& LLM advisory--human decides
& The LLM provides recommendations, but a human operator makes and executes
the decision. \\

& LLM generates--human reviews
& The LLM produces an artefact, such as code, a report, or a parameter set,
that is reviewed before execution. \\

& LLM autonomous--low risk
& The LLM directly issues commands in a bounded, low-stakes setting without
prior human approval. \\

& LLM autonomous--high risk
& The LLM directly issues commands with safety-critical equipment or occupant
implications without human approval. \\

\addlinespace
Safety reporting
& Explicit / not explicit
& The paper explicitly discusses safety constraints, fallback behaviour,
responsibility boundaries, governance, or relevant failure modes. \\

\bottomrule
\end{tabularx}
\end{table*}

Only pilot and operational deployments were treated as field evidence.
Because the coding was performed by one reviewer, explicit operational
definitions and publication of the paper-level assignments were used to
improve auditability. In the remaining text, the codebook labels are written
in readable prose form.
\textit{Study-quality appraisal.} In addition to the five main coding dimensions, each paper is assessed against five quality criteria that together distinguish weak from strong deployment evidence. These criteria are not used as gatekeepers for inclusion; the inclusion decision follows the eligibility rules in Section~\ref{sec:method_search}. They are instead an interpretation aid: studies with limited data realism, absent closed-loop validation, or missing safety and reproducibility reporting may still be relevant to the field, yet they should not be read as evidence of deployment maturity. The rubric thus helps readers calibrate how much weight to place on individual studies when assessing the deployment-readiness and evaluation-realism claims reported in Sections~\ref{sec:applications} and~\ref{sec:crosscutting}. Two of the five criteria are derived directly from the evidence-type code. Data realism counts papers coded as retrospective field data, lab study, pilot deployment, or operational deployment, while experimental or field realism counts papers coded as lab study, pilot deployment, or operational deployment. The remaining three are assessed against the thresholds in Table~\ref{tab:quality_rubric} and reported qualitatively.

\begin{table}[!h]\footnotesize
\centering
\caption{Study-quality appraisal rubric applied to all 66 corpus papers.}\label{tab:quality_rubric}
\begin{tabularx}{\linewidth}{>{\raggedright\arraybackslash}p{1.5cm} >{\raggedright\arraybackslash}X >{\raggedright\arraybackslash}p{2.2cm}}
\toprule
Criterion & Definition & Threshold for ``met'' \\
\midrule
Data realism &
  Evaluation uses empirical data or real-world source material rather than
  exclusively simulated or synthetically generated inputs &
  Retrospective field, lab, pilot, or operational evidence \\
Experimental or field realism &
  Evaluation extends beyond simulation-only or retrospective analysis to a
  structured experimental setting, pilot installation, or operational deployment &
  Lab, pilot, or operational evidence \\
Closed-loop validation &
  LLM output directly influences a physical or simulated control action and the response is measured &
  Closed-loop evaluation reported \\
Reproducibility &
  Code, dataset, or model weights are publicly available, or results are replicated across multiple runs &
  Public artefact or multi-run results \\
Safety / governance reporting &
  Paper explicitly discusses responsibility boundary, safety constraints, data governance, or failure modes &
  At least one of these dimensions addressed \\
\bottomrule
\end{tabularx}
\end{table}

\section{Application-Level Analysis}\label{sec:applications}
The following five subsections analyse each application category in turn.
Each subsection follows a consistent internal structure: scope and corpus,
methodological patterns, evidence realism, and deployment barriers, which enables
systematic comparison across categories and directly supports the cross-cutting
synthesis in Section~\ref{sec:crosscutting}.

\subsection{Fault Detection and Diagnosis (A1)}\label{sec:a1}

\subsubsection{Scope and Corpus}
The A1 corpus comprises nine primary-coded papers. AHU, chillers, and chilled-water systems dominate the scope. Heat pumps are a leading electrified heating technology and an important source of demand flexibility in buildings \cite{IEA_2023, Agora_2023}. However, within the reviewed corpus, no study addresses heat-pump fault detection, forecasting, or control using LLMs.
This is a structurally consequential gap for two reasons. First, heat pump fault signatures are physically distinct from the chiller and AHU patterns that dominate the corpus: defrost transients, variable-speed part-load behaviour, and reverse-cycle refrigerant-side faults manifest along sensor dimensions (suction superheat, discharge-pressure ratio, isentropic efficiency) that are largely absent from cooling-only fault libraries. Second, because existing LLM-based FDD knowledge bases and fine-tuning datasets are built around ASHRAE RP-1312 chiller and AHU benchmarks, applying them to heat pumps risks plausible but physically wrong fault explanations (failure mode FM4; see Section~\ref{sec:discussion_failures}). Whether LLM-based approaches add value over established heat-pump FDD methods is therefore an open question. Answering it would require a dedicated benchmark in which defrost-cycle labelling, part-load coefficient of performance trajectories, and refrigerant-state channels are first-class inputs.

Compared with the rest of the review, A1 is relatively well-grounded empirically: six papers use retrospective field data, two use lab or user-evaluation settings, and one is simulation-only (Table~\ref{tab:a1_summary}). This is stronger than the evidence base in A4 and less benchmark-centric than much of A3, but it still falls short of live deployment because no A1 paper reaches pilot or operational evidence level. One A1 paper broadens the category toward knowledge-graph-enabled facility-management integration rather than pure equipment-level diagnosis. This reinforces the point that current A1 work often blends diagnostic reasoning with data-integration support~\cite{ravandi_integration_2025}.

\subsubsection{Methodological Patterns}
Three methodological patterns emerge from the A1 corpus, each with distinct
strengths and limitations.

\emph{Knowledge scaffolding (RAG, graph, ontology)} reduces the LLM's exposure to raw numerical sensor noise by pre-filtering data through symbolic structures. \citet{deng_graph-constrained_2026} combine graph-constrained association rule mining with a knowledge-enhanced LLM: the GC-ARM framework achieves \qty{89.1}{\percent} accuracy (best of five tested LLMs, in the ChatGPT-4o with RAG configuration) on real office HVAC data while reducing the candidate rule space by more than \qty{95}{\percent}. \citet{chen_chiller_2026} deploy a Re2G (retrieve-rerank-generate) chiller expert system on industrial maintenance manuals, achieving a \qty{23.5}{\percent} improvement in retrieval relevancy and a \qty{31.2}{\percent} reduction in hallucinations versus a standard RAG baseline. However, these gains depend on the completeness and maintenance of the underlying manuals, ontologies, and rule bases.

\emph{Supervised fine-tuning} directly addresses the output instability of zero-shot LLMs on HVAC classification tasks. \citet{zhang_domain-specific_2025} apply LoRA fine-tuning with SMOTE (Synthetic Minority Over-sampling Technique) data augmentation to GPT-3.5, elevating FDD accuracy from \qty{29.5}{\percent} to near \qty{100}{\percent} on the ASHRAE RP-1312 benchmark and generalising to unseen Lawrence Berkeley National Laboratory (LBNL) AHU datasets at above \qty{98}{\percent} without additional fine-tuning; reaching comparable accuracy on VAV and chiller equipment, however, required separate fine-tuning for each class. \citet{langer_fault_2025} fine-tune DistilBERT on serialised Modelica simulation time series, achieving binary classification F1 scores of \qtyrange{82}{99}{\percent} without any feature engineering. These results challenge \emph{bigger is better} narratives: small fine-tuned models match or exceed larger zero-shot ones on well-specified tasks, but the contribution narrows once the task is well-defined and the label vocabulary is fixed.

\emph{Agentic workflows} extend LLMs beyond classification into multi-step orchestration of virtual model development, diagnosis, and calibration. \citet{li_ai_2025} demonstrate a GPT-4o agent that integrates a building digital twin with predefined Python toolkits, reducing digital twin root mean square error (RMSE) from \qty{1.04}{\degreeCelsius} to \qty{0.30}{\degreeCelsius} after calibrating an introduced sensor bias on a real lab HVAC system. \citet{qiu_coding-free_2025} show that an LLM-guided ML pipeline can develop a coding-free virtual flowmeter for chilled water systems with RMSE in the range \qtyrange{8.84}{9.05}{\cubic\metre\per\hour} (using the recommended configuration) through conversational interactions alone. The practical risk in agentic designs is that chaining LLM-mediated tool calls can add latency, API cost, and downstream error-propagation points---costs that are rarely quantified; \citet{zhang_automated_2024} report \$17.68 per analysis run as one of the few direct cost figures in the corpus.

\subsubsection{Evidence Realism}
The A1 corpus is comparatively well grounded in retrospective field data and controlled user-facing evaluations. Graph-constrained diagnostic reasoners and workflow-automation pipelines have been evaluated on real HVAC operational data, including large-office HVAC records and one-year chiller-plant data~\cite{deng_graph-constrained_2026,zhang_automated_2024,li_ai_2025,qiu_coding-free_2025} (the virtual-model calibration study of \citet{li_ai_2025} is primary-coded A3 and is cross-referenced here for its real-data grounding); related maintenance copilots have been evaluated against technical manuals and user ratings~\cite{chen_chiller_2026}. However, this stronger empirical grounding should not be mistaken for live deployment maturity. The available evaluations also rely on comparatively curated knowledge sources, including structured manuals, rule bases, ontologies, and annotated operational datasets. They therefore provide limited evidence that the reported performance would persist across building portfolios where documentation is incomplete, inconsistent, or outdated and where the supporting knowledge artefacts require continuous maintenance. A key safety nuance identified by \citet{zhang_domain-specific_2025} must still be named explicitly: near-perfect benchmark accuracy on a fine-tuned model does not guarantee reliable performance on unseen equipment types. The fine-tuned GPT-3.5 achieves near-perfect accuracy on the RP-1312 benchmark but requires separate fine-tuning for each new equipment class. The problem extends beyond fine-tuned models: even in zero-shot and few-shot settings, high headline accuracy on a specific benchmark is insufficient as a proxy for deployment readiness.

\subsubsection{Deployment Barriers}
Table~\ref{tab:a1_summary} summarises the A1 corpus. The binding constraints, expressed as the recurring failure modes (FM) catalogued in Section~\ref{sec:discussion_failures}, are: reasoning inconsistency that coexists with strong accuracy metrics (FM1); agentic pipeline latency of seconds to minutes, compatible with offline auditing but not real-time monitoring (FM5); and knowledge-base dependency: maintenance copilots and graph-based reasoners fail when the supporting ontology or manual set is incomplete or outdated (FM2). These barriers recur across A$\times$M cells and are synthesised in Section~\ref{sec:crosscutting_barriers}.

\subsection{Energy Load Forecasting (A2)}\label{sec:a2}
With only three primary-coded papers ($n=3$), A2 is too small for robust comparative conclusions and should be read as a boundary condition rather than a mapped subfield. All three evaluate on retrospective field data, which improves realism over simulation, but none tests a live forecasting pipeline. The evidence base is too thin to classify as anything other than research-only.

The shared finding is that LLMs function as indirect contributors (code generators, feature adapters, or descriptor layers) rather than direct forecasters. \citet{zhang_data-driven_2025} orchestrate a Bayesian-optimised GPT pipeline for two real buildings (R\textsuperscript{2}~=~0.95, \$0.318 for training and deploying one model); \citet{zhu_enhanced_2025} use a patch-reprogramming module to adapt time-series inputs for a frozen BERT-base backbone; and \citet{liu_towards_2025} generate metadata-derived descriptions to condition a diffusion model. This pattern sidesteps rather than solves the core semantic-to-physical gap. Deployment is further constrained by the difficulty of detecting semantic errors in generated code~\cite{zhang_data-driven_2025}. A strong non-LLM baseline compounds this limitation: purpose-built time-series models such as PatchTST and TiDE outperform backbone-repurposed LLMs on heating load benchmarks at substantially lower computational cost~\cite{neubauer_forecasting_2026}. Table~\ref{tab:a2_summary} lists the three papers.

Why LLMs may struggle structurally as direct forecasters is instructive. LLMs are not natively designed for continuous multivariate time-series inputs, so Time-LLM-style approaches require patching and cross-modal reprogramming to map numerical HVAC signals into the language model’s embedding space~\cite{zhu_enhanced_2025}. This additional representation layer can introduce redundant information and delayed responses to abrupt local changes, although its limitations are task- and horizon-dependent. Compared with specialised forecasting architectures, LLM-based approaches do not consistently provide an accuracy advantage and can incur higher computational costs~\cite{zhang_data-driven_2025}. The credible A2 role is therefore more likely indirect, including pipeline-code generation, feature documentation, or metadata conditioning, rather than replacing specialised numerical forecasters.

\subsection{Building Energy Modelling and Simulation (A3)}\label{sec:a3}
\subsubsection{Scope and Corpus}
With 32 papers, A3 is the largest application category by a substantial margin. The dominant cell in the full A$\times$M grid is A3$\times$M2 (\num{15} papers), but M1 is close behind with \num{14} papers, making A3 the most methodologically mixed category in the review.

The category spans five analytically distinct sub-clusters that follow the building-model lifecycle: \emph{NL-to-model generation} (NL = natural-language), \emph{retrofit and advisory analysis}, \emph{operation and maintenance model updating}, \emph{semantic interoperability and metadata-grounding}, and \emph{calibration, digital-twin, and urban-scale workflows}. This distinction matters because the validation burden differs by use case. NL-to-model generation for new design can tolerate exploratory alternatives under human modeller review. Retrofit analysis affects investment decisions and therefore requires physically meaningful calibration. O\&M model updating must reflect equipment replacements, control-sequence changes, and point-list drift over time. Semantic-interoperability workflows must preserve relationships across BAS, BIM, and ontology layers, while urban-scale workflows face data-volume and consistency constraints beyond those of single-building models. The A3 corpus also includes digital-twin, metadata, and simulation extensions \cite{choi_gpt-ubem_2024,choi_i-udt_2025,saif_metadata_2026,song_ontology-assisted_2024}, alongside related schema work outside the coded corpus~\cite{lee_metadata_2025}, retrofit and advisory studies \cite{chen_can_2025,ghani_conversational_2025,hidalgo-betanzos_can_2025,jurisevic_large_2025,shu_large_2025,rysanek_how_2023}, and newer model-generation, design-optimisation, and BEM-learning directions \cite{labib_leveraging_2025,liu_large_2026,shao_enhancing_2026,zhong_energai_2025}.

\subsubsection{Methodological Patterns}
\emph{NL-to-model generation} (NL-to-IDF: NL to EnergyPlus IDF) is a major A3 sub-cluster. \citet{jiang_eplus-llm_2024} and \citet{jiang_efficient_2025} report near-perfect EnergyPlus/IDF generation on bounded NL-to-model benchmarks, while \citet{jiang_prompt_2025} show that prompt-only approaches can generate valid IDF content but degrade on more complex cases. \citet{lu_automated_2025} and \citet{zhang_automatic_2025} extend this pattern into multi-agent workflow automation for model generation, calibration, or debugging. Across the cluster, the main reported value proposition is productivity, including reduced authoring time, improved executability, and higher syntax or model-generation success.

\emph{Retrofit and advisory analysis} constitutes a second sub-cluster in which LLMs serve as decision-support tools rather than model generators. Studies in this group apply prompting or context-grounded orchestration to assist practitioners with retrofit option screening, energy audit interpretation, and code-compliance queries~\cite{chen_can_2025,ghani_conversational_2025,hidalgo-betanzos_can_2025,jurisevic_large_2025,shu_large_2025,rysanek_how_2023}. This sub-cluster occupies a practically relevant intermediate position on the deployment gradient: operating in advisory mode reduces the immediate risk of model errors because outputs are reviewed by an engineer rather than executed directly, but current evaluations rarely validate recommendations against actual energy outcomes or real retrofit decisions.

\emph{Operation and maintenance model updating} is an emerging A3 role rather than a mature subfield. In real buildings, energy models decay after retrofits, sensor replacements, setpoint changes, and undocumented control-sequence edits. LLMs are potentially useful here because the evidence needed to update the model is distributed across work orders, commissioning notes, BAS trend logs, manuals, and operator comments. Conventional calibration algorithms can tune parameters once the model structure is known, but they do not by themselves recover missing narrative context about what changed in the building. Early LLM-agent prototypes begin to address this by recalibrating virtual models and sensors from BAS data, metadata, and design documents~\cite{li_ai_2025,li_ai_2025-1,saif_metadata_2026}. Current evidence remains thin, so this role should be treated as a future-facing extension of BEM workflow assistance rather than a validated deployment category.

\emph{Semantic interoperability and metadata-grounding} form a distinctive subcluster in the A3 corpus. \citet{ababsa_continuity_2025} report strong Brick tagging performance across four building datasets. \citet{jia_natural_2025} demonstrate a proof-of-concept natural-language interface for querying building sensor data through an LLM-integrated protocol, and \citet{li_rag_2026} use a RAG pipeline to improve coverage and structural preservation when transforming heterogeneous building documents into AI-ready representations for semantic data modelling. \citet{grauer_chatgpt_2025} extend this metadata-grounding direction multimodally, prompting foundation models with Brick schemas, textual spatial context, and floor-plan images to construct the graph structure of graph neural networks for indoor-climate prediction; the evaluation remains simulation-only, with manually built ground-truth graphs as the reference.

\emph{Calibration, digital-twin, and urban-scale workflows} are also well represented in A3, although the corpus codes them conservatively on readiness. \citet{li_ai_2025} demonstrate LLM-mediated virtual-model development and calibration support, while \citet{li_energyplus-mcp_2025}, \citet{zhang_large_2025}, and \citet{zhang_large_2026} provide tool infrastructure, reusable agent schemas, and RAG-based optimisation layers around simulation workflows. The critical risk across these A3 sub-clusters remains \emph{executable but wrong}: a workflow may run, compile, or return a valid query path while still encoding incorrect geometry, schedules, HVAC relationships, calibration assumptions, or semantic mappings. Urban-scale extensions~\cite{choi_gpt-ubem_2024,choi_i-udt_2025} push the frontier further: coordinating LLM-mediated modelling at city scale introduces data-volume and consistency challenges that substantially exceed those of single-building workflows, and these represent the most research-distant applications in A3.

\subsubsection{Evidence Realism}
Evidence realism in A3 is mixed. The corpus contains retrospective field studies, lab studies, simulation-only work, and two pilot studies. The codebook's \emph{conceptual} class is unpopulated because inclusion required evaluative evidence. Although this represents a broader evidence base than a purely synthetic NL-to-IDF benchmark literature, the reported evaluations still focus primarily on productivity, executability, tagging accuracy, and workflow speed. Evidence remains limited on whether generated models preserve physically and semantically correct geometry, schedules, HVAC topology, metadata relationships, and calibration assumptions across diverse buildings and portfolio-scale applications. Only a small minority of studies validate outputs against calibration criteria or through use in live buildings.

\subsubsection{Deployment Barriers}
The dominant A3 deployment risk is the semantic gap between executable and correct. First, syntactic validity does not guarantee semantic or physical fidelity. Generated IDF content may contain incorrect geometry, wall and window coordinates, or schedule values, particularly as model complexity increases~\cite{jiang_prompt_2025}. More broadly, generated models and data-grounding workflows may preserve incorrect topology, calibration assumptions, or metadata relationships unless their outputs are independently validated. Second, many workflows assume unusually clean documentation and manageable token budgets; real portfolios instead contain incomplete, inconsistent, multilingual, and image-heavy building records. Third, tool exposure or schema release alone does not guarantee readiness: even open or modular infrastructure requires end-to-end deployment validation before it can be treated as field-ready.

Table~\ref{tab:a3_summary} presents selected A3 papers; results include time reductions and coverage metrics discussed below.

\subsection{HVAC Control and Optimisation (A4)}\label{sec:a4}

\subsubsection{Scope and Corpus}
A4 is the highest-safety-criticality application category and the only one outside A3 with confirmed near-term evidence. Fifteen primary-coded papers span 2024--2026; two of them (both Office-in-the-Loop studies~\cite{sawada_office---loop_2024,sawada_office---loop_2025}) are near-term based on their real-office deployment (two periods spanning approximately 7.5 weeks) with real occupants and measured outcomes. The remaining 13 are research-only. Beyond direct setpoint generation, the category includes zero-shot scheduling~\cite{liu_llm-bem_2025}, explanation-centred policy support~\cite{zhang_building_2025} and Shapley-grounded explanation of machine-learning control signals~\cite{zhang_large_2024}, generalisation-oriented building energy management system  (BEMS) control frameworks~\cite{zhang_leveraging_2025}, physics-informed advisory wrappers around MPC~\cite{liang_physics-informed_2025}, and battery-health-aware scheduling support that couples LLM-based state-of-health diagnosis with charging-strategy recommendation~\cite{yang_llm-bas_2025}.

\subsubsection{Methodological Patterns}
\emph{Advisory, explanatory, and design-support roles} are the most credible part of the A4 design space. \citet{ko_darlin_2025}'s DARLIN system uses RAG over HVAC domain knowledge to generate cooling setpoints with human-readable rationales, applied directly in its simulated control loop (hence coded autonomous--low-risk), but bounded to a low-stakes comfort domain, while \citet{zhang_building_2025} frame explanation quality itself as the central contribution of an LLM layer over a BEMS. \citet{hu_autocontrol_2025} automate proportional–integral (PI) controller design from Brick models and validate the generated controllers in BOPTEST, and \citet{liu_co-llm_2026} use retrieval-augmented policy optimisation for a tertiary-hospital chiller system, reporting a \qty{1.9}{\percent} energy-saving gain over an expert policy. Across these studies, the more credible A4 pattern is not unconstrained autonomous control, but bounded LLM support for translating system context, explaining control logic, retrieving domain knowledge, or proposing actions that remain subject to simulation, optimisation, or human review.

\emph{Direct or quasi-direct control claims} still dominate the category numerically, but almost all remain research-only. Simulation-based control papers such as \citet{jin_democratizing_2024}, \citet{li_llm-assisted_2025}, and \citet{zhu_heating_2025} report promising performance under constrained conditions, yet they do not validate behaviour under real disturbances, network latency, operator intervention, or equipment faults. A4 therefore remains the most safety-critical and most simulation-dominated application category in this review, with 12 of 15 papers evaluated in simulation only and 13 of 15 coded research-only.

\emph{Real-building evidence} is concentrated in the Office-in-the-Loop programme~\cite{sawada_office---loop_2024,sawada_office---loop_2025} (the two near-term A4 papers) reporting \qty{47.9}{\percent} energy savings (LLM-only condition) over two deployment periods spanning approximately 7.5 weeks in a real office. Both remain human-in-the-loop, not autonomous safety-critical control. \citet{liang_physics-informed_2025} add a physics-informed advisory MPC wrapper evaluated at \$0.013--0.032 per tuning call.

\subsubsection{Evidence Realism}
A4 evidence remains dominated by simulation: 12 of the 15 primary-coded papers are simulation-only, two are near-term pilots (Office-in-the-Loop), and one is evaluated on retrospective field data from a real commercial HVAC system. \emph{Control} in the current A4 literature therefore mostly means simulated decision generation, not independently replicated live HVAC actuation.

\subsubsection{Deployment Barriers}\label{sec:A4_DB}
Safety remains the binding constraint for A4. High-frequency setpoint
generation requires physics guardrails, because prompt engineering alone cannot
enforce comfort, setpoint, and equipment-capacity constraints~\cite{zhu_heating_2025,jin_democratizing_2024}. Latency is a
second barrier: agentic, multi-step LLM designs add seconds of inference per
decision; the Co-LLM chiller-control framework reports approximately
\qty{2}{\second} per control step, comparable to model-based control~\cite{liu_co-llm_2026}. This
suits slow supervisory loops such as chiller sequencing (minutes), but remains
incompatible with the sub-second cycles of fast HVAC control. Structural integration adds a third layer of difficulty:
commercial BAS and legacy building infrastructure remain heterogeneous and poorly interoperable, so connecting a general-purpose LLM typically requires custom middleware~\cite{amangeldy_review_2025}. Finally, there is insufficient evidence to support deployment claims outside the Office-in-the-Loop programme. A4 systems are mainly validated in simulation or offline analysis. The hospital chiller case study is described by its authors as a field study, but the methods section reports evaluation in a simulation environment calibrated to the plant. The most credible near-term path is therefore an M2 advisory layer in which the LLM generates setpoint recommendations subject to physics-model approval, rather than issuing commands autonomously~\cite{liang_physics-informed_2025}.

Table~\ref{tab:a4_summary} summarises selected A4 papers across the spectrum from research-only simulation studies to the near-term real-office deployments.

\subsection{Thermal Comfort and Occupant Interaction (A5)}\label{sec:a5}

\subsubsection{Scope and Corpus}
The A5 corpus comprises seven primary-coded papers. Four are coded M1 and three M2, so prompt-centred, fine-tuned, and context-grounded comfort and feedback models are all represented, while multimodal work (M3) is absent from this family. The strongest real-building control evidence in this area sits in the secondary-coded Office-in-the-Loop studies discussed in Section~\ref{sec:a4}. The primary A5 corpus leans toward advisory interfaces, feedback mining, and behavioural modelling rather than direct comfort-control deployment.
It also includes a broader occupant-facing energy advisory application~\cite{kanayo_ai_2024}, which helps explain why the category spans comfort-specific studies and general occupant-interaction concepts. Table~\ref{tab:a5_summary} (Appendix) gives an overview of the primary-coded A5 papers.

\subsubsection{Methodological Patterns}
\emph{Conversational and context-grounded interfaces} enable occupants and building
analysts to interact with comfort and indoor environmental quality (IEQ) systems in natural language. \citet{arslan_monitoring_2025} develop ThermalComfortBot, an M2 context-grounded system for office IEQ monitoring that achieves \qty{94}{\percent} accuracy and \qty{92}{\percent} precision, outperforming a retrieval-only baseline (\qty{82}{\percent} accuracy) by integrating BIM, sensor, and contextual data through RAG-grounded orchestration. \citet{liu_integrating_2025} show that natural-language comfort requests can be routed to regression or RL models (regression fit: $r$~=~0.95), although the system is not yet integrated with live sensors.

\emph{Feedback mining and preference learning} apply LLMs to large-scale occupant feedback and behavioural data. \citet{sadick_what_2025} fine-tune IEQ-BERT on \num{14622} labelled entries from online reviews and social media, achieving \qty{93}{\percent} accuracy and F1~=~0.93 across five IEQ domains.
\citet{qaisar_dynamic_2025} show that few-shot prompting can classify occupancy
presence with high binary accuracy on office data, and
\citet{liu_conversational_2026} move toward conversational preference learning.
The key point is that A5 contains a genuine prompting- and fine-tuning-based (M1) subcluster concerned with behavioural inference and personalised preference signals, even though none of these papers yet demonstrates longitudinal deployment.

\emph{Digital occupants and comfort proxies} remain the main research-only
frontier of A5. \citet{sasaki_llm-driven_2025} use LLM agents as digital
occupants in an EnergyPlus co-simulation (primary-coded A4 and cross-referenced here, so it does not appear in Table~\ref{tab:a5_summary}), while \citet{liu_human---loop_2026}
extend this line toward human-in-the-loop virtual occupant simulation and
semantic decision support. These studies are intellectually important, but
they still depend on uncalibrated assumptions about how well synthetic
language behaviour captures real human thermal preferences.

\emph{Cross-over field evidence.} The strongest real-building comfort result in
the broader corpus remains the Office-in-the-Loop line of work discussed in
Section~\ref{sec:a4}~\cite{sawada_office---loop_2024,sawada_office---loop_2025}.
These papers are primary-coded A4 with a secondary A5 designation, which
better reflects their central contribution: comfort is important, but it is
embedded in a broader control-and-optimisation workflow.

\subsubsection{Evidence Realism}
Four of the seven primary-coded A5 papers use retrospective field data,
with no pilot study; two are lab studies, and one is simulation-only. Even so, the
category still lacks sustained deployment realism. No primary-coded A5 paper
reports a longitudinal study of preference drift or compliance decay, and most
results remain query-accuracy, classification, or simulated-behaviour
outcomes rather than evidence of long-term building use.

\subsubsection{Deployment Barriers}
The A5 deployment horizon is shaped by two interlocking constraints. \emph{Behavioural validity} is under-measured: stated willingness to adjust comfort preferences in questionnaire studies does not reliably predict observed behaviour in sustained real-building deployments, and no primary-coded A5 paper reports a longitudinal study covering seasonal variation or preference drift. \emph{Privacy and data governance} create hard procurement barriers: occupant presence detection, thermal imaging, and personalised preference logs are regulated data classes under the General Data Protection Regulation (GDPR) in European buildings and under analogous frameworks in many other jurisdictions. Two of the seven A5 papers rely on hosted commercial APIs (GPT-4o, Gemini-Pro) to process such signals, and a third uses a hosted GPT-3.5 endpoint, an architecture that is incompatible with strict data-residency requirements and would require on-premise or federated inference alternatives not yet validated in the HVAC context. The minimum evidence that would move an A5 comfort role from research-only toward near-term readiness is therefore a longitudinal deployment demonstrating preference stability across at least one full heating or cooling season, using a privacy-compatible inference architecture, a study design not yet present in the corpus.

\section{Cross-Cutting Methodological Analysis}\label{sec:crosscutting}
The preceding sections examined each application category in isolation. This section synthesises findings across all five categories through three complementary lenses (the A$\times$M distribution map, temporal and model trends, and the evidence realism gap), each of which informs the deployment-readiness assessment in Section~\ref{sec:discussion}.

\subsection{\texorpdfstring{The A1--A5 $\times$ M1--M3 Map}{The A1--A5 by M1--M3 Map}}\label{sec:axm_map}
\begin{figure}[h!]
  \centering
  \includegraphics[width=\linewidth]{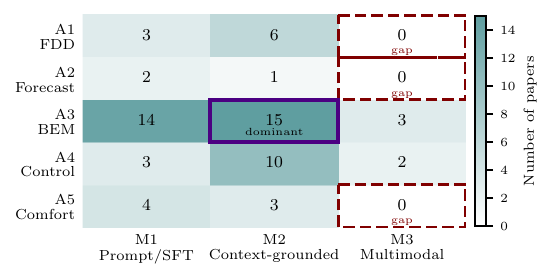}
  \caption{Application-by-method taxonomy (A1--A5 by M1--M3).
  Cell values show paper counts from the reviewed corpus. Violet border: dominant cell (A3/M2, n=15). Red border: gap cells.}
  \label{fig:axm_heatmap}
\end{figure}

Addressing \emph{RQ1}, the cross-tabulation (Fig.~\ref{fig:axm_heatmap}) reveals a strongly skewed concentration.
The dominant cell is A3$\times$M2 with 15 papers: context-grounded orchestration for building energy modelling constitutes the largest single cluster in the corpus. A3$\times$M1 is close behind with 14 papers. This dominance reflects a structural alignment between LLM strengths and BEM workflow characteristics: model generation is a multi-step, document-grounded task with clear success criteria (compilation, calibration); this makes it well-suited for orchestrated pipelines with tool use and iterative debugging.

\begin{figure*}[t!]
\centering
\includegraphics[width=\linewidth]{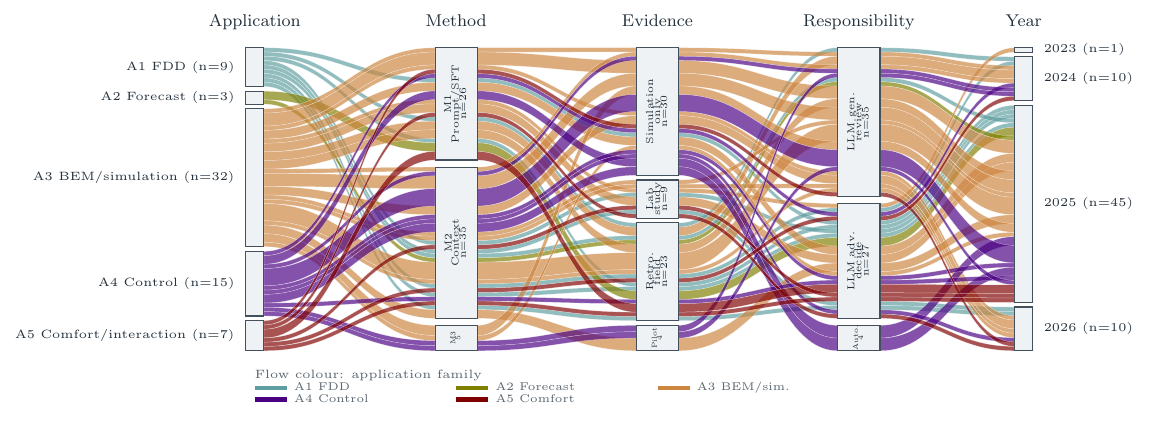}
\caption{Parallel-categories view of the reviewed corpus across application category, LLM method category, evidence type, responsibility boundary, and publication year ($N=66$). The figure complements the A$\times$M heatmap by showing how the main coding dimensions co-occur across papers. Flow colour indicates application category, allowing each A1--A5 ribbon group to be traced through method, evidence, responsibility boundary, and year. To read the figure, follow a ribbon left to right: each ribbon is a group of papers sharing the same category on every axis, and ribbon width is proportional to the number of papers in that group.}\label{fig:parallel_categories}
\end{figure*}

The sparse and empty cells indicate where little evidence is currently available, but they do not necessarily represent promising research gaps. A1$\times$M3, A2$\times$M3, and A5$\times$M3 contain no papers. For load forecasting in particular, this absence may reflect limited method fit, since specialised time-series models already process structured numerical inputs efficiently and multimodal LLMs add value only when complementary unstructured information contributes to prediction. A2$\times$M2 contains only one paper, while A4$\times$M3 contains two and A4$\times$M1 contains three. A similar argument applies to control: for well-posed tasks with structured inputs, established numerical, supervised, and optimisation-based methods remain more direct and verifiable than prompt-based approaches, which remain sensitive to prompt formulation~\cite{zhu_heating_2025,jin_democratizing_2024}. Thus, these control-oriented combinations remain sparsely studied rather than established subfields.

The A3 concentration reflects a structural fit between current LLM evaluation infrastructure and BEM tasks. \emph{Four asymmetries} explain the pattern. First, NL-to-code and NL-to-model tasks can be evaluated on static benchmarks with clean success criteria such as compilation rate or calibration accuracy, whereas A4 requires closed-loop real-building validation under safety constraints. Second, a \qty{100}{\percent} IDF compilation rate is a clean publishable result; a \qty{3}{\percent} simulation-only energy saving without field comparison is not. Third, BEM generation errors can be caught by engineers before use, while control errors affect occupants and equipment directly. Fourth, documentation and point lists are easier to share than operational time series, which are often restricted by facility data-governance constraints, incompletely archived, or corrupted by sensor faults and missing values. Together, these factors make A3 the natural entry point for academic LLM work, independent of where the largest operational value may lie.

The same asymmetry explains A4. Its simulation-heavy evidence base (12 of 15 papers) does not show that LLMs are close to operational control. It shows that accessible evaluation infrastructure for LLM-based controllers remains missing. Reinforcement-learning control, by contrast, already has standardised community-scale benchmark environments~\cite{nweye_citylearn_2025}. The Office-in-the-Loop programme, which provides the main A4 field evidence, required instrumented real offices, multi-month deployment, and direct occupant participation, resources most research groups cannot assemble. This does not mean BEM automation is more valuable than control; it means it is far easier to demonstrate.

Figure~\ref{fig:parallel_categories} complements the A$\times$M heatmap by showing how the reviewed papers distribute jointly across application category, LLM method category, evidence type, responsibility boundary, and publication year. Whereas Fig.~\ref{fig:axm_heatmap} shows the dominant cells of the taxonomy, the parallel-categories view makes the cross-dimensional structure of the corpus visible in a single synthesis figure.

\subsection{LLM Models and Temporal Trends}\label{sec:llm_trends}

\begin{figure}[h!]
  \centering
  \includegraphics[width=\linewidth]{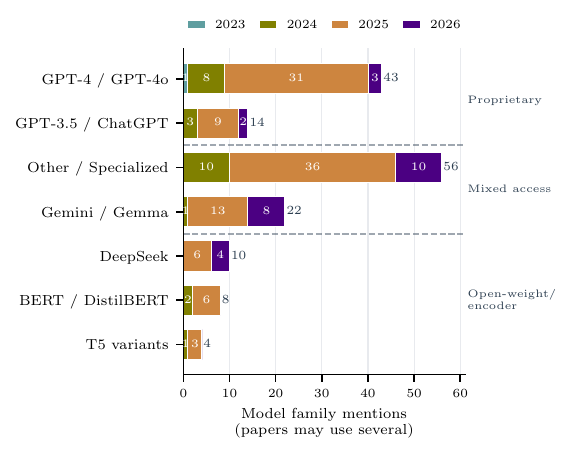}
  \caption{LLM model usage across the reviewed corpus, broken down by
  publication year. GPT-4/GPT-4o dominates; open-weight models
  appear primarily in fine-tuning studies.}
  \label{fig:llm_models}
\end{figure}

GPT-4 and GPT-4o form the largest single model group in the corpus (Fig.~\ref{fig:llm_models}), appearing across all five application categories and all publication years covered (2023--March~2026). This reflects the models' strong zero-shot and few-shot reasoning capability but also creates a systemic governance risk: the corpus is heavily dependent on a small proprietary-model ecosystem. A parallel concentration exists among authors: a few prolific groups (notably those of \citet{jiang_eplus-llm_2024,jiang_efficient_2025}, \citet{zhang_opportunities_2025,zhang_automatic_2025,zhang_large_2025}, and \citet{li_energyplus-mcp_2025,li_rag_2026,li_mcp_2026}) contribute a disproportionate share of the corpus, so both model- and group-level trends should be read with this dependence in mind. 

The corpus also includes open-weight or smaller encoder-style models (DistilBERT, T5 variants, BERT, and other models such as GPT-2 and LLaMA-family models), but these are concentrated in fine-tuning, representation-learning, or constrained benchmark studies rather than in
broadly validated deployment settings. The governance implication is direct:
building owners who require on-premise processing or strict data-residency
controls still face a shallow evidence base.

\begin{figure}[!h]
  \centering
  \includegraphics[width=\linewidth]{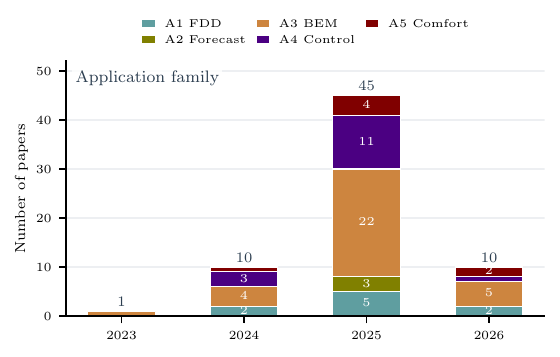}
  \vspace{0.6em}
  \includegraphics[width=\linewidth]{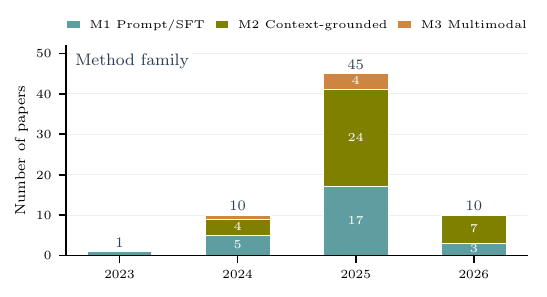}
  \caption{Papers per year 2023--2026 (March), stacked by application category
  A1--A5 (top) and method category M1--M3 (bottom). The field shows clear
  acceleration after 2024, while the method mix shifts toward
  context-grounded orchestration.}\label{fig:papers_per_year}
\end{figure}

The temporal distribution of papers shows clear acceleration after 2024 (Fig.~\ref{fig:papers_per_year}). The 2024--2025 window marks a qualitative inflection point, with foundational field-deployment evidence \cite{sawada_office---loop_2024} and new infrastructure contributions, including  MCP servers and agent schema libraries. The 2026 papers included through March suggest a maturing subfield that is moving from proof-of-concept studies toward more deployment-oriented evaluation. The method mix shifts in the same direction (Fig.~\ref{fig:papers_per_year}, bottom panel): context-grounded orchestration (M2) grows from 4 of 10 method codes in 2024 (\qty{40}{\percent}) to 24 of 45 in 2025 (\qty{53}{\percent}), overtaking prompting and supervised adaptation (M1) and becoming the most common method category across the full corpus (35 of 66 papers). The growing M2 share is consistent with the broader industry move from question-answering assistants toward task-oriented, tool-using systems. Because peer-reviewed publication trails fast-moving engineering practice, the corpus may underrepresent the most recent agentic designs (Section~\ref{sec:limitations}). The trend is steady and consistent in direction rather than abrupt, and the 2026 figures cover only the first quarter. Two caveats bound this trend. The annual cell counts are small, so the year-on-year shares should be read as indicative rather than precise. In addition, the M2 definition covers retrieval, tool use, memory, and agentic orchestration, and therefore absorbs by construction several system types that a narrower definition would separate; part of the apparent growth reflects that breadth.

\subsection{Evidence Realism Gap}

\begin{figure}[!h]
  \centering
  \includegraphics[width=\columnwidth]{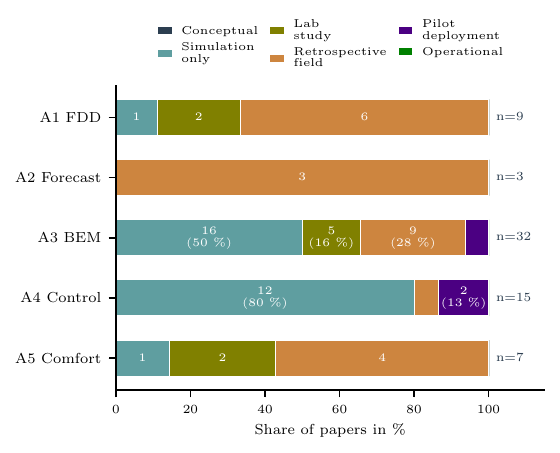}
  \caption{Evidence realism distribution per application category. Colours indicate evidence type from conceptual to operational deployment. The corpus shows mixed offline evidence in A1, A2, A3, and A5, but A4 remains strongly simulation-heavy; only four pilot studies and no operational deployments appear in the full corpus.}\label{fig:evidence_realism}
\end{figure}

Across all five application categories, the dominant evidence pattern is not
uniformly simulation-only but offline and pre-deployment (Fig.~\ref{fig:evidence_realism}). 

A1 combines
retrospective field, lab, and limited simulation evidence; A2 is entirely
retrospective field evaluation; A3 spans retrospective field, lab,
simulation-only, and pilot evidence; A4 remains strongly simulation-heavy; and
A5 mixes retrospective, lab, and simulation evidence. Field evidence in
the strict sense of pilot or operational deployment remains scarce: the corpus
contains only four pilot-coded papers and no operational deployments. These four
papers correspond to three distinct field deployments, as the two A4 pilots
report the same Office-in-the-Loop programme~\cite{sawada_office---loop_2024,sawada_office---loop_2025}. Pilot status and near-term readiness are not equivalent, however. The two
Office-in-the-Loop papers are the only pilots also coded near-term; the remaining
two pilot studies stay research-only because the demonstrated role is a
proof-of-concept interface rather than a reusable bounded artefact. The third
near-term paper is an A3 data-grounding workflow validated on operational
documents from real buildings but coded as a lab study.

This gap has a specific consequence for the deployment narrative. Papers
reporting strong R\textsuperscript{2}, high F1, or percentage energy savings
from simulation studies cannot be taken as evidence of field performance
without replication under real disturbances. The consistency problem
documented by current A1 studies---where repeated runs can produce different
outputs or plausible but incorrect reasoning---suggests
that single-run benchmark results systematically overestimate the reliability
of LLM-based systems.

Applying the study-quality rubric (Table~\ref{tab:quality_rubric}) across the
corpus quantifies this gap. \emph{Data realism} is met by 36 of 66 papers (empirical data or real-world source material rather than exclusively simulated or synthetic input) and is
strongest in A2 (3/3), A1 (8/9), and A5 (6/7), stands at 16/32 in A3, and is
weakest in A4 (3/15), the most simulation-bound category.  \emph{Experimental or field realism} is met by 13 of 66 papers, comprising nine lab studies and four pilot studies, and is absent in A2. This figure should not be interpreted as 13 studies validating an LLM on a physical HVAC system: most of the lab-coded studies are structured participant, expert, or other experimental evaluations rather than hardware or live-building tests. Physical-system validation is therefore substantially less common than the 13/66 figure alone suggests.

The remaining three criteria are weaker still and are not uniformly reported: closed-loop validation is confined to the four pilot studies and a minority of A4 simulations; reproducibility evidence (public artefacts or multi-run results) is limited because the corpus is dominated by proprietary models and single-run benchmarks (FM6); and explicit safety, governance, or failure-mode reporting is
uneven across studies. The rubric therefore confirms that the binding weakness is not the absence of task-performance metrics but the limited strength of \emph{deployment} evidence.

\subsection{Cross-Cutting Deployment Barriers}\label{sec:crosscutting_barriers}
Addressing \emph{RQ3}, five barriers recur across multiple A$\times$M cells and constrain deployment independently of any single application or method. Table~\ref{tab:barriers} organises these barriers by type, affected cells, consequence, and candidate mitigation. The table is derived from patterns visible across the corpus rather than from any single paper; it is intended to support RQ3 by translating
failure-mode analysis into cross-category structural constraints.

\begin{table*}[t]\footnotesize
\centering
\caption{Cross-cutting deployment barriers identified across the A$\times$M taxonomy.
Affected cells refer to primary A$\times$M combinations where the barrier
is most acute; it may appear less severely elsewhere. The table is derived from recurring patterns across the corpus rather than from any single paper.}\label{tab:barriers}
\begin{tabularx}{\textwidth}{p{2cm} p{5cm} p{1.6cm} p{3.2cm} X}
\toprule
Barrier type & Manifestation in corpus & Affected A$\times$M cells & Consequence & Candidate mitigation \\
\midrule
Semantic gap (executable vs.\ correct) &
  Generated models compile and produce plausible outputs while encoding incorrect geometry, schedules, or HVAC topology &
  A3$\times$M1, A3$\times$M2 &
  Executability is reported as validation; physical fidelity is not assessed &
  Physical fidelity checks beyond syntax; calibration against measured data \\
Temporal / control latency &
  Agentic systems report response times of seconds to minutes per cycle; HVAC loops require sub-second to low-second decisions &
  A4$\times$M2, A4$\times$M3 &
  Direct closed-loop LLM control infeasible; advisory roles only &
  Latency-constrained RAG designs; hard-coded physics-layer fallback controllers \\
Governance and data residency &
  Corpus relies heavily on proprietary cloud APIs; regulated occupant data (presence, occupant feedback text) cannot be sent externally &
  All cells, especially A5 &
  Deployment blocked in GDPR-constrained or on-premise environments &
  On-premise open-weight model validation; federated inference designs \\
Reproducibility (model dependence) &
  Results tied to API model versions that are updated silently; single-run benchmarks dominate &
  A1$\times$M2, A3$\times$M2, A4$\times$M2 &
  Single-run results overestimate reliability; cross-version replication fails &
  Open-weight replication; multi-run consistency reporting as standard \\
Safety / responsibility boundary &
  No formal constraint layer between LLM output and physical actuation in A4 prototypes &
  A4$\times$M2, A4$\times$M3 &
  High-autonomy deployment requires safety infrastructure absent from current designs &
  Constraint-enforcing wrappers; independent sensor watchdogs; hard-coded fallback controllers \\
Security / prompt injection &
  LLMs ingest manuals, work orders, and operator text; few designs validate input provenance or guard against adversarial or injected instructions &
  All cells, especially M2 &
  Untrusted ingested content can steer outputs or actions; BAS-connected systems expose an attack surface &
  Input-provenance checks; output validation; injection-resistant retrieval; model supply-chain trust \\
\bottomrule
\end{tabularx}
\end{table*}

\section{Discussion}\label{sec:discussion}
The cross-cutting analysis establishes the structural patterns in the corpus. This section translates that evidence base into four practical outputs: (i) a \emph{deployment-readiness assessment} (Section~\ref{sec:discussion_map}), (ii) \emph{safety-aware responsibility boundaries} (Section~\ref{sec:discussion_safety}), (iii) \emph{actionable practitioner guidance} (Section~\ref{sec:discussion_guidance}), and (iv) an identification of \emph{failure modes and future-work priorities} (Sections~\ref{sec:discussion_failures}--\ref{sec:discussion_future}). Taken together, these outputs are intended as a decision framework for where LLMs can create system-level energy value (through faster retrofit analysis, more scalable interoperability, and better human decision support) without prematurely crossing into unsafe autonomous control.

\emph{Why this matters for the energy transition.} The roles found defensible
here bear directly on the pace of the building energy transition, and the
mechanism is specific. Electrified heating, demand-flexible operation, and deep retrofits must be deployed across a large, heterogeneous, and poorly documented building stock, and that deployment is currently rate-limited less by control theory than by the per-building engineering effort needed to make each building legible: cleaning point names, reconstructing documentation, building and calibrating energy models, and integrating legacy systems. Metadata normalisation and document-grounded operator support reduce the integration labour that makes portfolio-scale analytics uneconomic; BEM workflow assistance lowers the cost of the retrofit analysis that underpins investment decisions; and advisory interfaces around MPC or RL let operators adopt advanced control they can interpret, override, and audit rather than reject as an opaque black box. The energy-transition contribution of LLMs is therefore primarily one of \emph{reduced integration and engineering friction and more interpretable and potentially more scalable adoption of established methods}: a more modest but more credible claim than that of autonomous energy-optimising agents, one in which the near-term payoff comes from removing the semantic and organisational barriers that have kept proven efficiency measures from scaling, not from handing control of physical plant to a language model.

\subsection{Deployment-Ready vs.\ Research-Only: The Core Map}

\label{sec:discussion_map}
Addressing \emph{RQ2}: no paper is classified as ready-now. That status would require sustained or repeated operational deployment of a reusable, bounded artefact under human oversight, and the corpus contains no such evidence (Section~\ref{sec:crosscutting}). Three papers are near-term and 63 are research-only.

The three near-term cases fall into two distinct tracks. The first is \emph{A3 infrastructure tooling}: \citet{li_rag_2026} deploy a RAG pipeline across 13 real buildings, achieving an approximately 30\% improvement in extraction coverage and structural preservation for heterogeneous documents (e.g.\ images and tables); this role is a bounded, human-supervised task with no safety-critical autonomy. \citet{li_energyplus-mcp_2025} provide a related open-source MCP server exposing EnergyPlus as a tool-callable interface, but as validation is limited to an example model rather than real building data, that role is classified research-only. The second track is \emph{A4 real-office control support}: \citet{sawada_office---loop_2024,sawada_office---loop_2025} report a closed-loop deployment in a real office building (two periods spanning approximately 7.5 weeks), achieving \qty{47.9}{\percent} energy savings (LLM-only condition) under multimodal LLM supervision; the 2025 study extends the same experiment with an offline agentic-reasoning reanalysis rather than a second deployment. This is the only A4 role with real-building evidence, and its near-term classification reflects the combination of empirical grounding and bounded decision scope.

The contrast with \citet{jia_natural_2025} illustrates the coding logic: readiness is not a monotonic function of evidence level. \citet{jia_natural_2025} reach pilot-level evidence but remain research-only because the demonstrated role is a proof-of-concept interface whose 4K-token context limit precludes operational use, whereas \citet{li_rag_2026} are coded near-term on lab-level evidence because the role is bounded, human-supervised, and validated on operational data from 13 real buildings. Deployment readiness therefore combines evidence realism with role boundedness and artefact reusability, rather than following the evidence hierarchy alone.

Every other role in the corpus (FDD reasoners, load forecasters, BEM generators, advisory control wrappers, and comfort interfaces) remains at the research-only stage, typically evaluated in simulation or on a single retrospective dataset without live deployment. A4 remains predominantly research-only (13 of 15 papers; 12 of 15 simulation-only). A1, the most empirically grounded category (6 of 9 papers use retrospective field data), still lacks live deployment evidence. Near-term readiness is therefore concentrated in infrastructure tooling and bounded office-scale control, not in autonomous diagnosis, load forecasting, or comfort operation. Table~\ref{tab:deployment_readiness} summarises representative roles by evidence basis and readiness classification.

\begin{table*}[t]\footnotesize
\centering
\caption{Deployment readiness of representative LLM roles in HVAC operations.}\label{tab:deployment_readiness}
\begin{tabularx}{\textwidth}{p{6.0cm} p{0.5cm} p{2.0cm} p{6cm} X}
\toprule
Role & Cat. & Deployment & Evidence basis & Representative papers \\
\midrule
EnergyPlus MCP server (tool infrastructure) & A3 & Research-only & Example-model retrofit demo; open-source artefact; no real-building validation & \cite{li_energyplus-mcp_2025} \\
RAG pipeline for heterogeneous building data & A3 & Near-term & 13 real buildings; ${\sim}$\qty{30}{\percent} extraction-coverage gain & \cite{li_rag_2026} \\
Office HVAC control via multimodal LLM & A4 & Near-term & Real office, two periods (${\approx}$7.5 weeks); \qty{47.9}{\percent} energy savings (LLM-only condition) & \cite{sawada_office---loop_2024,sawada_office---loop_2025} \\
Chiller FDD with knowledge graph & A1 & Research-only & Retrospective field data; \qty{89}{\percent} accuracy; single building & \cite{deng_graph-constrained_2026} \\
Chiller maintenance copilot & A1 & Research-only & Lab evaluation; real manuals & \cite{chen_chiller_2026} \\
NL-to-IDF automated modelling & A3 & Research-only & Simulation-only; \qty{100}{\percent} accuracy on benchmark & \cite{jiang_efficient_2025} \\
Brick / point-name tagging & A3 & Research-only & Retrospective field data; three real buildings + one synthetic & \cite{ababsa_continuity_2025} \\
Load prediction pipeline & A2 & Research-only & Retrospective field data; two real buildings; R$^2$=0.95 & \cite{zhang_data-driven_2025} \\
Advisory MPC wrapper & A4 & Research-only & Physics model trained on plant data; controller evaluated in one-week simulation; \$0.013--0.032 / call & \cite{liang_physics-informed_2025} \\
Comfort / IEQ advisory interface & A5 & Research-only & Retrospective field data; \qty{94}{\percent} accuracy & \cite{arslan_monitoring_2025} \\
Direct closed-loop LLM control & A4 & Research-only & Simulation-only; no safety validation & \cite{zhu_heating_2025,li_llm-assisted_2025} \\
Digital occupant proxy & A5 & Research-only & Simulation-only; no physiological validation & \cite{liu_human---loop_2026} \\
\bottomrule
\end{tabularx}
\end{table*}

Five bounded, low-risk workflow roles emerge as the most defensible areas for near-term practitioner attention: metadata normalisation and point-name cleanup, grounded document
and knowledge-base support, MPC/RL advisory interfaces, BEM workflow automation, and technician training and procedural support. These are broader role types rather than individual readiness classifications, and the strict near-term label is assigned only where the available evidence meets the operational criteria defined in Section~\ref{sec:method_coding}. Each role pairs a potential efficiency gain with a constraint that limits the LLM's authority: human validation of semantic mappings~\cite{zheng_mastering_2025,ababsa_continuity_2025,forth_semantic_2024}, retrieval-quality control~\cite{arslan_monitoring_2025,ko_darlin_2025}, explanation-fidelity checks~\cite{zhang_building_2025,chen_customized_2025}, engineer review of generated models~\cite{lu_automated_2025,jiang_efficient_2025}, and an explanatory rather than executive remit, respectively.

Four roles remain research-only. High-frequency closed-loop HVAC control without physics guardrails: no paper in the corpus has validated this at adequate safety margins outside simulation. Fully autonomous agentic operation: error propagation in multi-step chains is not bounded by current designs. Unvalidated synthetic occupant proxies used for policy deployment: LLM-agent preference simulations remain uncalibrated against real physiology. These classifications reflect current evidence, not permanent limits. As LLM capabilities advance and safety infrastructure (physics guardrails, formal constraint layers, error-recovery protocols) matures alongside field-validation evidence, these roles may become viable; the precondition is rigorous field testing, not model capability alone. Taken together, the current evidence supports LLMs primarily as bounded semantic and workflow-support tools, while roles involving autonomous physical decision-making remain at the research stage.

\subsection{Safety-Aware Responsibility Boundaries}\label{sec:discussion_safety}
Four responsibility-boundary levels, defined in
Section~\ref{sec:method_coding}, should be treated as explicit system-design
decisions rather than implementation defaults:
\begin{enumerate}
  \item \emph{LLM-advisory-human-decides}: the LLM generates a recommendation; a human operator decides and acts.
  \item \emph{LLM-generates-human-reviews}: the LLM generates an artefact (code, report, parameter set); a human reviews before execution.
  \item \emph{LLM-autonomous-low-risk}: the LLM issues commands with immediate effect in low-stakes domains (comfort setpoints; occupancy flags).
  \item \emph{LLM-autonomous-high-risk}: the LLM issues commands with direct safety implications (equipment sequencing, emergency ventilation, emergency boiler shutdown) without human oversight.
\end{enumerate}

Further addressing \emph{RQ2}, the responsibility-boundary audit defined in Section~\ref{sec:method_coding} reveals a consistent pattern: no paper in the corpus deploys an LLM in an autonomous high-risk role. The four papers coded \textit{LLM-autonomous-low-risk} (\citet{ko_darlin_2025}, \citet{li_llm-assisted_2025}, \citet{zhang_leveraging_2025}, and \citet{zhu_heating_2025}) are all confined to low-stakes comfort or advisory settings, and all acknowledge prediction fragility while stopping short of formal safety analyses.

\begin{figure}[!h]
  \centering
  \includegraphics[width=\columnwidth]{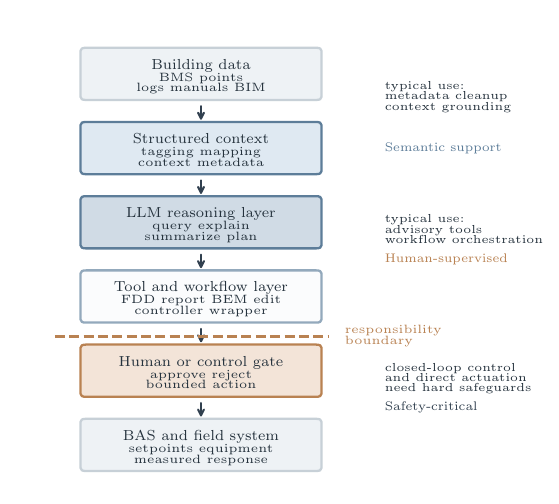}\caption{LLM--HVAC cyber-physical stack showing the responsibility boundary between semantic and workflow support and operational action. The labels describe architectural responsibility, not the deployment-readiness classifications assigned to corpus studies.}\label{fig:llm_hvac_stack}
\end{figure}

Figure~\ref{fig:llm_hvac_stack} maps the four responsibility boundary levels defined above onto the LLM--HVAC cyber-physical stack. 

The stack also includes data sources, a semantic layer, the LLM, tools and outputs, the decision layer, actuation, and the physical plant. The dashed responsibility boundary is the key divide, separating the LLM’s semantic role from the human oversight zone. No reviewed paper crosses this boundary autonomously for safety-critical actuation; all autonomous deployments in the corpus are confined to levels~1--3.

The absence of \textit{LLM-autonomous-high-risk} deployment reflects consistent field-wide restraint: researchers universally position LLMs in advisory or human-reviewed roles when physical actuation, equipment safety, or occupant health could be affected. This restraint is well-founded---\citet{zhang_domain-specific_2025} demonstrated that even a fine-tuned LLM achieving near-perfect fault-label accuracy may retain unreliable causal reasoning, illustrating that high classification performance alone does not justify removing human oversight from the actuation loop. Any future deployment elevating an LLM beyond an advisory role will require explicit safety layers: constraint-enforcing wrappers, hard-coded fallback controllers, and independent sensor watchdogs, none of which appear in current prototypes. A further absent dimension is \emph{security}: LLMs ingesting manuals, work orders, or operator text are exposed to prompt injection, so building-connected deployments must treat input provenance, output validation, and supply-chain trust as explicit safety controls (cf.\ IEC~62443).

\subsection{Where Language Reasoning Adds Value}\label{sec:discussion_language_reasoning}
Addressing \emph{RQ4}, the reviewed evidence suggests that LLMs are most useful when the limiting factor is not numerical prediction or control optimality, but the interpretation and coordination of heterogeneous information. HVAC tasks benefit from language reasoning when they require translating between human intent, building-specific documentation, point names, logs, manuals, code, and simulation tools. They benefit less when the task is already a well-posed numerical optimisation, time-series prediction \cite{babakhani_moe_2026,darbandi_lstm_2025,neubauer_multistep_2025,neubauer_fi_2024}, data imputation \cite{yu_tcn_2025,yu_imputation_2025} or ontology-mapping problem with clean inputs and stable labels. Table~\ref{tab:language_reasoning_fit} maps this boundary across six task types.

\begin{table*}[t]\footnotesize
\centering
\caption{Fit-for-purpose boundary between language reasoning and conventional HVAC methods. Apart from the task types, all entries state what current evidence supports rather than permanent capability limits, and may shift as tool-using agents mature.}\label{tab:language_reasoning_fit}
\begin{tabularx}{\textwidth}{p{3.6cm} X X X}
\toprule
Task type & Better default tool & Where LLMs add value & Deployment boundary \\
\midrule
Load forecasting &
  Time-series ML, transfer learning, probabilistic forecasting &
  Generating metadata descriptions, building pipeline code, explaining forecast drivers &
  Human-reviewed analytics; not autonomous scheduling evidence \\
Closed-loop control and optimisation &
  MPC, RL, rule-based supervisory control, safety filters &
  Explaining controller actions, translating operator intent, proposing slow-time-scale tuning candidates &
  LLM advisory only; physics layer decides \\
Point naming and metadata alignment &
  Brick/Haystack, ontologies, graph rules, supervised taggers &
  Resolving ambiguous labels, extracting context from manuals and drawings, suggesting mappings for review &
  Human validation for spatial and equipment relations \\
Fault diagnosis and maintenance support &
  FDD algorithms, rule libraries, expert systems, knowledge graphs &
  Linking alarms, manuals, logs, and work orders into understandable diagnostic narratives &
  Recommendation only; technician or engineer acts \\
BEM generation and retrofit analysis &
  EnergyPlus/OpenStudio, calibration routines, optimisation engines &
  Translating requirements into model edits, orchestrating tools, documenting assumptions, comparing retrofit narratives &
  Modeller review before investment or control use \\
Technician training and procedural O\&M &
  Manuals, standard operating procedures, training programmes &
  Grounded step-by-step explanations, safety reminders, context-specific procedure lookup &
  Training and decision support; no unsupervised field action \\
\bottomrule
\end{tabularx}
\end{table*}

Two boundaries of this map deserve emphasis. First, LLMs appear least advantageous when inputs are already structured and the task has a stable numerical objective. In short-horizon load forecasting, for example, PatchTST outperforms the LLM-based models evaluated by \citet{zhu_enhanced_2025}, although the relative performance changes for longer forecasting horizons. Current LLM-based closed-loop and optimisation approaches also remain predominantly simulation-based and retain output-consistency, latency, and safety constraints~\cite{li_llm-assisted_2025,zhu_heating_2025}. Even the most operationally grounded case was evaluated in a plant-calibrated simulation environment, and it reports only a \qty{1.9}{\percent} gain over an expert baseline~\cite{liu_co-llm_2026}. Similarly, LLM-based metadata mapping can assist with ambiguous point labels, but tagging accuracy still degrades on complex equipment configurations and naming conventions, so human oversight remains necessary~\cite{zheng_mastering_2025,ababsa_continuity_2025}. These findings support using LLMs as semantic or advisory layers rather than assuming that they should replace specialised numerical and rule-based methods. Second, conventional baselines are not themselves fully field-proven. Both MPC and RL rest on more mature, verifiable formulations, but real-building demonstrations remain limited relative to simulation studies. The fit-for-purpose boundary should therefore be interpreted as a statement about current evidence, task structure, and risk rather than as a permanent capability ceiling.

\subsection{Practical Guidance}\label{sec:discussion_guidance}
Further addressing \emph{RQ4}, five bounded roles merit near-term practitioner attention. In each case the LLM reduces engineering effort by bridging a semantic gap; in each case a hard constraint defines where the role ends.

\emph{Point-name cleanup.} Few-shot prompting maps heterogeneous BAS point names to Brick or Haystack schemas without large labelled datasets~\cite{ababsa_continuity_2025,zheng_mastering_2025}. Automated relation accuracy for complex spatial relationships remains only partial~\cite{forth_semantic_2024}, so human validation remains necessary; deploy as human-in-the-loop, not autonomous pipeline.

\emph{Grounded document and knowledge-base support.} RAG systems improve answer precision for domain-specific queries by anchoring responses in retrieved building documents~\cite{li_rag_2026,chen_chiller_2026}. The binding constraint: retrieval quality determines output quality. Treat the knowledge base as a first-class engineering artefact requiring systematic curation.

\emph{MPC/RL advisory interfaces.} LLM wrappers around physics-based controllers explain rationale, translate operator intent, and propose slow-time-scale tuning, at \$0.013--0.032 per call~\cite{liang_physics-informed_2025}. Non-negotiable constraint: the LLM must not override the physics solver; a hard-coded fallback is required.

\emph{BEM workflow automation.} Multi-agent frameworks reduce expert BEM authoring time by over \qty{90}{\percent}~\cite{lu_automated_2025,zhang_automatic_2025}. The remaining gap is semantic validation: a generated IDF may compile correctly while encoding wrong geometry, schedules, or HVAC topology. Qualified engineer review before any investment or control use.

\emph{Technician training and procedural O\&M.} Grounded LLMs turn manuals, alarm logs, and work-order histories into step-by-step procedural support for less-experienced staff. The role is explanatory, not executive: retrieve and explain trusted procedures; do not generate new maintenance actions.

\emph{Quick start by readiness tier.}
\emph{Bounded trial roles (human-in-the-loop):} Point-name cleanup and metadata alignment; document-grounded operator and knowledge-base support; BEM workflow assistance; MPC/RL advisory interfaces; technician training and O\&M procedural support.
\emph{Validate first (research-only):} FDD reasoning without causal-reasoning verification; autonomous multi-step agentic operation; direct closed-loop control without physics guardrails; unvalidated synthetic occupant proxies.
\emph{Minimum package for any near-term role:} source grounding over trusted documents, human review before execution of safety-adjacent outputs, hard-coded fallback controller, audit logging of prompts and outputs.

Across all five guidance areas, a minimum deployment package applies regardless of the specific use case (Table~\ref{tab:deployment_package}).

\begin{table}[!h]\footnotesize
\centering
\caption{Minimum deployment package for LLM roles in HVAC operations. The failure-mode column references the taxonomy in Section~\ref{sec:discussion_failures}.}\label{tab:deployment_package}
\begin{tabularx}{\linewidth}{p{3.2cm} X p{1.2cm}}
\toprule
Requirement & Purpose & Failure mode if omitted \\
\midrule
Source grounding via retrieval over trusted documents, metadata, or logs &
  Anchors outputs in verified, building-specific context and reduces hallucination &
  FM2, FM3 \\
Human review before execution of safety-adjacent outputs &
  Keeps outputs advisory, with decision authority retained by qualified humans &
  FM4 \\
Hard-coded safety constraints outside the LLM layer &
  Enforces physical limits regardless of LLM failure mode or latency &
  FM5 \\
Fallback controller or operator override independent of LLM availability &
  Maintains operational continuity under API outage, rate-limiting, or error &
  FM5, FM6 \\
Logging of prompts, retrieved sources, outputs, and overrides &
  Enables auditability and post-hoc failure analysis &
  FM1, FM6 \\
Task-specific evaluation on real building data before deployment &
  Validates performance under actual sensor noise, point naming, and topology &
  FM1, FM3 \\
\bottomrule
\end{tabularx}
\end{table}

A concise reference of cost, latency, and readiness figures for the main deployment archetypes is provided in Table~\ref{tab:practical_archetypes} (\ref{app:supplementary}). Defensible LLM roles and minimum guardrails for five HVAC stakeholder groups are summarised in Table~\ref{tab:stakeholder_guidance} (\ref{app:supplementary}).

\subsection{Failure Modes and Evaluation Gaps}\label{sec:discussion_failures}
Completing the answer to \emph{RQ3}, the corpus reveals six recurring failure modes (FM) that standard evaluation metrics do not adequately capture.

\emph{FM1: Reasoning brittleness} is the most dangerous for HVAC applications.
Current A1 studies already show that strong top-line performance can mask
unsafe instability: \citet{xiao_exploring_2024} report run-to-run variation in
generated recommendations, while \citet{zhang_domain-specific_2025} show that near-perfect
fault-label accuracy can coexist with incorrect causal reasoning. A fault
classifier that produces the right label for the wrong reason cannot support
reliable diagnostic workflows.

\emph{FM2: Semantic hallucination in generated logic} affects orchestrator designs in A1 and A2. An LLM generating Python code for a data pipeline may produce syntactically valid code that silently implements an incorrect feature engineering step (normalising by the wrong reference period, selecting an incorrect lag, or misidentifying a sensor relationship) \cite{zhang_data-driven_2025,qiu_coding-free_2025}. This failure is not visible in downstream model accuracy until the pipeline is tested on out-of-sample data with different distributional properties.

\emph{FM3: ``Executable but wrong''} in BEM generation affects A3. A generated IDF file that compiles and runs may produce simulation outputs that are numerically plausible but physically incorrect: wrong floor areas, incorrect HVAC schedules, or mis-specified system types~\cite{liu_large_2026}. Current evaluation practice in A3 primarily assesses executability, not physical fidelity.

\emph{FM4: Plausible but unfaithful explanations} affect advisory systems in A4 and A5. A language model generating an explanation for a control decision or a comfort recommendation may produce text that is coherent, persuasive, and incorrect~\cite{zhang_building_2025}. Unlike numerical errors, explanation infidelity is hard to detect without domain expertise.

\emph{FM5: Latency incompatible with real-time control} is a structural barrier for A4. HVAC control loops typically operate at timescales ranging from seconds (e.g.\ fast VAV damper control) to several minutes (e.g.\ chiller sequencing); current agentic designs do not meet the tighter end of this range for direct control and would need consistently bounded end-to-end latency~\cite{sawada_office---loop_2025}.

\emph{FM6: Proprietary model dependence} blocks reproducibility and governance. The corpus remains strongly dependent on proprietary model families, especially GPT-4/GPT-4o and Gemini variants. Paper results cannot be reproduced cleanly when API models are updated silently, and building owners subject to data residency regulations cannot assume that cloud-hosted inference is acceptable~\cite{hong_ai_2025}. This dependence is a systemic reproducibility and governance problem that the field has not yet addressed systematically.

The corpus also reveals a \emph{cost and latency evaluation gap} that cuts across A1, A3, and A4. Table~\ref{tab:cost_latency_profile} (\ref{app:supplementary}) assembles  the principal directly reported cost and latency values identified in the reviewed corpus. Even this sparse subset spans orders of magnitude and reflects fundamentally different deployment contexts: offline audits, automated model-generation workflows, query-based simulation tools, and advisory control calls. For practitioners making deployment decisions, the main result is not a single benchmark value but the absence of a comparable reporting standard across most papers. Future work should therefore standardise cost-per-call and end-to-end latency reporting alongside accuracy and energy savings.

Standard metrics such as accuracy, F1, R\textsuperscript{2} and energy savings percentages do not expose failure modes FM1 through FM4. Additional evaluation targets should be adopted as standard practice: repeatability across prompt variants and model versions, calibration and uncertainty quantification, semantic plausibility checks against domain knowledge, and systematic failure-case audits covering at least the fault and edge-case categories relevant to the deployment environment.

\subsection{Future Work}\label{sec:discussion_future}
LLM capabilities in reasoning, tool use, and multimodal processing continue to improve with each model generation \cite{minaee_llm-survey_2025}, and these gains may reduce some technical barriers, although field validation, governance, and safety requirements remain independent constraints. Translating that capability progress into HVAC deployment, however, requires the field evidence infrastructure described below.

\emph{FW1: Portfolio-scale benchmarks with privacy-preserving release.}
The evidence-realism gap is most severe in fault detection and control (A1, A4), where field-validated systems remain limited to a small number of buildings and equipment types. As \citet{amangeldy_review_2025} argue, progress requires open, high-fidelity benchmarks that pair multivariate building and IoT data with standardised metadata and occupant feedback, alongside privacy-preserving evaluation designs that enable broader and more comparable validation. A specific and currently unaddressed target within this priority is a dedicated heat-pump FDD benchmark: no reviewed paper covers heat pumps (Section~\ref{sec:a1}), yet their defrost-cycle, part-load, and refrigerant-side fault signatures are physically distinct from the chiller and AHU patterns that dominate existing datasets, so reusing cooling-only fault libraries risks physically wrong explanations (FM4).

\emph{FW2: Context-grounded orchestration evaluation under operational latency and governance budgets.}
The cost and latency of LLM-based HVAC systems are under-evaluated. Only five papers quantify API cost directly (Table~\ref{tab:cost_latency_profile}): \citet{xiao_exploring_2024} report approximately \$5 per energy audit, \citet{liang_physics-informed_2025} report \$0.013--0.032 per MPC tuning call, \citet{zhang_automated_2024} report \$17.68 per chiller-plant analysis run, \citet{zhang_data-driven_2025} report \$0.318 for training and deploying one model, and \citet{quan_autobee_2025} report \$0.07--0.10 per analysis query. These data points are insufficient for practitioners to make deployment decisions, and they must be read with caution: API token pricing has fallen rapidly and varies by more than an order of magnitude across providers and model tiers, so any absolute figure reflects the model version and date tested rather than a stable deployment cost.

Systematic ablations are needed comparing M1 (prompting and supervised adaptation), M2 (context-grounded orchestration), and M3 (multimodal input) designs under matched operational constraints (latency, cost, and on-premise versus cloud governance) representative of each intended use case. Such ablations would allow the field to identify which designs are viable under real operational budgets rather than only under academic conditions.

\emph{FW3: Principled integration with physics-based and RL controllers.}
The most important technical gap in A4 is the absence of formal frameworks for LLM--MPC/RL integration with verifiable safety properties. Current designs either treat the LLM as a direct controller (with the risks identified in Section~\ref{sec:A4_DB}) or as an advisory wrapper without specifying what happens when the LLM fails. Near-term examples that begin to address this gap include \citet{hu_autocontrol_2025} (automated PI synthesis with defined fallback), \citet{liang_physics-informed_2025} (LLM tuner with MPC constraint enforcement), and \citet{sawada_office---loop_2025} (human-in-the-loop with occupant feedback as a safeguard). Physics-informed machine learning offers a precedent here, since it embeds physical priors directly in data-driven building models rather than wrapping them~\cite{jiang_physics_2025}.
Formal integration frameworks should specify the responsibility boundary, the fallback controller behaviour, the conditions under which the LLM layer is bypassed, and the safety test coverage required for promotion to higher autonomy levels.

\emph{FW4: Governance-preserving deployment architectures.}
Five A5 papers and several A1/A4 systems ingest presence data, occupant feedback text, or detailed occupancy logs, data classes regulated under GDPR in European buildings and under analogous frameworks in other jurisdictions. The corpus does not include a single paper that evaluates an on-premise or federated deployment under real data-governance constraints. Three concrete directions address this gap: (1)~on-premise open-weight model validation, demonstrating which HVAC tasks can be handled by locally hosted models without cloud API access; (2)~federated inference designs that allow portfolio-scale learning without centralising occupant data; and (3)~differential privacy mechanisms for occupant preference models that make individual-level data inaccessible to the LLM while preserving aggregate preference signals. These are not merely regulatory compliance exercises; they are a prerequisite for deployment in public-sector and healthcare buildings where data governance requirements are strictest and the energy-efficiency need is greatest.

The failure modes map directly onto the future-work priorities: FM1 and FM2 (reasoning inconsistency; hallucination in generated logic) require the field-validated benchmarks and matched-constraint ablations of FW1--FW2; FM3 (executable but wrong BEM) requires the physics-integrated validation of FW3; FM4 (unfaithful explanations) requires explanation-fidelity evaluation within FW2; FM5 (control latency) is addressed by latency-constrained design evaluation in FW2 and formal LLM--MPC integration in FW3; and FM6 (proprietary model dependence) maps directly onto the governance-preserving deployment architectures of FW4. Table~\ref{tab:fm_fw_map} makes this mapping explicit.

\begin{table}[!h]\footnotesize
\centering
{\setlength{\tabcolsep}{2pt}
\caption{Failure modes (FM1--FM6) mapped to future-work priorities (FW1--FW4). Filled circles indicate a direct relationship.}\label{tab:fm_fw_map}
\begin{tabular}{lcccc}
\toprule
 & FW1 & FW2 & FW3 & FW4 \\
\midrule
FM1 Reasoning brittleness          & $\bullet$ & $\bullet$ &            &            \\
FM2 Semantic hallucination         & $\bullet$ & $\bullet$ &            &            \\
FM3 Executable but wrong           &           & $\bullet$ & $\bullet$  &            \\
FM4 Unfaithful explanations        &           & $\bullet$ &            &            \\
FM5 Latency / real-time control    &           & $\bullet$ & $\bullet$  &            \\
FM6 Proprietary model dependence   &           &           &            & $\bullet$  \\
\bottomrule
\end{tabular}
}
\end{table}

In addition to the four immediate priorities, five near-term capability developments will shift which barriers remain binding, without changing the deployment-readiness classification of the current corpus. Field validation, governance, and safety architecture remain prerequisites for any movement from research-only to near-term status.

\emph{Reasoning models and FM1 brittleness.}
LLMs augmented with extended test-time computation~\cite{minaee_llm-survey_2025} directly target FM1 (the gap between correct outputs and correct causal reasoning~\cite{zhang_domain-specific_2025}) through stronger intermediate reasoning and self-checking in A1 diagnostics and A3 model generation. The trade-off is latency: acceptable offline, but problematic for short-cycle A4 control.

\emph{Long-context inference and retrieval limitations.}
Frontier context windows large enough to ingest full building management system (BMS) point catalogues, seasonal histories, and manuals without chunking~\cite{minaee_llm-survey_2025} can reduce FM2 (semantic hallucination from incomplete or mis-ranked context in M2 systems). The constraint is governance: sending complete operational logs to cloud models intensifies the FW4 data-residency concerns, making large-context \emph{on-premise} deployment the more relevant near-term target.

\emph{MCP maturation and the integration barrier.}
MCP offers an open standard for exposing external tools as callable interfaces~\cite{li_mcp_2026}; \citet{li_energyplus-mcp_2025} demonstrate it for EnergyPlus. As comparable interfaces emerge for BIM platforms, ontology services, and building data systems, the project-specific engineering effort that currently limits A3$\times$M2 and A4$\times$M2 workflows is likely to fall. The surrounding infrastructure, not only the model, then becomes an important source of capability in its own right.

\emph{On-premise open-weight models and governed deployment.}
Fine-tuned open-weight models already in this corpus achieve competitive performance on bounded HVAC tasks without proprietary APIs: DistilBERT achieves F1 of \qtyrange{82}{99}{\percent}~\cite{langer_fault_2025}, and T5-family models report \qty{100}{\percent} IDF accuracy~\cite{jiang_efficient_2025}. The strongest FDD result in the corpus is a LoRA-adapted GPT-3.5~\cite{zhang_domain-specific_2025}, but that model is closed-weight and API-only, so it does not demonstrate the on-premise capability at issue here. For document-heavy O\&M workflows (procedure lookup, manual retrieval, work-order triage) locally hosted, retrieval-grounded systems provide knowledge access without cloud transfer, directly addressing FM6 and the FW4 governance gap.

\emph{From context engineering to harness engineering.}
The corpus already shows a shift from prompt-centred use (M1) toward context-grounded orchestration (M2). The next step is \emph{harness engineering} (capability that depends on orchestration logic, tool design, memory management, safety constraints, and fallback protocols) not a new taxonomy axis but a maturation within M2. It shifts the central question from whether the model can perform a task to whether the overall system can do so safely and reliably under operational conditions.

Professional associations such as ASHRAE are launching educational and research activities on generative AI for their HVAC engineer members through the MTG (Multi-disciplinary Task Group) on generative AI, to tackle some of the challenges of applying LLMs to the HVAC industry. 

\subsection{Limitations of this Review}\label{sec:limitations}
This review has several \emph{scope and methodological limitations}. First, the evidence base is deliberately limited to Scopus-indexed, English-language, peer-reviewed publications. This boundary is appropriate for the present review because the objective is not to catalogue every emerging prototype, poster, or vendor demonstration, but to compare deployment claims against a corpus with stable bibliographic metadata, reproducible screening, and a minimum level of reviewed evaluative content. In a still-small literature concentrated in a limited set of core journals and conference venues, a single curated index keeps deployment-readiness claims anchored to peer-reviewed evidence and improves replicability. The trade-off is that grey literature, arXiv preprints, and some conference items outside Scopus are excluded by design. Excluding these sources may omit both leading-edge capability demonstrations and negative or failed deployments. The readiness findings should therefore be read as statements about the Scopus-indexed, peer-reviewed corpus rather than as bounds on the field as a whole.

Second, the time window from January 2023 to March 2026 deliberately excludes earlier LLM-adjacent building work (chatbot interfaces, pre-GPT-4 language models) but may miss foundational infrastructure work that informs current deployments. \citet{hong_sota_2020} survey the preceding decade of machine-learning applications across the building life cycle, including the adoption barriers already identified there.

In particular, the broader MCP tool ecosystem has expanded rapidly around the 1~April 2026 search date. References to this wider ecosystem, beyond the single MCP contribution in the corpus,~\cite{li_energyplus-mcp_2025} are post-search observations rather than coded evidence. The same applies to task-oriented agentic systems more broadly, which often appear in repositories and preprints before indexed venues.

Third, all coding (application category, method category, evidence type, deployment readiness, and responsibility boundary) was performed by a single rater without formal inter-rater reliability testing; Cohen's kappa and disagreement rates are therefore not reported. This is an important limitation because several coding dimensions, particularly deployment readiness and responsibility boundary, involve structured interpretation; explicit coding definitions and the full publication of paper-level assignments improve auditability, but independent replication remains desirable.

Two mitigations partially address this. First, every code is operationally defined with explicit, auditable thresholds (for example, \emph{research-only} requires simulation-only evidence or a fundamental capability gap, and \emph{near-term} requires demonstrated feasibility on real data with a specific unresolved barrier) and the complete per-paper coding for all 66 studies is provided as machine-readable supplementary material (see Data Availability) for independent re-assessment. Second, the M1--M3 taxonomy deliberately compresses within-category heterogeneity (e.g.\ prompting vs.\ fine-tuning within M1), which aids synthesis but abstracts over subcategory nuances visible only in the paper-level analysis. These labels should therefore be read as structured assessments rather than objective measurements, and as a snapshot of the evidence base up to March 2026 rather than a fixed judgement of long-term maturity.

\section{Conclusion}\label{sec:conclusion}
This review mapped 66 peer-reviewed studies on LLMs for HVAC operations across five application categories and three method categories. It interprets them through the joint lens of evidence realism, deployment readiness, and responsibility boundary. The corpus is concentrated in building energy modelling, with context-grounded orchestration (M2) the dominant methodological cluster, while multimodal work remains sparse and control evidence is substantially weaker than publication counts alone suggest. The contribution is therefore not only a literature map, but also a deployment-oriented framework for judging where LLMs are \emph{technically plausible} and where they are \emph{operationally defensible}.

Evidence remains dominated by offline and pre-deployment evaluation: only four pilot studies were identified, and standard benchmark metrics overestimate reliability because they rarely capture run-to-run consistency, semantic plausibility, or robustness to real-world disturbances. The deployment-readiness coding accordingly found no ready-now role, three near-term cases, and 63 research-only papers (Section~\ref{sec:discussion_map}). The binding barriers are not primarily capability gaps (reasoning models, long-context inference, MCP tool interfaces, and open-weight models keep lowering the technical floor) but field validation, safety architecture, and governance, which benchmark scores cannot substitute for, and which define the four priorities identified here (FW1--FW4).

On current evidence, LLMs are most defensible in HVAC as documentation interfaces, integration accelerators, semantic support layers, and advisory wrappers around established tools and controllers. These roles address the semantic, organisational, and institutional barriers that have long limited advanced methods in real, heterogeneous, and frequently modified building stock. Their present value lies in reducing integration friction, improving access to operational knowledge, and supporting human decision-making rather than replacing it. Progress toward autonomous HVAC roles will continue to require field validation, explicit safety architectures, and governance-preserving deployment that go beyond benchmark accuracy alone. As models and orchestration frameworks advance rapidly, the boundary between research-only and deployable roles will move quickly. Disciplined, evidence-based judgement of \emph{where} language models belong in HVAC operations will therefore matter more, not less.

\section*{CRediT authorship contribution statement}
\textbf{Alexander Neubauer:} Conceptualization, Methodology, Investigation, Formal analysis, Data curation, Visualization, Writing - original draft, Writing - review \& editing.
\textbf{Tianzhen Hong:} Conceptualization, Writing - review \& editing, Supervision, Resources.
\textbf{Han Li:} Conceptualization, Writing - review \& editing.
\textbf{Mengbo Yu:} Writing - review \& editing.
\textbf{Amin Darbandi:} Writing - review \& editing.
\textbf{Yannick Fürst:} Writing - review \& editing.
\textbf{Martin Kriegel:} Project administration, Funding acquisition, Supervision, Resources.

\section*{Acknowledgements}
This work is funded by the German Federal Ministry for Economic Affairs and Energy (BMWE) in the framework of the research program EnOB: RR-RLT 03EN1109. The support is gratefully acknowledged. The first author gratefully acknowledges Lawrence Berkeley National Laboratory for hosting a research stay during which this manuscript was prepared.

\section*{Data Availability}
The coded corpus underlying all review-level counts, figures, and the deployment-readiness synthesis is available as supplementary material: an XLSX of 66 peer-reviewed studies with application-category, method-category, evidence-type, deployment-readiness, and responsibility-boundary fields (Supplementary Material S1).

\section*{Declaration of Competing Interest}
The authors declare the following competing interests. This manuscript is submitted to a journal for which Tianzhen Hong serves as Executive Editor and to a special issue for which Han Li serves as Guest Editor. The authors request that the manuscript be handled independently by another editor, without any involvement of Tianzhen Hong or Han Li in the editorial decision-making or peer-review process. In addition, two studies included in the review corpus were co-authored by members of the author team, namely \citet{li_energyplus-mcp_2025} and \citet{li_rag_2026}. Both were screened, coded, and discussed using the same criteria as all other included studies. The authors declare no other competing interests.

\section*{Declaration of Generative AI and AI-assisted technologies in the writing process}
During the preparation of this work, the authors used ChatGPT, Claude, and NotebookLM to assist with language editing, readability improvement, and structured extraction of information from the reviewed literature. After using these tools, the authors reviewed and edited the content as needed.
\appendix
\setcounter{table}{0}
\renewcommand{\thetable}{A\arabic{table}}

\section{Corpus Summary Tables}\label{app:corpus_tables}
The following tables provide per-application-category summaries of the reviewed papers, listing system scope, LLM methodology category, key result, primary limitation, and deployment-readiness classification. Tables~\ref{tab:a1_summary}, \ref{tab:a2_summary}, and~\ref{tab:a5_summary} cover the complete primary-coded sets for A1, A2, and A5; Tables~\ref{tab:a3_summary} and~\ref{tab:a4_summary} present selected studies from the larger A3 and A4 corpora, namely those discussed in the corresponding application sections. The complete paper-level coding for all 66 studies is provided as Supplementary Material S1.

\begin{table*}[!h]\footnotesize
\centering
\caption{Primary-coded A1 papers: fault detection and diagnosis. Depl.\ = deployment readiness.}\label{tab:a1_summary}
{\setlength{\tabcolsep}{3pt}
\begin{tabularx}{\textwidth}{p{0.4cm} X p{2.4cm} X p{2.3cm} p{1.9cm} p{1.5cm}}
\toprule
Ref & System scope & LLM methodology & Key result & Primary limitation & Evidence & Deployment \\
\midrule
\cite{zhang_automated_2024} & Chiller plant, Shenzhen & Workflow automation; itemset interpretation & GPT-4 recall \qty{89.17}{\percent}, cost \$17.68 & API cost; offline audit only & Retrospective field data & research-only \\
\cite{zhang_domain-specific_2025} & AHU/VAV/chiller & LoRA fine-tuning; supervised FDD classifier & Accuracy \qty{29.5}{\percent}$\rightarrow$\qty{100}{\percent} after fine-tuning & Correct labels, flawed causal reasoning & Retrospective field data & research-only \\
\cite{langer_fault_2025} & Buildings' HVAC fault detection & Fine-tuned DistilBERT benchmark & F1 \qtyrange{82}{99}{\percent} & Simulation-data benchmark; no field data & Simulation-only & research-only \\
\cite{ravandi_integration_2025} & Facility management work orders & GPT-3.5 advisory NL-to-action workflow & Advisory NL-to-action integration & Work-order completeness dependency & Retrospective field data & research-only \\
\cite{qiu_coding-free_2025} & Metro station chilled water & LLM-guided ML virtual flowmeter & RMSE \qtyrange{8.84}{9.05}{\cubic\metre\per\hour} & Single-building validation only & Retrospective field data & research-only \\
\cite{chen_chiller_2026} & Industrial chiller systems & RAG-based maintenance copilot & Hallucinations reduced by \qty{31.2}{\percent} & Knowledge-base curation required & Lab study & research-only \\
\cite{deng_graph-constrained_2026} & Large office HVAC & GC-ARM + knowledge-enhanced LLM & Accuracy \qty{89.1}{\percent}, rule space reduced \qty{95}{\percent} & Ontology upkeep at scale & Retrospective field data & research-only \\
\cite{chen_customized_2025} & Human--AI O\&M collaboration & Customized maintenance assistant & \qty{96.3}{\percent} fault-diagnosis accuracy & Tool development burden remains high & Lab study & research-only \\
\cite{xiao_exploring_2024} & Unstructured building data & Multi-agent optimisation workflow & RAG retrieval F1 0.8894; ${\sim}$\$5/audit & Run-to-run variation in recommendations & Retrospective field data & research-only \\
\bottomrule
\end{tabularx}
}
\end{table*}

\begin{table*}[!h]\footnotesize
\centering
\caption{Primary-coded A2 papers: energy load forecasting.}\label{tab:a2_summary}
{\setlength{\tabcolsep}{3pt}
\begin{tabularx}{\textwidth}{p{0.4cm} X p{2.4cm} X p{2.3cm} p{1.9cm} p{1.5cm}}
\toprule
Ref & System scope & LLM methodology & Key result & Primary limitation & Evidence & Deployment \\
\midrule
\cite{liu_towards_2025} & Building Genome Project & Metadata-to-text conditioning for diffusion forecasting & MAE 10.58, CRPS 12.01 & No prospective real-building validation & Retrospective field data & research-only \\
\cite{zhang_data-driven_2025} & Two real buildings (Shenzhen, London) & Bayesian-optimised GPT pipeline & Avg.\ R\textsuperscript{2}~=~0.95, cost \$0.318 per model trained and deployed & Semantic errors in generated code & Retrospective field data & research-only \\
\cite{zhu_enhanced_2025} & Fujian commercial centre & Frozen BERT-base; patch reprogramming & MAE reduced \qty{17.3}{\percent}, RMSE \qty{19.3}{\percent} & Short-horizon lag; single dataset & Retrospective field data & research-only \\
\bottomrule
\end{tabularx}
}
\end{table*}

\begin{table*}[!h]\footnotesize
\centering
\caption{Selected primary-coded A3 papers: building energy modelling, simulation, and interoperability.}\label{tab:a3_summary}
{\setlength{\tabcolsep}{3pt}
\begin{tabularx}{\textwidth}{p{0.4cm} X p{2.4cm} X p{2.3cm} p{1.9cm} p{1.5cm}}
\toprule
Ref & System scope & LLM methodology & Key result & Primary limitation & Evidence & Deployment \\
\midrule
\cite{jiang_eplus-llm_2024} & 152 synthetic test cases & Fine-tuned T5; NL to IDF & \qty{100}{\percent} IDF accuracy, $>\qty{95}{\percent}$ effort reduction & Synthetic cases; no physical fidelity & Simulation-only & research-only \\
\cite{jiang_efficient_2025} & 402 modelling cases & LoRA-tuned T5-11B & \qty{100}{\percent} accuracy (geometry/parameter/simulation match), $>\qty{98}{\percent}$ effort reduction & Bounded benchmark setting & Simulation-only & research-only \\
\cite{jiang_prompt_2025} & 648 LLM case studies & Few-shot prompting across 18 LLMs & ${\sim}$\qty{25}{\percent} (one-shot) to \qty{31.5}{\percent} (two-shot) of cases rated Perfect & Performance drops on complex cases & Simulation-only & research-only \\
\cite{ababsa_continuity_2025} & 3 real buildings + 1 synthetic & Few-shot BMS-to-Brick tagging & F1 \qtyrange{92}{98}{\percent} for Brick tagging & Spatial relation inference under-tested & Retrospective field data & research-only \\
\cite{elsayed_user-friendly_2025} & $>$50 validation instances & Text/image to Honeybee/EnergyPlus & \qty{100}{\percent} convergence, avg.\ 7.2~s (text), 13.4~s (image) & Image-to-mesh geometry over-simplification & Simulation-only & research-only \\
\cite{jia_natural_2025} & building sensor system & LLM-integrated sensor-data protocol & Pilot natural-language interaction & Token limit of 4K constrains use & Pilot deployment & research-only \\
\cite{lu_automated_2025} & Real retrofit scenario & Multi-agent OpenStudio workflow & Time reduced by \qty{90}{\percent}; meets Guideline 14 & Single retrofit; GPT-4 dependency & Retrospective field data & research-only \\
\cite{li_energyplus-mcp_2025} & EnergyPlus IDF examples & MCP server; 35 EnergyPlus tools & End-to-end tool access via natural language & No deployment validation beyond tooling & Simulation-only & research-only \\
\cite{li_rag_2026} & 13 LBNL buildings & RAG pipeline for building-data grounding & \qty{30}{\percent} coverage and structure-preservation gain & No sustained operational deployment & Lab study & near-term \\
\cite{zhang_automatic_2025} & 10 iUnit trials & Four-agent debug loop & \qty{100}{\percent} success; expert time \qty{4}{\hour}$\rightarrow$\qty{9}{\minute} & Semantic validation missing & Lab study & research-only \\
\bottomrule
\end{tabularx}
}
\end{table*}

\begin{table*}[!h]\footnotesize
\centering
\caption{Selected primary-coded A4 papers: HVAC control and optimisation. Rows show the studies discussed in Section~\ref{sec:a4}; the advisory-MPC wrapper of \citet{liang_physics-informed_2025} is characterised in Tables~\ref{tab:deployment_readiness} and~\ref{tab:cost_latency_profile} instead.}\label{tab:a4_summary}
{\setlength{\tabcolsep}{3pt}
\begin{tabularx}{\textwidth}{p{0.4cm} X p{2.4cm} X p{2.3cm} p{1.9cm} p{1.5cm}}
\toprule
Ref & System scope & LLM methodology & Key result & Primary limitation & Evidence & Deployment \\
\midrule
\cite{sawada_office---loop_2024} & Real office, 30 workers, two periods (${\approx}$7.5 weeks) & Multimodal prompting; IoT + occupant feedback & \qty{47.9}{\percent} energy savings, \qty{26.4}{\percent} comfort gain (LLM-only condition) & Prediction divergence in dynamic weather & Pilot deployment & near-term \\
\cite{liu_co-llm_2026} & Tertiary hospital HVAC system & Retrieval-grounded optimisation support & \qty{1.9}{\percent} energy-saving gain & Output consistency remains fragile & Simulation-only & research-only \\
\cite{hu_autocontrol_2025} & Building energy-system control design & End-to-end controller synthesis & Comparable control quality without manual design & Simulation-only validation & Simulation-only & research-only \\
\cite{ko_darlin_2025} & Multi-zone office building & RAG-based setpoint-control interface & Interpretable setpoints with energy savings & Uncertain LLM outputs & Simulation-only & research-only \\
\cite{zhang_building_2025} & Building energy management system & RCoT explanation layer & Improved explanation quality & Simulation-tested; no deployment validation & Simulation-only & research-only \\
\cite{zhu_heating_2025} & HVAC temperature and humidity control & LLM-enhanced MPC strategy & Simulation-only energy reduction claims & No real-building validation & Simulation-only & research-only \\
\cite{jin_democratizing_2024} & EV charging and HVAC control & ChatGPT-based decision support & Simulation proof of concept & Relies on LLM reasoning quality & Simulation-only & research-only \\
\bottomrule
\end{tabularx}
}
\end{table*}

\begin{table*}[!t]\footnotesize
\centering
\caption{Primary-coded A5 papers: thermal comfort and occupant interaction.}\label{tab:a5_summary}
{\setlength{\tabcolsep}{3pt}
\begin{tabularx}{\textwidth}{p{0.4cm} X p{2.4cm} X p{2.3cm} p{1.9cm} p{1.5cm}}
\toprule
Ref & System scope & LLM methodology & Key result & Primary limitation & Evidence & Deployment \\
\midrule
\cite{arslan_monitoring_2025} & Office IEQ monitoring & Context-grounded comfort interface & \qty{94}{\percent} accuracy, \qty{92}{\percent} precision & Limited generalisation beyond the case setting & Retrospective field data & research-only \\
\cite{sadick_what_2025} & \num{14622} IEQ feedback entries & Fine-tuned IEQ-BERT & \qty{93}{\percent} accuracy, F1~=~0.93 & Public web-app deployment; no BAS integration & Retrospective field data & research-only \\
\cite{qaisar_dynamic_2025} & Office data, Hebei and National University of Singapore & Few-shot ICL for dynamic cooling behaviour & \qty{95.8}{\percent} binary occupancy-detection accuracy (Gemini) & No longitudinal preference learning & Retrospective field data & research-only \\
\cite{liu_integrating_2025} & Conversational comfort requests & Function-calling LLM with RL interface & \qty{95}{\percent} function-call accuracy & Research-only control integration & Lab study & research-only \\
\cite{liu_conversational_2026} & Conversational preference learning & M1 comfort dialogue model & Month-long participant study in a real office yielding 411 trajectories; model trained and evaluated offline on that dataset & Behaviour drift not tested & Lab study & research-only \\
\cite{liu_human---loop_2026} & Persona-based comfort simulation & Human-in-the-loop preference modelling & Simulation-only preference proxy & Real physiology not validated & Simulation-only & research-only \\
\cite{kanayo_ai_2024} & 10-participant user study & LangChain energy-advisory chatbot & \qty{93}{\percent} entity-extraction accuracy & No multilingual support for the audio feature & Retrospective field data & research-only \\
\bottomrule
\end{tabularx}
}
\end{table*}

\clearpage
\section{Supplementary Reference Tables}\label{app:supplementary}

\begin{table*}[t!]\footnotesize
\centering
\caption{Fit-for-purpose deployment archetypes distilled from this corpus.
The \emph{Use horizon} column reports role-level practitioner guidance synthesised across the corpus; it is not a per-study deployment-readiness code.
Empty cost/latency cells indicate that the source papers did not report a direct figure.}\label{tab:practical_archetypes}
\begin{tabularx}{\textwidth}{X p{3.3cm} p{3.8cm} p{1.7cm} p{1.9cm} p{1.5cm}}
\toprule
Use case & Typical inputs & Typical outputs & Reported cost / latency & Use horizon & Evidence \\
\midrule
Point-name cleanup and metadata alignment &
BAS point lists; metadata exports &
Normalised names; Brick / Haystack tags &
 &
Bounded trial &
\cite{ababsa_continuity_2025,forth_semantic_2024} \\
Grounded document and knowledge-base support &
Manuals; BAS documents; retrieved tables and images &
Grounded answers; diagnostic support; structured metadata &
5--20\,s &
Bounded trial &
\cite{arslan_monitoring_2025,chen_chiller_2026,li_rag_2026} \\
BEM workflow assistance &
Natural-language model descriptions; IDF / OpenStudio files &
Generated or debugged models; calibration support &
7.2\,s (text path) &
Bounded trial &
\cite{elsayed_user-friendly_2025,lu_automated_2025,zhang_automatic_2025} \\
Structured NL-to-IDF generation on bounded tasks &
Constrained natural-language model descriptions &
Executable IDF candidates &
 &
Validate first &
\cite{jiang_efficient_2025} \\
Controller advisory or optimisation support &
Forecasts; controller state; comfort and constraint signals &
Explanations; tuning suggestions; slow-time-scale advisory setpoints &
\$0.013--0.032 / call &
Validate first &
\cite{liang_physics-informed_2025,liu_co-llm_2026} \\
Natural-language tool access to EnergyPlus &
Natural-language simulation requests &
Tool calls; model queries; workflow automation &
 &
Bounded trial &
\cite{li_energyplus-mcp_2025} \\
Occupant feedback classification &
Textual IEQ feedback &
Complaint classes; triage signal &
 &
Research stage &
\cite{sadick_what_2025} \\
Technician training and procedural O\&M support &
Manuals; work orders; alarm histories; control sequences &
Step-by-step explanations; safety reminders; procedure lookup &
 &
Bounded trial &
\cite{chen_chiller_2026,chen_customized_2025} \\
\bottomrule
\end{tabularx}
\end{table*}

\begin{table*}[!h]\footnotesize
\centering
\caption{Defensible LLM roles, main risks, and required guardrails by stakeholder group.}\label{tab:stakeholder_guidance}
\begin{tabularx}{\textwidth}{p{2.5cm} X p{3.2cm} p{3.8cm}}
\toprule
Stakeholder & Defensible LLM role & Main risk & Required guardrail \\
\midrule
Facility operator &
  Document-grounded troubleshooting; alarm interpretation; maintenance procedure lookup &
  Hallucinated diagnosis &
  Source citation; human approval before action \\
Controls engineer &
  BEM or MPC workflow support; automated model calibration assistance &
  Wrong assumptions propagated to model or controller &
  Simulation check; qualified review before investment or control use \\
Building owner / asset manager &
  Portfolio point-name normalisation; metadata cleanup and alignment &
  Inconsistent tagging at scale &
  Ontology alignment check; spot-check audit trail \\
Occupant &
  Comfort feedback interface; preference capture &
  Privacy exposure; data residency &
  Consent mechanism; on-premise processing \\
Regulator / auditor &
  Explanation and traceability of control decisions &
  Unverifiable or incomplete reasoning chains &
  Logged evidence chain; human expert countersignature \\
\bottomrule
\end{tabularx}
\end{table*}

\begin{table*}[!h]\footnotesize
\centering
\caption{Reported cost and latency figures in the reviewed corpus. Empty cells indicate that the source paper did not report a direct value.}\label{tab:cost_latency_profile}
\begin{tabularx}{\textwidth}{p{0.7cm} p{0.8cm} p{4.2cm} p{2.2cm} p{2.3cm} X}
\toprule
Paper & Cat. & Use case & Reported cost & Reported latency & Context \\
\midrule
\cite{xiao_exploring_2024} & A1 & Energy audit / optimisation workflow & $\sim$\$5 / audit &  & Multi-agent optimisation from unstructured data \\
\cite{deng_graph-constrained_2026} & A1 & Diagnostic reasoning &  & 2--5\,s / inference & Real office HVAC operational data \\
\cite{zhang_automated_2024} & A1 & Chiller-plant workflow automation & \$17.68 / analysis run &  & Offline audit on real plant data \\
\cite{zhang_data-driven_2025} & A2 & Automated load-prediction pipeline & \$0.318 / training and deployment &  & Two real buildings; Bayesian prompt search \\
\cite{li_rag_2026} & A3 & RAG building-data grounding &  & 5--20\,s & 13-building campus-scale pipeline \\
\cite{elsayed_user-friendly_2025} & A3 & Text/image to Honeybee-EnergyPlus &  & 7.2\,s (text) & User-facing model-generation interface \\
\cite{liang_physics-informed_2025} & A4 & MPC tuning call & \$0.013--0.032 / call &  & Advisory wrapper around physics-based control \\
\cite{quan_autobee_2025} & A3 & Building-performance analysis (incl.\ PMV) & \$0.07--0.10 / query &  & Simulation-based EnergyPlus workflow with IDF modification and RAG reporting \\
\bottomrule
\end{tabularx}
\end{table*}

\clearpage

\bibliographystyle{unsrtnat}
\bibliography{bibliographic}

\end{document}

%% file: figures/graphical_abstract_revised.pdf_tex
\begingroup%
  \makeatletter%
  \providecommand\color[2][]{%
    \errmessage{(Inkscape) Color is used for the text in Inkscape, but the package 'color.sty' is not loaded}%
    \renewcommand\color[2][]{}%
  }%
  \providecommand\transparent[1]{%
    \errmessage{(Inkscape) Transparency is used (non-zero) for the text in Inkscape, but the package 'transparent.sty' is not loaded}%
    \renewcommand\transparent[1]{}%
  }%
  \providecommand\rotatebox[2]{#2}%
  \newcommand*\fsize{\dimexpr\f@size pt\relax}%
  \newcommand*\lineheight[1]{\fontsize{\fsize}{#1\fsize}\selectfont}%
  \ifx\svgwidth\undefined%
    \setlength{\unitlength}{650.24999784bp}%
    \ifx\svgscale\undefined%
      \relax%
    \else%
      \setlength{\unitlength}{\unitlength * \real{\svgscale}}%
    \fi%
  \else%
    \setlength{\unitlength}{\svgwidth}%
  \fi%
  \global\let\svgwidth\undefined%
  \global\let\svgscale\undefined%
  \makeatother%
  \begin{picture}(1,0.50294009)%
    \lineheight{1}%
    \setlength\tabcolsep{0pt}%
    \put(0.50062666,0.48872249){\color[rgb]{0.14117647,0.19215686,0.23529412}\makebox(0,0)[t]{\lineheight{1.25}\smash{\begin{tabular}[t]{c}\textbf{{\large Large Language Models for HVAC Operations in Building Energy Systems}}\end{tabular}}}}%
    \put(0.5006274,0.47026805){\color[rgb]{0.41960784,0.4627451,0.50196078}\makebox(0,0)[t]{\lineheight{1.25}\smash{\begin{tabular}[t]{c}{\normalsize A Critical Review of Methods, Applications, and Deployment Readiness}\end{tabular}}}}%
    \put(0,0){\includegraphics[width=\unitlength,page=1]{graphical_abstract_revised.pdf}}%
    \put(0.14936563,0.42874556){\color[rgb]{1,1,1}\makebox(0,0)[t]{\lineheight{1.25}\smash{\begin{tabular}[t]{c}\textbf{1. HVAC operational bottlenecks}\end{tabular}}}}%
    \put(0.14936563,0.4045241){\color[rgb]{0.14117647,0.19215686,0.23529412}\makebox(0,0)[t]{\lineheight{1.25}\smash{\begin{tabular}[t]{c}\textbf{Review question}\end{tabular}}}}%
    \put(0.14936564,0.38952987){\color[rgb]{0.33333333,0.38039216,0.41960784}\makebox(0,0)[t]{\lineheight{1.25}\smash{\begin{tabular}[t]{c}How can LLMs support HVAC operations,\end{tabular}}}}%
    \put(0.14936563,0.37684244){\color[rgb]{0.33333333,0.38039216,0.41960784}\makebox(0,0)[t]{\lineheight{1.25}\smash{\begin{tabular}[t]{c}what evidence exists, and what is deployment-ready?\end{tabular}}}}%
    \put(0,0){\includegraphics[width=\unitlength,page=2]{graphical_abstract_revised.pdf}}%
    \put(0.02364475,0.34916078){\color[rgb]{0.14117647,0.19215686,0.23529412}\makebox(0,0)[lt]{\lineheight{1.25}\smash{\begin{tabular}[t]{l}\textbf{Problem}\end{tabular}}}}%
    \put(0.02364475,0.33185974){\color[rgb]{0.25882353,0.29803922,0.33333333}\makebox(0,0)[lt]{\lineheight{1.25}\smash{\begin{tabular}[t]{l}Point names, metadata, and manuals\end{tabular}}}}%
    \put(0.02364475,0.31801891){\color[rgb]{0.25882353,0.29803922,0.33333333}\makebox(0,0)[lt]{\lineheight{1.25}\smash{\begin{tabular}[t]{l}are fragmented across BAS portfolios.\end{tabular}}}}%
    \put(0,0){\includegraphics[width=\unitlength,page=3]{graphical_abstract_revised.pdf}}%
    \put(0.02364475,0.2684226){\color[rgb]{0.14117647,0.19215686,0.23529412}\makebox(0,0)[lt]{\lineheight{1.25}\smash{\begin{tabular}[t]{l}\textbf{Consequence}\end{tabular}}}}%
    \put(0.02364475,0.25112156){\color[rgb]{0.25882353,0.29803922,0.33333333}\makebox(0,0)[lt]{\lineheight{1.25}\smash{\begin{tabular}[t]{l}Operational data stay hard to query,\end{tabular}}}}%
    \put(0.02364475,0.23728073){\color[rgb]{0.25882353,0.29803922,0.33333333}\makebox(0,0)[lt]{\lineheight{1.25}\smash{\begin{tabular}[t]{l}reuse, validate, and explain.\end{tabular}}}}%
    \put(0,0){\includegraphics[width=\unitlength,page=4]{graphical_abstract_revised.pdf}}%
    \put(0.02364475,0.18768442){\color[rgb]{0.12156863,0.30980392,0.44705882}\makebox(0,0)[lt]{\lineheight{1.25}\smash{\begin{tabular}[t]{l}\textbf{LLM role}\end{tabular}}}}%
    \put(0.02364475,0.17038338){\color[rgb]{0.17647059,0.25882353,0.32156863}\makebox(0,0)[lt]{\lineheight{1.25}\smash{\begin{tabular}[t]{l}Semantic interface between documents,\end{tabular}}}}%
    \put(0.02364475,0.15654255){\color[rgb]{0.17647059,0.25882353,0.32156863}\makebox(0,0)[lt]{\lineheight{1.25}\smash{\begin{tabular}[t]{l}operators, tools, and HVAC workflows.\end{tabular}}}}%
    \put(0.02364475,0.13808811){\color[rgb]{0.31764706,0.37647059,0.42745098}\makebox(0,0)[lt]{\lineheight{1.25}\smash{\begin{tabular}[t]{l}Translate intent - normalise names - explain outputs\end{tabular}}}}%
    \put(0,0){\includegraphics[width=\unitlength,page=5]{graphical_abstract_revised.pdf}}%
    \put(0.05594002,0.07003736){\color[rgb]{0.14117647,0.19215686,0.23529412}\makebox(0,0)[t]{\lineheight{1.25}\smash{\begin{tabular}[t]{c}\textbf{BAS}\end{tabular}}}}%
    \put(0.05594002,0.05504313){\color[rgb]{0.36862745,0.41176471,0.44705882}\makebox(0,0)[t]{\lineheight{1.25}\smash{\begin{tabular}[t]{c}points\end{tabular}}}}%
    \put(0.14936563,0.07003736){\color[rgb]{0.14117647,0.19215686,0.23529412}\makebox(0,0)[t]{\lineheight{1.25}\smash{\begin{tabular}[t]{c}\textbf{Docs}\end{tabular}}}}%
    \put(0.14936563,0.05504313){\color[rgb]{0.36862745,0.41176471,0.44705882}\makebox(0,0)[t]{\lineheight{1.25}\smash{\begin{tabular}[t]{c}manuals\end{tabular}}}}%
    \put(0.24279124,0.07003736){\color[rgb]{0.14117647,0.19215686,0.23529412}\makebox(0,0)[t]{\lineheight{1.25}\smash{\begin{tabular}[t]{c}\textbf{Users}\end{tabular}}}}%
    \put(0.24279124,0.05504313){\color[rgb]{0.36862745,0.41176471,0.44705882}\makebox(0,0)[t]{\lineheight{1.25}\smash{\begin{tabular}[t]{c}operators\end{tabular}}}}%
    \put(0,0){\includegraphics[width=\unitlength,page=6]{graphical_abstract_revised.pdf}}%
    \put(0.50000001,0.42874556){\color[rgb]{1,1,1}\makebox(0,0)[t]{\lineheight{1.25}\smash{\begin{tabular}[t]{c}\textbf{2. Application-method map and evidence base}\end{tabular}}}}%
    \put(0.36505191,0.37504692){\color[rgb]{0.35294118,0.4,0.43921569}\makebox(0,0)[lt]{\lineheight{1.25}\smash{\begin{tabular}[t]{l}Applications\end{tabular}}}}%
    \put(0.3995774,0.4031871){\color[rgb]{0.35294118,0.4,0.43921569}\makebox(0,0)[lt]{\lineheight{1.25}\smash{\begin{tabular}[t]{l}Methods\end{tabular}}}}%
    \put(0.46885814,0.4045241){\color[rgb]{0.35294118,0.4,0.43921569}\makebox(0,0)[t]{\lineheight{1.25}\smash{\begin{tabular}[t]{c}M1\end{tabular}}}}%
    \put(0.53575549,0.4045241){\color[rgb]{0.35294118,0.4,0.43921569}\makebox(0,0)[t]{\lineheight{1.25}\smash{\begin{tabular}[t]{c}M2\end{tabular}}}}%
    \put(0.60265284,0.4045241){\color[rgb]{0.35294118,0.4,0.43921569}\makebox(0,0)[t]{\lineheight{1.25}\smash{\begin{tabular}[t]{c}M3\end{tabular}}}}%
    \put(0.46885814,0.39183667){\color[rgb]{0.41960784,0.4627451,0.50196078}\makebox(0,0)[t]{\lineheight{1.25}\smash{\begin{tabular}[t]{c}prompt /\\supervised\end{tabular}}}}%
    \put(0.53575549,0.39183667){\color[rgb]{0.41960784,0.4627451,0.50196078}\makebox(0,0)[t]{\lineheight{1.25}\smash{\begin{tabular}[t]{c}grounded\\orchestration\end{tabular}}}}%
    \put(0.60265284,0.39183667){\color[rgb]{0.41960784,0.4627451,0.50196078}\makebox(0,0)[t]{\lineheight{1.25}\smash{\begin{tabular}[t]{c}multimodal\\input\end{tabular}}}}%
    \put(0,0){\includegraphics[width=\unitlength,page=7]{graphical_abstract_revised.pdf}}%
    \put(0.42964246,0.3572346){\color[rgb]{0.14117647,0.19215686,0.23529412}\makebox(0,0)[rt]{\lineheight{1.25}\smash{\begin{tabular}[t]{r}A1 FDD\end{tabular}}}}%
    \put(0.46885814,0.3572346){\color[rgb]{1,1,1}\makebox(0,0)[t]{\lineheight{1.25}\smash{\begin{tabular}[t]{c}3\end{tabular}}}}%
    \put(0.53575549,0.3572346){\color[rgb]{1,1,1}\makebox(0,0)[t]{\lineheight{1.25}\smash{\begin{tabular}[t]{c}6\end{tabular}}}}%
    \put(0.60265284,0.3572346){\color[rgb]{0.48627451,0.54117647,0.58823529}\makebox(0,0)[t]{\lineheight{1.25}\smash{\begin{tabular}[t]{c}0\end{tabular}}}}%
    \put(0,0){\includegraphics[width=\unitlength,page=8]{graphical_abstract_revised.pdf}}%
    \put(0.42964246,0.32263252){\color[rgb]{0.14117647,0.19215686,0.23529412}\makebox(0,0)[rt]{\lineheight{1.25}\smash{\begin{tabular}[t]{r}A2 Forecast\end{tabular}}}}%
    \put(0.46885814,0.32263252){\color[rgb]{0.14117647,0.19215686,0.23529412}\makebox(0,0)[t]{\lineheight{1.25}\smash{\begin{tabular}[t]{c}2\end{tabular}}}}%
    \put(0.53575549,0.32263252){\color[rgb]{0.14117647,0.19215686,0.23529412}\makebox(0,0)[t]{\lineheight{1.25}\smash{\begin{tabular}[t]{c}1\end{tabular}}}}%
    \put(0.60265284,0.32263252){\color[rgb]{0.48627451,0.54117647,0.58823529}\makebox(0,0)[t]{\lineheight{1.25}\smash{\begin{tabular}[t]{c}0\end{tabular}}}}%
    \put(0,0){\includegraphics[width=\unitlength,page=9]{graphical_abstract_revised.pdf}}%
    \put(0.42964246,0.28803044){\color[rgb]{0.14117647,0.19215686,0.23529412}\makebox(0,0)[rt]{\lineheight{1.25}\smash{\begin{tabular}[t]{r}A3 BEM\end{tabular}}}}%
    \put(0.46885814,0.28803044){\color[rgb]{1,1,1}\makebox(0,0)[t]{\lineheight{1.25}\smash{\begin{tabular}[t]{c}14\end{tabular}}}}%
    \put(0.53575549,0.28803044){\color[rgb]{1,1,1}\makebox(0,0)[t]{\lineheight{1.25}\smash{\begin{tabular}[t]{c}\textbf{15}\end{tabular}}}}%
    \put(0.60265284,0.28803044){\color[rgb]{0.14117647,0.19215686,0.23529412}\makebox(0,0)[t]{\lineheight{1.25}\smash{\begin{tabular}[t]{c}3\end{tabular}}}}%
    \put(0,0){\includegraphics[width=\unitlength,page=10]{graphical_abstract_revised.pdf}}%
    \put(0.42964246,0.25342837){\color[rgb]{0.14117647,0.19215686,0.23529412}\makebox(0,0)[rt]{\lineheight{1.25}\smash{\begin{tabular}[t]{r}A4 Control\end{tabular}}}}%
    \put(0.46885814,0.25342837){\color[rgb]{0.14117647,0.19215686,0.23529412}\makebox(0,0)[t]{\lineheight{1.25}\smash{\begin{tabular}[t]{c}3\end{tabular}}}}%
    \put(0.53575549,0.25342837){\color[rgb]{1,1,1}\makebox(0,0)[t]{\lineheight{1.25}\smash{\begin{tabular}[t]{c}10\end{tabular}}}}%
    \put(0.60265284,0.25342837){\color[rgb]{0.14117647,0.19215686,0.23529412}\makebox(0,0)[t]{\lineheight{1.25}\smash{\begin{tabular}[t]{c}2\end{tabular}}}}%
    \put(0,0){\includegraphics[width=\unitlength,page=11]{graphical_abstract_revised.pdf}}%
    \put(0.42964246,0.21882629){\color[rgb]{0.14117647,0.19215686,0.23529412}\makebox(0,0)[rt]{\lineheight{1.25}\smash{\begin{tabular}[t]{r}A5 Comfort\end{tabular}}}}%
    \put(0.46885814,0.21882629){\color[rgb]{0.14117647,0.19215686,0.23529412}\makebox(0,0)[t]{\lineheight{1.25}\smash{\begin{tabular}[t]{c}4\end{tabular}}}}%
    \put(0.53575549,0.21882629){\color[rgb]{1,1,1}\makebox(0,0)[t]{\lineheight{1.25}\smash{\begin{tabular}[t]{c}3\end{tabular}}}}%
    \put(0.60265284,0.21882629){\color[rgb]{0.48627451,0.54117647,0.58823529}\makebox(0,0)[t]{\lineheight{1.25}\smash{\begin{tabular}[t]{c}0\end{tabular}}}}%
    \put(0,0){\includegraphics[width=\unitlength,page=12]{graphical_abstract_revised.pdf}}%
    \put(0.50000001,0.18883782){\color[rgb]{0.14117647,0.19215686,0.23529412}\makebox(0,0)[t]{\lineheight{1.25}\smash{\begin{tabular}[t]{c}\textbf{Method families}\end{tabular}}}}%
    \put(0.3581315,0.17384359){\color[rgb]{0.33333333,0.38039216,0.41960784}\makebox(0,0)[lt]{\lineheight{1.25}\smash{\begin{tabular}[t]{l}M1: prompting and supervised adaptation\end{tabular}}}}%
    \put(0.3581315,0.16000276){\color[rgb]{0.33333333,0.38039216,0.41960784}\makebox(0,0)[lt]{\lineheight{1.25}\smash{\begin{tabular}[t]{l}M2: retrieval, tools, memory, and agentic workflows\end{tabular}}}}%
    \put(0.3581315,0.14616193){\color[rgb]{0.33333333,0.38039216,0.41960784}\makebox(0,0)[lt]{\lineheight{1.25}\smash{\begin{tabular}[t]{l}M3: multimodal inputs (images, plans, heatmaps)\end{tabular}}}}%
    \put(0,0){\includegraphics[width=\unitlength,page=13]{graphical_abstract_revised.pdf}}%
    \put(0.50000001,0.12886089){\color[rgb]{0.14117647,0.19215686,0.23529412}\makebox(0,0)[t]{\lineheight{1.25}\smash{\begin{tabular}[t]{c}\textbf{Evidence realism}\end{tabular}}}}%
    \put(0.36966552,0.10579284){\color[rgb]{0.35294118,0.4,0.43921569}\makebox(0,0)[lt]{\lineheight{1.25}\smash{\begin{tabular}[t]{l}Simulation only\end{tabular}}}}%
    \put(0,0){\includegraphics[width=\unitlength,page=14]{graphical_abstract_revised.pdf}}%
    \put(0.64186853,0.10809964){\color[rgb]{0.14117647,0.19215686,0.23529412}\makebox(0,0)[lt]{\lineheight{1.25}\smash{\begin{tabular}[t]{l}30\end{tabular}}}}%
    \put(0.36966552,0.08503159){\color[rgb]{0.35294118,0.4,0.43921569}\makebox(0,0)[lt]{\lineheight{1.25}\smash{\begin{tabular}[t]{l}Retrospective field\end{tabular}}}}%
    \put(0,0){\includegraphics[width=\unitlength,page=15]{graphical_abstract_revised.pdf}}%
    \put(0.60149944,0.0873384){\color[rgb]{0.14117647,0.19215686,0.23529412}\makebox(0,0)[lt]{\lineheight{1.25}\smash{\begin{tabular}[t]{l}23\end{tabular}}}}%
    \put(0.36966552,0.06427035){\color[rgb]{0.35294118,0.4,0.43921569}\makebox(0,0)[lt]{\lineheight{1.25}\smash{\begin{tabular}[t]{l}Lab study\end{tabular}}}}%
    \put(0,0){\includegraphics[width=\unitlength,page=16]{graphical_abstract_revised.pdf}}%
    \put(0.52422147,0.06657715){\color[rgb]{0.14117647,0.19215686,0.23529412}\makebox(0,0)[lt]{\lineheight{1.25}\smash{\begin{tabular}[t]{l}9\end{tabular}}}}%
    \put(0.36966552,0.0435091){\color[rgb]{0.35294118,0.4,0.43921569}\makebox(0,0)[lt]{\lineheight{1.25}\smash{\begin{tabular}[t]{l}Pilot deployment\end{tabular}}}}%
    \put(0,0){\includegraphics[width=\unitlength,page=17]{graphical_abstract_revised.pdf}}%
    \put(0.4965398,0.0458159){\color[rgb]{0.14117647,0.19215686,0.23529412}\makebox(0,0)[lt]{\lineheight{1.25}\smash{\begin{tabular}[t]{l}4\end{tabular}}}}%
    \put(0.50000001,0.01352063){\color[rgb]{0.33333333,0.38039216,0.41960784}\makebox(0,0)[t]{\lineheight{1.25}\smash{\begin{tabular}[t]{c}Dominant cell: A3 x M2 (15 papers).\end{tabular}}}}%
    \put(0.50000002,0.00198661){\color[rgb]{0.33333333,0.38039216,0.41960784}\makebox(0,0)[t]{\lineheight{1.25}\smash{\begin{tabular}[t]{c}A$\times$M cells show primary papers with assigned M1--M3 methods; evidence groups sum to $N = 66$.\end{tabular}}}}%
    \put(0,0){\includegraphics[width=\unitlength,page=18]{graphical_abstract_revised.pdf}}%
    \put(0.85063439,0.42874556){\color[rgb]{1,1,1}\makebox(0,0)[t]{\lineheight{1.25}\smash{\begin{tabular}[t]{c}\textbf{3. Deployment logic in the field}\end{tabular}}}}%
    \put(0,0){\includegraphics[width=\unitlength,page=19]{graphical_abstract_revised.pdf}}%
    \put(0.85063439,0.39645028){\color[rgb]{0.12156863,0.30980392,0.44705882}\makebox(0,0)[t]{\lineheight{1.25}\smash{\begin{tabular}[t]{c}\textbf{LLM semantic layer}\end{tabular}}}}%
    \put(0.85063439,0.38145605){\color[rgb]{0.21176471,0.31764706,0.38823529}\makebox(0,0)[t]{\lineheight{1.25}\smash{\begin{tabular}[t]{c}prompting - RAG - agents - multimodal fusion\end{tabular}}}}%
    \put(0.85063439,0.33647335){\color[rgb]{0.14117647,0.19215686,0.23529412}\makebox(0,0)[t]{\lineheight{1.25}\smash{\begin{tabular}[t]{c}\textbf{Verification and guardrails}\end{tabular}}}}%
    \put(0.85063439,0.32147912){\color[rgb]{0.33333333,0.38039216,0.41960784}\makebox(0,0)[t]{\lineheight{1.25}\smash{\begin{tabular}[t]{c}schema checks - units - retrieval grounding - review\end{tabular}}}}%
    \put(0.85063439,0.28341683){\color[rgb]{0.14117647,0.19215686,0.23529412}\makebox(0,0)[t]{\lineheight{1.25}\smash{\begin{tabular}[t]{c}\textbf{Operator or physics layer}\end{tabular}}}}%
    \put(0.85063439,0.2684226){\color[rgb]{0.33333333,0.38039216,0.41960784}\makebox(0,0)[t]{\lineheight{1.25}\smash{\begin{tabular}[t]{c}human review, MPC, RL, simulation, BAS tools\end{tabular}}}}%
    \put(0.85063439,0.20959907){\color[rgb]{0.14117647,0.19215686,0.23529412}\makebox(0,0)[t]{\lineheight{1.25}\smash{\begin{tabular}[t]{c}\textbf{Plant and actuation}\end{tabular}}}}%
    \put(0.85063439,0.19460484){\color[rgb]{0.33333333,0.38039216,0.41960784}\makebox(0,0)[t]{\lineheight{1.25}\smash{\begin{tabular}[t]{c}bounded execution in the real HVAC system\end{tabular}}}}%
    \put(0,0){\includegraphics[width=\unitlength,page=20]{graphical_abstract_revised.pdf}}%
    \put(0.98558249,0.23612733){\color[rgb]{0.40784314,0.45098039,0.49019608}\makebox(0,0)[rt]{\lineheight{1.25}\smash{\begin{tabular}[t]{r}responsibility boundary\end{tabular}}}}%
    \put(0,0){\includegraphics[width=\unitlength,page=21]{graphical_abstract_revised.pdf}}%
    \put(0.85063439,0.11271325){\color[rgb]{0.14117647,0.19215686,0.23529412}\makebox(0,0)[t]{\lineheight{1.25}\smash{\begin{tabular}[t]{c}\textbf{Observed trends and next steps}\end{tabular}}}}%
    \put(0.71107268,0.09771902){\color[rgb]{0.33333333,0.38039216,0.41960784}\makebox(0,0)[lt]{\lineheight{1.25}\smash{\begin{tabular}[t]{l}Trends: agentic systems rising; multimodal still sparse\\63 of 66 papers remain research-only\end{tabular}}}}%
    \put(0.71107268,0.06311694){\color[rgb]{0.33333333,0.38039216,0.41960784}\makebox(0,0)[lt]{\lineheight{1.25}\smash{\begin{tabular}[t]{l}Future work: context-grounded evaluation under\end{tabular}}}}%
    \put(0.71107268,0.04927611){\color[rgb]{0.33333333,0.38039216,0.41960784}\makebox(0,0)[lt]{\lineheight{1.25}\smash{\begin{tabular}[t]{l}governance constraints, physics-integrated control\end{tabular}}}}%
    \put(0,0){\includegraphics[width=\unitlength,page=22]{graphical_abstract_revised.pdf}}%
    \put(0.93690889,0.15538915){\color[rgb]{0.14117647,0.19215686,0.23529412}\makebox(0,0)[t]{\lineheight{1.25}\smash{\begin{tabular}[t]{c}\textbf{Ready now}\end{tabular}}}}%
    \put(0.93690889,0.14385512){\color[rgb]{0.33333333,0.38039216,0.41960784}\makebox(0,0)[t]{\lineheight{1.25}\smash{\begin{tabular}[t]{c}0 papers\end{tabular}}}}%
    \put(0.85063439,0.15538915){\color[rgb]{0.14117647,0.19215686,0.23529412}\makebox(0,0)[t]{\lineheight{1.25}\smash{\begin{tabular}[t]{c}\textbf{Near-term}\end{tabular}}}}%
    \put(0.85063439,0.14385512){\color[rgb]{0.33333333,0.38039216,0.41960784}\makebox(0,0)[t]{\lineheight{1.25}\smash{\begin{tabular}[t]{c}3 papers\end{tabular}}}}%
    \put(0.7652826,0.15538915){\color[rgb]{0.14117647,0.19215686,0.23529412}\makebox(0,0)[t]{\lineheight{1.25}\smash{\begin{tabular}[t]{c}\textbf{Research-only}\end{tabular}}}}%
    \put(0.7652826,0.14385512){\color[rgb]{0.33333333,0.38039216,0.41960784}\makebox(0,0)[t]{\lineheight{1.25}\smash{\begin{tabular}[t]{c}63 papers\end{tabular}}}}%
  \end{picture}%
\endgroup%

%% file: figures/llm_semantic_workflow_layer.pdf_tex
\begingroup%
  \makeatletter%
  \providecommand\color[2][]{%
    \errmessage{(Inkscape) Color is used for the text in Inkscape, but the package 'color.sty' is not loaded}%
    \renewcommand\color[2][]{}%
  }%
  \providecommand\transparent[1]{%
    \errmessage{(Inkscape) Transparency is used (non-zero) for the text in Inkscape, but the package 'transparent.sty' is not loaded}%
    \renewcommand\transparent[1]{}%
  }%
  \providecommand\rotatebox[2]{#2}%
  \newcommand*\fsize{\dimexpr\f@size pt\relax}%
  \newcommand*\lineheight[1]{\fontsize{\fsize}{#1\fsize}\selectfont}%
  \ifx\svgwidth\undefined%
    \setlength{\unitlength}{510.23622047bp}%
    \ifx\svgscale\undefined%
      \relax%
    \else%
      \setlength{\unitlength}{\unitlength * \real{\svgscale}}%
    \fi%
  \else%
    \setlength{\unitlength}{\svgwidth}%
  \fi%
  \global\let\svgwidth\undefined%
  \global\let\svgscale\undefined%
  \makeatother%
  \begin{picture}(1,0.60000001)%
    \lineheight{1}%
    \setlength\tabcolsep{0pt}%
    \put(0,0){\includegraphics[width=\unitlength,page=1]{llm_semantic_workflow_layer.pdf}}%
    \put(0.16062483,0.57044747){\color[rgb]{1,1,1}\makebox(0,0)[t]{\lineheight{1.25}\smash{\begin{tabular}[t]{c}\textbf{Fragmented source data}\end{tabular}}}}%
    \put(0,0){\includegraphics[width=\unitlength,page=2]{llm_semantic_workflow_layer.pdf}}%
    \put(0.06973696,0.50736377){\color[rgb]{0.14117647,0.19215686,0.23529412}\makebox(0,0)[lt]{\lineheight{1.25}\smash{\begin{tabular}[t]{l}\textbf{BAS}\\\textbf{points}\end{tabular}}}}%
    \put(0.06973696,0.47713604){\color[rgb]{0.2745098,0.31764706,0.35294118}\makebox(0,0)[lt]{\lineheight{1.25}\smash{\begin{tabular}[t]{l}tags, units\end{tabular}}}}%
    \put(0,0){\includegraphics[width=\unitlength,page=3]{llm_semantic_workflow_layer.pdf}}%
    \put(0.20232634,0.50736377){\color[rgb]{0.14117647,0.19215686,0.23529412}\makebox(0,0)[lt]{\lineheight{1.25}\smash{\begin{tabular}[t]{l}\textbf{Time-series}\end{tabular}}}}%
    \put(0.20232634,0.49518972){\color[rgb]{0.14117647,0.19215686,0.23529412}\makebox(0,0)[lt]{\lineheight{1.25}\smash{\begin{tabular}[t]{l}\textbf{sensor data}\end{tabular}}}}%
    \put(0.20232634,0.48007586){\color[rgb]{0.2745098,0.31764706,0.35294118}\makebox(0,0)[lt]{\lineheight{1.25}\smash{\begin{tabular}[t]{l}trends and\\alarms\end{tabular}}}}%
    \put(0,0){\includegraphics[width=\unitlength,page=4]{llm_semantic_workflow_layer.pdf}}%
    \put(0.06973696,0.43210603){\color[rgb]{0.14117647,0.19215686,0.23529412}\makebox(0,0)[lt]{\lineheight{1.25}\smash{\begin{tabular}[t]{l}\textbf{Manuals}\end{tabular}}}}%
    \put(0.06973696,0.40775795){\color[rgb]{0.2745098,0.31764706,0.35294118}\makebox(0,0)[lt]{\lineheight{1.25}\smash{\begin{tabular}[t]{l}specs, SOPs\end{tabular}}}}%
    \put(0,0){\includegraphics[width=\unitlength,page=5]{llm_semantic_workflow_layer.pdf}}%
    \put(0.20232634,0.43172599){\color[rgb]{0.14117647,0.19215686,0.23529412}\makebox(0,0)[lt]{\lineheight{1.25}\smash{\begin{tabular}[t]{l}\textbf{BIM /}\\\textbf{topology}\end{tabular}}}}%
    \put(0.20232634,0.40263762){\color[rgb]{0.2745098,0.31764706,0.35294118}\makebox(0,0)[lt]{\lineheight{1.25}\smash{\begin{tabular}[t]{l}spaces and\\topology\end{tabular}}}}%
    \put(0,0){\includegraphics[width=\unitlength,page=6]{llm_semantic_workflow_layer.pdf}}%
    \put(0.06973696,0.3568483){\color[rgb]{0.14117647,0.19215686,0.23529412}\makebox(0,0)[lt]{\lineheight{1.25}\smash{\begin{tabular}[t]{l}\textbf{Semantic}\end{tabular}}}}%
    \put(0.06973696,0.34467424){\color[rgb]{0.14117647,0.19215686,0.23529412}\makebox(0,0)[lt]{\lineheight{1.25}\smash{\begin{tabular}[t]{l}\textbf{ontologies}\end{tabular}}}}%
    \put(0.06973696,0.32956003){\color[rgb]{0.2745098,0.31764706,0.35294118}\makebox(0,0)[lt]{\lineheight{1.25}\smash{\begin{tabular}[t]{l}Brick,\\Haystack\end{tabular}}}}%
    \put(0,0){\includegraphics[width=\unitlength,page=7]{llm_semantic_workflow_layer.pdf}}%
    \put(0.20232634,0.35646791){\color[rgb]{0.14117647,0.19215686,0.23529412}\makebox(0,0)[lt]{\lineheight{1.25}\smash{\begin{tabular}[t]{l}\textbf{Maintenance}\\\textbf{logs}\end{tabular}}}}%
    \put(0.20232634,0.32738134){\color[rgb]{0.2745098,0.31764706,0.35294118}\makebox(0,0)[lt]{\lineheight{1.25}\smash{\begin{tabular}[t]{l}work orders,\\notes\end{tabular}}}}%
    \put(0,0){\includegraphics[width=\unitlength,page=8]{llm_semantic_workflow_layer.pdf}}%
    \put(0.06973696,0.28121017){\color[rgb]{0.14117647,0.19215686,0.23529412}\makebox(0,0)[lt]{\lineheight{1.25}\smash{\begin{tabular}[t]{l}\textbf{Operator}\\\textbf{input}\end{tabular}}}}%
    \put(0.06973696,0.24918206){\color[rgb]{0.2745098,0.31764706,0.35294118}\makebox(0,0)[lt]{\lineheight{1.25}\smash{\begin{tabular}[t]{l}questions,\\intents\end{tabular}}}}%
    \put(0,0){\includegraphics[width=\unitlength,page=9]{llm_semantic_workflow_layer.pdf}}%
    \put(0.20232634,0.27865074){\color[rgb]{0.14117647,0.19215686,0.23529412}\makebox(0,0)[lt]{\lineheight{1.25}\smash{\begin{tabular}[t]{l}\textbf{Tools and}\end{tabular}}}}%
    \put(0.20232634,0.26647668){\color[rgb]{0.14117647,0.19215686,0.23529412}\makebox(0,0)[lt]{\lineheight{1.25}\smash{\begin{tabular}[t]{l}\textbf{databases}\end{tabular}}}}%
    \put(0.20232634,0.24842267){\color[rgb]{0.2745098,0.31764706,0.35294118}\makebox(0,0)[lt]{\lineheight{1.25}\smash{\begin{tabular}[t]{l}simulation\\tools\end{tabular}}}}%
    \put(0,0){\includegraphics[width=\unitlength,page=10]{llm_semantic_workflow_layer.pdf}}%
    \put(0.06439061,0.20743955){\color[rgb]{0.14117647,0.19215686,0.23529412}\makebox(0,0)[lt]{\lineheight{1.25}\smash{\begin{tabular}[t]{l}\textbf{Optional multimodal sources}\end{tabular}}}}%
    \put(0.06439061,0.19637223){\color[rgb]{0.36862745,0.41176471,0.44705882}\makebox(0,0)[lt]{\lineheight{1.25}\smash{\begin{tabular}[t]{l}thermal images, floor plans, camera\end{tabular}}}}%
    \put(0,0){\includegraphics[width=\unitlength,page=11]{llm_semantic_workflow_layer.pdf}}%
    \put(0.46322796,0.49740319){\color[rgb]{0.12156863,0.30980392,0.44705882}\makebox(0,0)[t]{\lineheight{1.25}\smash{\begin{tabular}[t]{c}\textbf{LLM semantic and}\\\textbf{workflow layer}\end{tabular}}}}%
    \put(0,0){\includegraphics[width=\unitlength,page=12]{llm_semantic_workflow_layer.pdf}}%
    \put(0.46322796,0.40599116){\color[rgb]{0.19215686,0.25098039,0.30980392}\makebox(0,0)[t]{\lineheight{1.25}\smash{\begin{tabular}[t]{c}Interprets, aligns, grounds, and routes\end{tabular}}}}%
    \put(0.46322796,0.3938171){\color[rgb]{0.19215686,0.25098039,0.30980392}\makebox(0,0)[t]{\lineheight{1.25}\smash{\begin{tabular}[t]{c}heterogeneous building context.\end{tabular}}}}%
    \put(0,0){\includegraphics[width=\unitlength,page=13]{llm_semantic_workflow_layer.pdf}}%
    \put(0.40347371,0.37499367){\color[rgb]{0.25490196,0.31764706,0.37254902}\makebox(0,0)[t]{\lineheight{1.25}\smash{\begin{tabular}[t]{c}semantic reasoning\end{tabular}}}}%
    \put(0.52356999,0.37498254){\color[rgb]{0.25490196,0.31764706,0.37254902}\makebox(0,0)[t]{\lineheight{1.25}\smash{\begin{tabular}[t]{c}context grounding\end{tabular}}}}%
    \put(0.40347371,0.34734915){\color[rgb]{0.25490196,0.31764706,0.37254902}\makebox(0,0)[t]{\lineheight{1.25}\smash{\begin{tabular}[t]{c}orchestration\end{tabular}}}}%
    \put(0.52356999,0.34736726){\color[rgb]{0.25490196,0.31764706,0.37254902}\makebox(0,0)[t]{\lineheight{1.25}\smash{\begin{tabular}[t]{c}multimodal fusion\end{tabular}}}}%
    \put(0,0){\includegraphics[width=\unitlength,page=14]{llm_semantic_workflow_layer.pdf}}%
    \put(0.68884369,0.50293684){\color[rgb]{1,1,1}\makebox(0,0)[t]{\lineheight{1.25}\smash{\begin{tabular}[t]{c}\textbf{Enabled tasks}\end{tabular}}}}%
    \put(0,0){\includegraphics[width=\unitlength,page=15]{llm_semantic_workflow_layer.pdf}}%
    \put(0.68984138,0.46749667){\color[rgb]{0.25490196,0.31764706,0.37254902}\makebox(0,0)[t]{\lineheight{1.25}\smash{\begin{tabular}[t]{c}point names\end{tabular}}}}%
    \put(0.68984138,0.44205226){\color[rgb]{0.25490196,0.31764706,0.37254902}\makebox(0,0)[t]{\lineheight{1.25}\smash{\begin{tabular}[t]{c}grounded QA\end{tabular}}}}%
    \put(0.68984138,0.41660793){\color[rgb]{0.25490196,0.31764706,0.37254902}\makebox(0,0)[t]{\lineheight{1.25}\smash{\begin{tabular}[t]{c}fault explanation\end{tabular}}}}%
    \put(0.68984138,0.39116354){\color[rgb]{0.25490196,0.31764706,0.37254902}\makebox(0,0)[t]{\lineheight{1.25}\smash{\begin{tabular}[t]{c}model support\end{tabular}}}}%
    \put(0.68984138,0.36571919){\color[rgb]{0.25490196,0.31764706,0.37254902}\makebox(0,0)[t]{\lineheight{1.25}\smash{\begin{tabular}[t]{c}control advice\end{tabular}}}}%
    \put(0.68984138,0.3402748){\color[rgb]{0.25490196,0.31764706,0.37254902}\makebox(0,0)[t]{\lineheight{1.25}\smash{\begin{tabular}[t]{c}comfort interaction\end{tabular}}}}%
    \put(0,0){\includegraphics[width=\unitlength,page=16]{llm_semantic_workflow_layer.pdf}}%
    \put(0.88451997,0.57044747){\color[rgb]{1,1,1}\makebox(0,0)[t]{\lineheight{1.25}\smash{\begin{tabular}[t]{c}\textbf{HVAC application families}\end{tabular}}}}%
    \put(0,0){\includegraphics[width=\unitlength,page=17]{llm_semantic_workflow_layer.pdf}}%
    \put(0.84281846,0.50293684){\color[rgb]{0.12156863,0.30980392,0.44705882}\makebox(0,0)[lt]{\lineheight{1.25}\smash{\begin{tabular}[t]{l}\textbf{A1}\end{tabular}}}}%
    \put(0.86848095,0.50736377){\color[rgb]{0.14117647,0.19215686,0.23529412}\makebox(0,0)[lt]{\lineheight{1.25}\smash{\begin{tabular}[t]{l}\textbf{Fault detection}\end{tabular}}}}%
    \put(0.86848095,0.49518972){\color[rgb]{0.14117647,0.19215686,0.23529412}\makebox(0,0)[lt]{\lineheight{1.25}\smash{\begin{tabular}[t]{l}\textbf{and diagnosis}\end{tabular}}}}%
    \put(0,0){\includegraphics[width=\unitlength,page=18]{llm_semantic_workflow_layer.pdf}}%
    \put(0.84281846,0.44428008){\color[rgb]{0.12156863,0.30980392,0.44705882}\makebox(0,0)[lt]{\lineheight{1.25}\smash{\begin{tabular}[t]{l}\textbf{A2}\end{tabular}}}}%
    \put(0.86848095,0.44428008){\color[rgb]{0.14117647,0.19215686,0.23529412}\makebox(0,0)[lt]{\lineheight{1.25}\smash{\begin{tabular}[t]{l}\textbf{Load forecasting}\end{tabular}}}}%
    \put(0,0){\includegraphics[width=\unitlength,page=19]{llm_semantic_workflow_layer.pdf}}%
    \put(0.84281846,0.38562329){\color[rgb]{0.12156863,0.30980392,0.44705882}\makebox(0,0)[lt]{\lineheight{1.25}\smash{\begin{tabular}[t]{l}\textbf{A3}\end{tabular}}}}%
    \put(0.86848095,0.38562329){\color[rgb]{0.14117647,0.19215686,0.23529412}\makebox(0,0)[lt]{\lineheight{1.25}\smash{\begin{tabular}[t]{l}\textbf{BEM and}\\\textbf{simulation}\end{tabular}}}}%
    \put(0,0){\includegraphics[width=\unitlength,page=20]{llm_semantic_workflow_layer.pdf}}%
    \put(0.84281846,0.32696652){\color[rgb]{0.12156863,0.30980392,0.44705882}\makebox(0,0)[lt]{\lineheight{1.25}\smash{\begin{tabular}[t]{l}\textbf{A4}\end{tabular}}}}%
    \put(0.86848095,0.32696652){\color[rgb]{0.14117647,0.19215686,0.23529412}\makebox(0,0)[lt]{\lineheight{1.25}\smash{\begin{tabular}[t]{l}\textbf{Control and}\\\textbf{optimisation}\end{tabular}}}}%
    \put(0,0){\includegraphics[width=\unitlength,page=21]{llm_semantic_workflow_layer.pdf}}%
    \put(0.84281846,0.26830977){\color[rgb]{0.12156863,0.30980392,0.44705882}\makebox(0,0)[lt]{\lineheight{1.25}\smash{\begin{tabular}[t]{l}\textbf{A5}\end{tabular}}}}%
    \put(0.86848095,0.26723743){\color[rgb]{0.14117647,0.19215686,0.23529412}\makebox(0,0)[lt]{\lineheight{1.25}\smash{\begin{tabular}[t]{l}\textbf{Thermal comfort}\end{tabular}}}}%
    \put(0.86848095,0.25506343){\color[rgb]{0.14117647,0.19215686,0.23529412}\makebox(0,0)[lt]{\lineheight{1.25}\smash{\begin{tabular}[t]{l}\textbf{and interaction}\end{tabular}}}}%
    \put(0,0){\includegraphics[width=\unitlength,page=22]{llm_semantic_workflow_layer.pdf}}%
    \put(0.43366134,0.2309942){\color[rgb]{0.14117647,0.19215686,0.23529412}\makebox(0,0)[lt]{\lineheight{1.25}\smash{\begin{tabular}[t]{l}\textbf{Physical HVAC control layer}\end{tabular}}}}%
    \put(0.43366134,0.21882012){\color[rgb]{0.36862745,0.41176471,0.44705882}\makebox(0,0)[lt]{\lineheight{1.25}\smash{\begin{tabular}[t]{l}BAS logic, MPC, schedules, plant actuation\end{tabular}}}}%
    \put(0.43366134,0.20775276){\color[rgb]{0.36862745,0.41176471,0.44705882}\makebox(0,0)[lt]{\lineheight{1.25}\smash{\begin{tabular}[t]{l}guarded handoff; no unrestricted LLM autonomy.\end{tabular}}}}%
    \put(0,0){\includegraphics[width=\unitlength,page=23]{llm_semantic_workflow_layer.pdf}}%
    \put(0.5006524,0.15874336){\color[rgb]{1,1,1}\makebox(0,0)[t]{\lineheight{1.25}\smash{\begin{tabular}[t]{c}\textbf{Verification and deployment guardrails}\end{tabular}}}}%
    \put(0,0){\includegraphics[width=\unitlength,page=24]{llm_semantic_workflow_layer.pdf}}%
    \put(0.11571552,0.12332796){\color[rgb]{0.25490196,0.31764706,0.37254902}\makebox(0,0)[t]{\lineheight{1.25}\smash{\begin{tabular}[t]{c}retrieval grounding\end{tabular}}}}%
    \put(0.29962979,0.12332796){\color[rgb]{0.25490196,0.31764706,0.37254902}\makebox(0,0)[t]{\lineheight{1.25}\smash{\begin{tabular}[t]{c}schema and unit checks\end{tabular}}}}%
    \put(0.48354405,0.12332796){\color[rgb]{0.25490196,0.31764706,0.37254902}\makebox(0,0)[t]{\lineheight{1.25}\smash{\begin{tabular}[t]{c}human review\end{tabular}}}}%
    \put(0.6674583,0.12332796){\color[rgb]{0.25490196,0.31764706,0.37254902}\makebox(0,0)[t]{\lineheight{1.25}\smash{\begin{tabular}[t]{c}physics / MPC verification\end{tabular}}}}%
    \put(0.86848095,0.12332796){\color[rgb]{0.25490196,0.31764706,0.37254902}\makebox(0,0)[t]{\lineheight{1.25}\smash{\begin{tabular}[t]{c}bounded execution in BAS\end{tabular}}}}%
    \put(0,0){\includegraphics[width=\unitlength,page=25]{llm_semantic_workflow_layer.pdf}}%
    \put(0.50065237,0.07791753){\color[rgb]{0.25882353,0.31372549,0.35686275}\makebox(0,0)[t]{\lineheight{1.25}\smash{\begin{tabular}[t]{c}\textbf{Corpus method transition}\end{tabular}}}}%
    \put(0,0){\includegraphics[width=\unitlength,page=26]{llm_semantic_workflow_layer.pdf}}%
    \put(0.17987167,0.06020983){\color[rgb]{0.14117647,0.19215686,0.23529412}\makebox(0,0)[t]{\lineheight{1.25}\smash{\begin{tabular}[t]{c}\textbf{Beginning}\end{tabular}}}}%
    \put(0.17987167,0.02811461){\color[rgb]{0.36862745,0.41176471,0.44705882}\makebox(0,0)[t]{\lineheight{1.25}\smash{\begin{tabular}[t]{c}M1 prompting and fine-tuning\end{tabular}}}}%
    \put(0.17987167,0.01483383){\color[rgb]{0.41568627,0.4627451,0.50196078}\makebox(0,0)[t]{\lineheight{1.25}\smash{\begin{tabular}[t]{c}isolated tasks\end{tabular}}}}%
    \put(0,0){\includegraphics[width=\unitlength,page=27]{llm_semantic_workflow_layer.pdf}}%
    \put(0.50065237,0.06020983){\color[rgb]{0.14117647,0.19215686,0.23529412}\makebox(0,0)[t]{\lineheight{1.25}\smash{\begin{tabular}[t]{c}\textbf{Current frontier}\end{tabular}}}}%
    \put(0.50065237,0.02811461){\color[rgb]{0.36862745,0.41176471,0.44705882}\makebox(0,0)[t]{\lineheight{1.25}\smash{\begin{tabular}[t]{c}retrieval, tools, workflows\end{tabular}}}}%
    \put(0.50065237,0.01483383){\color[rgb]{0.41568627,0.4627451,0.50196078}\makebox(0,0)[t]{\lineheight{1.25}\smash{\begin{tabular}[t]{c}retrieval, tools, workflow integration\end{tabular}}}}%
    \put(0,0){\includegraphics[width=\unitlength,page=28]{llm_semantic_workflow_layer.pdf}}%
    \put(0.82143308,0.06020983){\color[rgb]{0.14117647,0.19215686,0.23529412}\makebox(0,0)[t]{\lineheight{1.25}\smash{\begin{tabular}[t]{c}\textbf{Future direction}\end{tabular}}}}%
    \put(0.82143308,0.02811461){\color[rgb]{0.36862745,0.41176471,0.44705882}\makebox(0,0)[t]{\lineheight{1.25}\smash{\begin{tabular}[t]{c}governed integration\end{tabular}}}}%
  \end{picture}%
\endgroup%

%% file: figures/methods_overview_revised.pdf_tex
\begingroup%
  \makeatletter%
  \providecommand\color[2][]{%
    \errmessage{(Inkscape) Color is used for the text in Inkscape, but the package 'color.sty' is not loaded}%
    \renewcommand\color[2][]{}%
  }%
  \providecommand\transparent[1]{%
    \errmessage{(Inkscape) Transparency is used (non-zero) for the text in Inkscape, but the package 'transparent.sty' is not loaded}%
    \renewcommand\transparent[1]{}%
  }%
  \providecommand\rotatebox[2]{#2}%
  \newcommand*\fsize{\dimexpr\f@size pt\relax}%
  \newcommand*\lineheight[1]{\fontsize{\fsize}{#1\fsize}\selectfont}%
  \ifx\svgwidth\undefined%
    \setlength{\unitlength}{558.6665176bp}%
    \ifx\svgscale\undefined%
      \relax%
    \else%
      \setlength{\unitlength}{\unitlength * \real{\svgscale}}%
    \fi%
  \else%
    \setlength{\unitlength}{\svgwidth}%
  \fi%
  \global\let\svgwidth\undefined%
  \global\let\svgscale\undefined%
  \makeatother%
  \begin{picture}(1,0.33495022)%
    \lineheight{1}%
    \setlength\tabcolsep{0pt}%
    \put(0,0){\includegraphics[width=\unitlength,page=1]{methods_overview_revised.pdf}}%
    \put(0.15454432,0.31588483){\color[rgb]{1,1,1}\makebox(0,0)[t]{\lineheight{1.25}\smash{\begin{tabular}[t]{c}\textbf{1. Review scope and search}\end{tabular}}}}%
    \put(0,0){\includegraphics[width=\unitlength,page=2]{methods_overview_revised.pdf}}%
    \put(0.0351276,0.27589046){\color[rgb]{0.14117647,0.19215686,0.23529412}\makebox(0,0)[lt]{\lineheight{1.25}\smash{\begin{tabular}[t]{l}\textbf{Background framing}\end{tabular}}}}%
    \put(0.0351276,0.25806983){\color[rgb]{0.30196078,0.34509804,0.38039216}\makebox(0,0)[lt]{\lineheight{1.25}\smash{\begin{tabular}[t]{l}HVAC bottlenecks and LLM methods\end{tabular}}}}%
    \put(0.0351276,0.24071891){\color[rgb]{0.30196078,0.34509804,0.38039216}\makebox(0,0)[lt]{\lineheight{1.25}\smash{\begin{tabular}[t]{l}Sec.~\ref{sec:background}, \ref{sec:background_hvac},\ref{sec:background_methods}\end{tabular}}}}%
    \put(0,0){\includegraphics[width=\unitlength,page=3]{methods_overview_revised.pdf}}%
    \put(0.0351276,0.19292858){\color[rgb]{0.12156863,0.30980392,0.44705882}\makebox(0,0)[lt]{\lineheight{1.25}\smash{\begin{tabular}[t]{l}\textbf{Search strategy and corpus}\end{tabular}}}}%
    \put(0.0351276,0.17242298){\color[rgb]{0.25490196,0.31764706,0.37254902}\makebox(0,0)[lt]{\lineheight{1.25}\smash{\begin{tabular}[t]{l}Scopus query, year window, screening\end{tabular}}}}%
    \put(0.0351276,0.15507216){\color[rgb]{0.25490196,0.31764706,0.37254902}\makebox(0,0)[lt]{\lineheight{1.25}\smash{\begin{tabular}[t]{l}logic, and PRISMA flow\end{tabular}}}}%
    \put(0.0351276,0.13772127){\color[rgb]{0.25490196,0.31764706,0.37254902}\makebox(0,0)[lt]{\lineheight{1.25}\smash{\begin{tabular}[t]{l}Sec.~\ref{sec:method_search}\end{tabular}}}}%
    \put(0,0){\includegraphics[width=\unitlength,page=4]{methods_overview_revised.pdf}}%
    \put(0.05645027,0.091978){\color[rgb]{0.14117647,0.19215686,0.23529412}\makebox(0,0)[t]{\lineheight{1.25}\smash{\begin{tabular}[t]{c}\textbf{Records}\end{tabular}}}}%
    \put(0.05645027,0.0746271){\color[rgb]{0.37254902,0.41568627,0.45098039}\makebox(0,0)[t]{\lineheight{1.25}\smash{\begin{tabular}[t]{c}142\end{tabular}}}}%
    \put(0.14311664,0.091978){\color[rgb]{0.14117647,0.19215686,0.23529412}\makebox(0,0)[t]{\lineheight{1.25}\smash{\begin{tabular}[t]{c}\textbf{Full text}\end{tabular}}}}%
    \put(0.14311664,0.0746271){\color[rgb]{0.37254902,0.41568627,0.45098039}\makebox(0,0)[t]{\lineheight{1.25}\smash{\begin{tabular}[t]{c}104\end{tabular}}}}%
    \put(0.23047087,0.091978){\color[rgb]{0.14117647,0.19215686,0.23529412}\makebox(0,0)[t]{\lineheight{1.25}\smash{\begin{tabular}[t]{c}\textbf{Corpus}\end{tabular}}}}%
    \put(0.23047087,0.0746271){\color[rgb]{0.37254902,0.41568627,0.45098039}\makebox(0,0)[t]{\lineheight{1.25}\smash{\begin{tabular}[t]{c}66\end{tabular}}}}%
    \put(0.14380447,0.03851609){\color[rgb]{0.4,0.44705882,0.48627451}\makebox(0,0)[t]{\lineheight{1.25}\smash{\begin{tabular}[t]{c}Positioning against prior reviews:\end{tabular}}}}%
    \put(0.14380447,0.02116517){\color[rgb]{0.4,0.44705882,0.48627451}\makebox(0,0)[t]{\lineheight{1.25}\smash{\begin{tabular}[t]{c}Sec.~\ref{sec:background_reviews}\end{tabular}}}}%
    \put(0,0){\includegraphics[width=\unitlength,page=5]{methods_overview_revised.pdf}}%
    \put(0.50341363,0.31318743){\color[rgb]{1,1,1}\makebox(0,0)[t]{\lineheight{1.25}\smash{\begin{tabular}[t]{c}\textbf{2. Taxonomy and coding procedure}\end{tabular}}}}%
    \put(0,0){\includegraphics[width=\unitlength,page=6]{methods_overview_revised.pdf}}%
    \put(0.37310581,0.27586968){\color[rgb]{0.14117647,0.19215686,0.23529412}\makebox(0,0)[lt]{\lineheight{1.25}\smash{\begin{tabular}[t]{l}\textbf{Taxonomy definition}\end{tabular}}}}%
    \put(0.37310581,0.2553677){\color[rgb]{0.30196078,0.34509804,0.38039216}\makebox(0,0)[lt]{\lineheight{1.25}\smash{\begin{tabular}[t]{l}Application families A1--A5\end{tabular}}}}%
    \put(0.37310581,0.23801987){\color[rgb]{0.30196078,0.34509804,0.38039216}\makebox(0,0)[lt]{\lineheight{1.25}\smash{\begin{tabular}[t]{l}Method families M1--M3\end{tabular}}}}%
    \put(0.37310581,0.22067205){\color[rgb]{0.30196078,0.34509804,0.38039216}\makebox(0,0)[lt]{\lineheight{1.25}\smash{\begin{tabular}[t]{l}Sec.~\ref{sec:method_coding}\end{tabular}}}}%
    \put(0,0){\includegraphics[width=\unitlength,page=7]{methods_overview_revised.pdf}}%
    \put(0.37829491,0.16879816){\color[rgb]{0.14117647,0.19215686,0.23529412}\makebox(0,0)[t]{\lineheight{1.25}\smash{\begin{tabular}[t]{c}\textbf{Evidence}\end{tabular}}}}%
    \put(0.37829491,0.14987328){\color[rgb]{0.37254902,0.41568627,0.45098039}\makebox(0,0)[t]{\lineheight{1.25}\smash{\begin{tabular}[t]{c}Conceptual\\to\end{tabular}}}}%
    \put(0.37829491,0.12604763){\color[rgb]{0.37254902,0.41568627,0.45098039}\makebox(0,0)[t]{\lineheight{1.25}\smash{\begin{tabular}[t]{c}deployment\end{tabular}}}}%
    \put(0.46333371,0.16879816){\color[rgb]{0.14117647,0.19215686,0.23529412}\makebox(0,0)[t]{\lineheight{1.25}\smash{\begin{tabular}[t]{c}\textbf{Readiness}\end{tabular}}}}%
    \put(0.46333371,0.15384737){\color[rgb]{0.37254902,0.41568627,0.45098039}\makebox(0,0)[t]{\lineheight{1.25}\smash{\begin{tabular}[t]{c}ReadyNow,\end{tabular}}}}%
    \put(0.46333289,0.13937674){\color[rgb]{0.37254902,0.41568627,0.45098039}\makebox(0,0)[t]{\lineheight{1.25}\smash{\begin{tabular}[t]{c}NearTerm,\\ResearchOnly\end{tabular}}}}%
    \put(0.54635393,0.16879816){\color[rgb]{0.14117647,0.19215686,0.23529412}\makebox(0,0)[t]{\lineheight{1.25}\smash{\begin{tabular}[t]{c}\textbf{Safety}\end{tabular}}}}%
    \put(0.54635393,0.15255824){\color[rgb]{0.37254902,0.41568627,0.45098039}\makebox(0,0)[t]{\lineheight{1.25}\smash{\begin{tabular}[t]{c}explicit /\end{tabular}}}}%
    \put(0.54635393,0.12604763){\color[rgb]{0.37254902,0.41568627,0.45098039}\makebox(0,0)[t]{\lineheight{1.25}\smash{\begin{tabular}[t]{c}not explicit\end{tabular}}}}%
    \put(0.62254546,0.16879816){\color[rgb]{0.14117647,0.19215686,0.23529412}\makebox(0,0)[t]{\lineheight{1.25}\smash{\begin{tabular}[t]{c}\textbf{Boundary}\end{tabular}}}}%
    \put(0.62254546,0.15255824){\color[rgb]{0.37254902,0.41568627,0.45098039}\makebox(0,0)[t]{\lineheight{1.25}\smash{\begin{tabular}[t]{c}human\\review to\end{tabular}}}}%
    \put(0.62254546,0.12604763){\color[rgb]{0.37254902,0.41568627,0.45098039}\makebox(0,0)[t]{\lineheight{1.25}\smash{\begin{tabular}[t]{c}autonomy\end{tabular}}}}%
    \put(0,0){\includegraphics[width=\unitlength,page=8]{methods_overview_revised.pdf}}%
    \put(0.36236592,0.09232589){\color[rgb]{0.12156863,0.30980392,0.44705882}\makebox(0,0)[lt]{\lineheight{1.25}\smash{\begin{tabular}[t]{l}\textbf{Output of coding stage}\end{tabular}}}}%
    \put(0.36236592,0.0771929){\color[rgb]{0.25490196,0.31764706,0.37254902}\makebox(0,0)[lt]{\lineheight{1.25}\smash{\begin{tabular}[t]{l}Primary-category A$\times$M map for $N = 66$ papers\end{tabular}}}}%
    \put(0.36236592,0.06521414){\color[rgb]{0.25490196,0.31764706,0.37254902}\makebox(0,0)[lt]{\lineheight{1.25}\smash{\begin{tabular}[t]{l}with evidence and readiness fields for synthesis\end{tabular}}}}%
    \put(0,0){\includegraphics[width=\unitlength,page=9]{methods_overview_revised.pdf}}%
    \put(0.85619574,0.31319987){\color[rgb]{1,1,1}\makebox(0,0)[t]{\lineheight{1.25}\smash{\begin{tabular}[t]{c}\textbf{3. Synthesis path through the paper}\end{tabular}}}}%
    \put(0,0){\includegraphics[width=\unitlength,page=10]{methods_overview_revised.pdf}}%
    \put(0.74889445,0.27857543){\color[rgb]{0.14117647,0.19215686,0.23529412}\makebox(0,0)[lt]{\lineheight{1.25}\smash{\begin{tabular}[t]{l}\textbf{Application-level analysis}\end{tabular}}}}%
    \put(0.74889445,0.25806983){\color[rgb]{0.30196078,0.34509804,0.38039216}\makebox(0,0)[lt]{\lineheight{1.25}\smash{\begin{tabular}[t]{l}A1 to A5 family sections\end{tabular}}}}%
    \put(0.74889445,0.24071891){\color[rgb]{0.30196078,0.34509804,0.38039216}\makebox(0,0)[lt]{\lineheight{1.25}\smash{\begin{tabular}[t]{l}Secs.~\ref{sec:a1}--\ref{sec:a5}\end{tabular}}}}%
    \put(0,0){\includegraphics[width=\unitlength,page=11]{methods_overview_revised.pdf}}%
    \put(0.74889445,0.18077948){\color[rgb]{0.12156863,0.30980392,0.44705882}\makebox(0,0)[lt]{\lineheight{1.25}\smash{\begin{tabular}[t]{l}\textbf{Cross-cutting synthesis}\end{tabular}}}}%
    \put(0.74889445,0.16564389){\color[rgb]{0.25490196,0.31764706,0.37254902}\makebox(0,0)[lt]{\lineheight{1.25}\smash{\begin{tabular}[t]{l}A$\times$M concentration (A3$\times$M2 = 15);\\readiness 0 / 3 / 63; evidence gap\\Sec.~\ref{sec:crosscutting}\end{tabular}}}}%
    \put(0,0){\includegraphics[width=\unitlength,page=12]{methods_overview_revised.pdf}}%
    \put(0.74889445,0.0829836){\color[rgb]{0.14117647,0.19215686,0.23529412}\makebox(0,0)[lt]{\lineheight{1.25}\smash{\begin{tabular}[t]{l}\textbf{Discussion outputs}\end{tabular}}}}%
    \put(0.74889357,0.06902546){\color[rgb]{0.30196078,0.34509804,0.38039216}\makebox(0,0)[lt]{\lineheight{1.25}\smash{\begin{tabular}[t]{l}deployment map,responsibility \\boundary, practical guidance, failure \\modes, future work\end{tabular}}}}%
    \put(0.74889445,0.02129933){\color[rgb]{0.30196078,0.34509804,0.38039216}\makebox(0,0)[lt]{\lineheight{1.25}\smash{\begin{tabular}[t]{l}Secs.~\ref{sec:discussion_map}-\ref{sec:discussion_future}\end{tabular}}}}%
    \put(0.49986,0.03964373){\color[rgb]{0.36470588,0.41176471,0.44705882}\makebox(0,0)[t]{\lineheight{1.25}\smash{\begin{tabular}[t]{c}Core methodological logic:\end{tabular}}}}%
    \put(0.49986,0.02340377){\color[rgb]{0.36470588,0.41176471,0.44705882}\makebox(0,0)[t]{\lineheight{1.25}\smash{\begin{tabular}[t]{c}define scope - screen corpus - code papers -\\synthesize patterns - derive discussion outputs\end{tabular}}}}%
  \end{picture}%
\endgroup%